\documentclass[journal,onecolumn,draftclsnofoot]{./IEEEtran}
\usepackage{hyperref}
\usepackage{url}
\usepackage{tcolorbox}
\usepackage{hyperref}

\usepackage{cite}
\usepackage{graphicx}
\usepackage{soul}
\usepackage{amssymb, amsmath}
\usepackage{cleveref}
\usepackage{subcaption}

\begin{document}
%
\title{Robust Federated Learning by Mixture of Experts}
%
%
%

\author{Saeedeh Parsaeefard , Sayed Ehsan Etesami \& Alberto Leon Garcia,~\IEEEmembership{Life~Fellow,~IEEE} 
\thanks{S. Parsaeefard, SE. Etesami and A. Leon Garcia are with the Department of Electrical and Computer Engineering, University of Toronto, Toronto, email:
    \texttt{\{saeideh.fard,alberto.leongarcia\}@utoronto.ca},
    \texttt{\{ehsan.etesami\}@mail.utoronto.ca}
}}
\maketitle

\begin{abstract}
    We present a novel weighted average model based on the mixture of experts (MoE) concept to provide robustness in Federated learning (FL) against the poisoned/corrupted/outdated local models. These threats along with the non-IID nature of data sets can considerably diminish the accuracy of the FL model. Our proposed MoE-FL setup relies on the trust between users and the server where the users share a portion of their public data sets with the server. The server applies a robust aggregation method by solving the optimization problem or the Softmax method to highlight the outlier cases and to reduce their adverse effect on the FL process. Our experiments illustrate that MoE-FL outperforms the performance of the traditional aggregation approach for high rate of poisoned data from attackers.  
\end{abstract}


%
\IEEEpeerreviewmaketitle

\section{Introduction and Motivation}
With the unprecedented growth of smart and highly capable end-user devices, e.g., IoT sensors and smart phones, the power of local data sets is unlocked for future machine learning (ML) applications. In this context, federated learning (FL) is a leading distributed approach where a specific set of end-users participate in the ML training process using their local data sets subject to orchestration by one server \cite{FederatedMachineLearning}. FL brings a lot of practical advantages where privacy is the most highlighted ones. 

\begin{figure}[ht!]
\begin{center}
\includegraphics[width=0.85\textwidth]{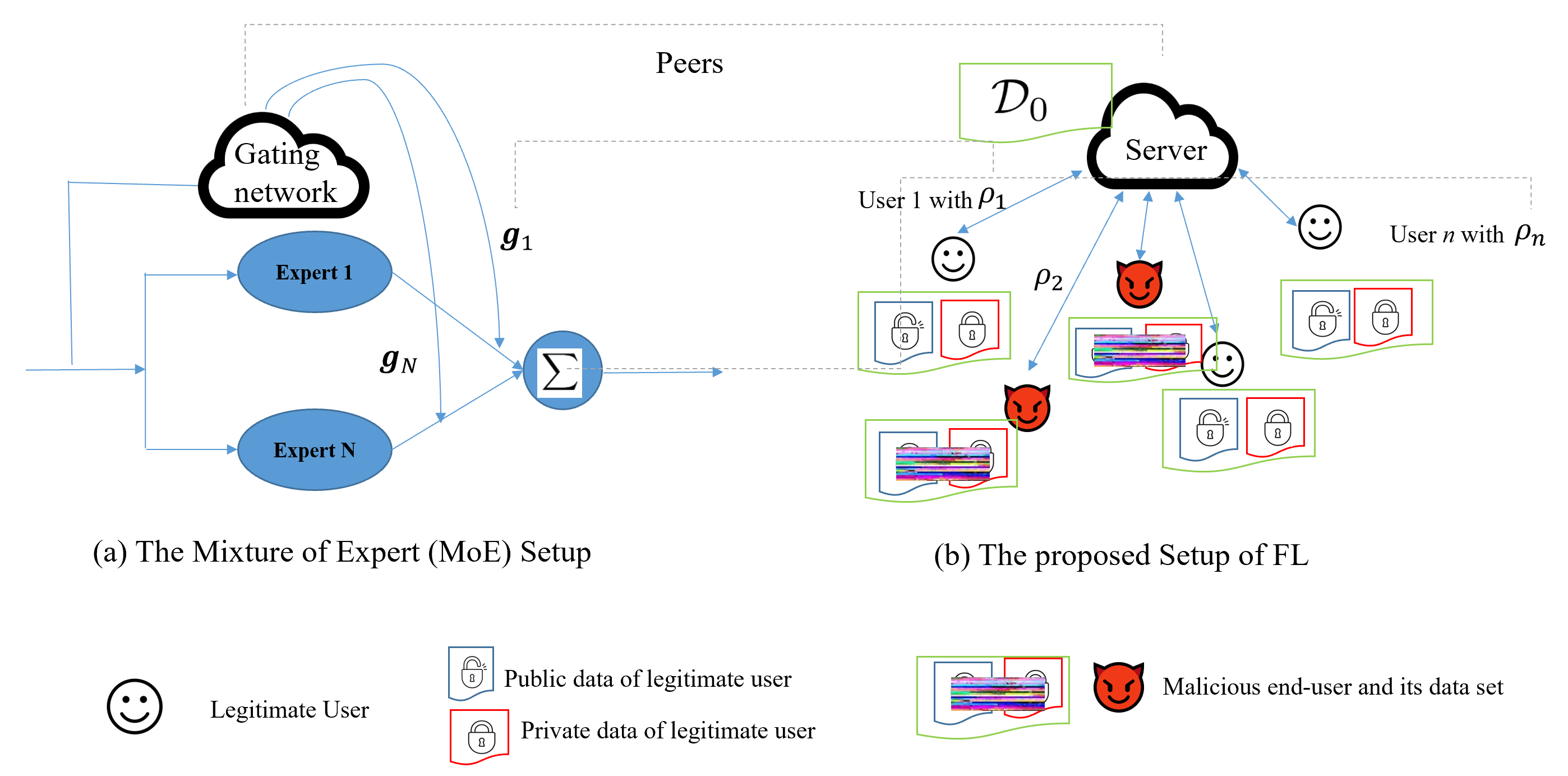}
\end{center}
\caption{The interplay between MoE and FL (a) In MoE, one gating network determines the effect of expert $n$ by $g_n$ at the output \cite{Adaptivemixture}; (b) In FL  where there exist both legitimate users and attackers, server finds $\rho_n$ to reach a weighted average to diminish the effect of attackers}\label{fig1}
\end{figure}

Commonly applied, the server utilizes an element-wise aggregation of the updated users' local models per each round and sends the result back for next iteration. This process is naturally vulnerable against poisoned/corrupted models resulting form suspicious behaviours of attackers or the outdated and noisy versions of users' local data sets (see Fig. \ref{fig1} (b)) \cite{7478523,tramer2020ensemble,DBA}. Consequently, robustness against this type of threats is of interest in this context. There exists a vast body of literature study the effects of these threats in FL, e.g.,   \cite{bagdasaryan2019backdoor,abs-1911-07963,DBA}. The most obvious way to handle these attacks is to find the outlier models and eliminate them from the aggregation process in the server by clipping approaches or robust aggregation \cite{bagdasaryan2019backdoor,abs-1911-07963,NIPS2017_6617, Huber,pillutla2019robust}. However, their applications in FL is challenging due to the non-IID nature of local data sets. 

In this paper, we revisit the FL setup by the concept of mixture of experts (MoE) which is well-known ensemble method \cite{Adaptivemixture,Rida99localexperts} and has been successfully applied for diverse problems, e.g.,  \cite{SVMBengio,Adaptivemixture}. Generally, MOE can provide promising combination methods and it has inherent connection to FL as shown in Fig. \ref{fig1}. In MoE, there exists a set of experts in learning process who are controlled by one gating network imposing a more sophisticated aggregation mechanism to reach a better output. In our presented approach in FL, the server is equipped with public data set which can be applied to calculate the weighted average of models by \textit{optimization problems} or predetermined roles such as \textit{softmax} similar to MoE. We show how this approach helps to reach a robust aggregation against attackers.  

To evaluate this approach, we consider a worst case scenario where the attackers send the reverse values of the aggregated model of legitimate users in each round. We consider both IID and non-IID data sets for users within the server. We also evaluate MoE for both pure and poisoned data sets (impure) of server. For all these scenarios, our proposed approach considerably outperforms the traditional average based algorithm called FedAvg \cite{McMahan2016CommunicationEfficientLO} at least by 50\% in test accuracy. This improvement comes with the price of slower convergence for MoE.

\section{Proposed Setup for FL in Presence of Attackers}
For a set of users $\mathcal{N}=\{1,\cdots,N\}$, 
each user $n$ has local data set $\textbf{z}_n=\{\textbf{x}_n(\text{Features}), \textbf{y}_n(\text{Labels})\} \in \mathcal{D}_n$. The goal is to find a model $\textbf{w}$ from \cite{koloskova2020unified}
$$\min_{\textbf{w} \in \mathbb{R}^d} V(\textbf{w}) := \frac{1}{N} \sum_{n \in \mathcal{N}} V_n(\textbf{w}),\; \forall n \in \mathcal{N},$$
where $V_n(\textbf{w})=\mathbb{E}_{\textbf{z}_n\in \mathcal{D}_n} v_n(\textbf{w},\textbf{z}_n)$ is a loss function of user $n$. Assume that $\textbf{z}_n \in \mathcal{D}_n$ has secure data which cannot be shared  ($\textbf{z}_n^{\textbf{s}}$), and public/less-sensitive data allowed to be shared by third parties ($\textbf{z}_n^{\textbf{p}}$). Here, the server denoted by $0$ is trustworthy and its data set is $\{\textbf{z}_1^{\textbf{p}}, \cdots \textbf{z}_n^{\textbf{p}}\} \in \mathcal{D}_0$. Among users, there is a set of attackers $\mathcal{E} \subseteq \mathcal{N}$ trying to manipulate the learning process. To mathematically represent the behaviour of attackers, let's modify the local loss of $n \in \mathcal{E}$ as $\tilde{V}_n(\textbf{w})=\mathbb{E}_{\textbf{z}_n\in \widetilde{\mathcal{D}}_n} \tilde{v}_n(\textbf{w},\textbf{z}_n)$ in which attackers change $\mathcal{D}_n$ to $\widetilde{\mathcal{D}}_n$ to poison or to manipulate local models derived  by $\widetilde{\mathcal{D}}_n$. 
Depending on the capabilities of attackers, they can use different approaches. For example, attackers alter their distribution of data where $\widetilde{\mathcal{D}}_n \neq \mathcal{D}_n$ once and update the model based on manipulated data, i.e., \textit{Static data poisoning}. In \textit{Adaptive data poisoning}, in each iteration, the attackers adjust $\widetilde{\mathcal{D}}_n(\textbf{w})$. 

\textbf{Mixture of Experts based algorithm for FL (MoE-FL)} contains two major steps which are summarized in Table 1. First step involves to learn \textbf{local mode update} where in each time round $t$, each user solves its own local problem from 
\begin{equation}\label{local}
    \min_{\textbf{w}_n \in \mathbb{R}^d} V_n(\textbf{w}_n), \,\,\, \forall n \in \mathcal{N}/\mathcal{E}, \, \text{or} 
\min_{\textbf{w}_n \in \mathbb{R}^d} \widetilde{V}_n(\textbf{w}_n), \,\,\, \forall n \in \mathcal{E},
\end{equation}
to derive $\textbf{w}_n^t$. 
2) The second step deals with the aggregation of the users' local models in the server. For our model, the sever applies \textbf{Mixture of Expert Aggregation (MoEA)}. In this part, the server aims to assigns $\rho_n \in [0,1]$ for all $n$ where $\sum_{n \in \mathcal{N} }\rho_n =1$ and the server applies the following weighted average for the server as

\begin{eqnarray}\label{updatesss}
    \phi(\textbf{w}) : \longrightarrow \textbf{w}^t = \sum_{n \in \mathcal{N}} \rho_n^t\textbf{w}_n^t. 
\end{eqnarray}
To derive the values of $\rho_n$, we assume that the server also updates its own model from $\mathcal{D}_0$ as

\begin{equation} \label{2server}
    \min_{\textbf{w}_0 \in \textbf{W}_0} V_0(\textbf{w}_0, \textbf{z}_0), \,\,\,\,\,\, \forall (\textbf{x}_0, \textbf{y}_0) \in \mathcal{D}_0. 
\end{equation}

Then, we apply two main methods from MoE context: I) Softmax algorithm to find  $\rho_n$  as \begin{eqnarray}\label{MOEsoft}\rho_n(\textbf{w}_0^t,\textbf{w}_n^t)=\frac{\exp((\textbf{w}_0^t)^T\textbf{w}_n^t)}{\sum_{n\in \mathcal{N}} \exp((\textbf{w}_0^t)^T\textbf{w}_n)}, \quad \forall n \in \mathcal{N}.\end{eqnarray} II) Optimization approach where the server applies the following optimization problem 
\begin{eqnarray}\label{MOEoptimization}
{\min_{\boldsymbol{\rho}=[\rho_1, \cdots, \rho_N]}  \sum_{n=1}^N \rho_n^t\| \textbf{w}_n^t- \textbf{w}_0^t\|,} \quad \text{subject to:}
\sum_{n=1}^N \rho_n^t=1.
\end{eqnarray}

Algorithm 1 presents MoE-FL in Table 1. Note that For $\rho_n=\frac{1}{N}$, MOE-FL converges to FedAvg \cite{McMahan2016CommunicationEfficientLO}.

\begin{table*}[t]
	\caption{Algorithm 1: Mixture of Expert based FL  (MoE-FL)} \centering \vspace{-.2 in }\begin{tabular}{l}
	\\	\hline \hline \textbf{Step 0}: Users send their public data to the server and the server assigns $T$ and $\textbf{w}^t$
		\\ \textbf{Iterative Algorithm}: For $t=1,\cdots,T$ and $1\ll T$, 
		\\ $\quad $ \textbf{Local Update}:  For all $n \in \mathcal{N}$, in parallel: derive $\textbf{w}_n^t$ from \eqref{local} and send to the server 
		\\ $\quad$ \textbf{MoEA}: Server applies mixture of server aggregation from \eqref{2server}  \\ If $\|\textbf{w}^{t-1}-\textbf{w}^{t}\|_{2}\leq \zeta$, End; Otherwise $t=t+1$, continue;
		\\ \hline \hline
		\vspace{-0.0 in}
	\end{tabular}\label{distributedalgorith2}
\end{table*}

\section{Performance Study of the MoE-FL}
The question is how we can analyse the robustness of MoE-FL against the attackers. Basically, our proposed approach by MoEA. Basically, \eqref{MOEoptimization} falls into the robust statistic i.e., \cite{Huber}, robust Byzantine, i.e., \cite{NIPS2017_6617}, and robust aggregation with geometric mean \cite{pillutla2019robust}. Here, instead of $\varepsilon$-robust estimation of geometric mean of the central parameter, we consider weights of server $\textbf{w}_0$. Therefore, our algorithm is more practical from computation perspective, while it is more sensitive to $\mathcal{D}_0$ as we will show through some scenarios as follows. 

\begin{tcolorbox}
Assumptions to study MoE-FL.

\textbf{I)} To show the heterogeneity in $V_n(\textbf{w})$, we assume that at the optimal model, there is a bounded gradient as $ \|\nabla V_n\| \leq  \eta_n^1$ for $n \in \mathcal{N}/\mathcal{E}$ and  $ \|\nabla \widetilde{V}_n\| \leq  \eta_n^2$ for $n \in \mathcal{E}$ and $ \frac{1}{N-E} \sum _{n \in \mathcal{N}} \eta_n^1 \leq \eta^1$ and $ \frac{1}{E} \sum _{n \in \mathcal{E}} \eta_n^2 \leq \eta^2$ which can capture the level of heterogeneity among users and the legitimate and attackers.

\textbf{II)} We assume that $\sigma_n^1=\mathbb{E}_{\textbf{z}_n}\|\nabla v_n (\textbf{w}_n, \textbf{z}_n)- \nabla V_n(\textbf{w}_n)\|$ for $n \in \mathcal{N}/\mathcal{E}$ and $\sigma_n^2=E\|\nabla \tilde{v}_n (\textbf{w}_n, \textbf{x}_n)- \nabla \tilde{V}_n(\textbf{w}_n)\|$ for $n \in \mathcal{E}$; and  $ \frac{1}{N-E} \sum _{n \in \mathcal{N}} \sigma_n^1 \leq \sigma^1$ and $ \frac{1}{E} \sum _{n \in \mathcal{E}} \sigma_n^2 \leq \sigma^2$.

\textbf{III)} $V_n$ and $\tilde{V}_n$ are smooth and differentiable for each $\textbf{z}_n \in \mathcal{D}_n$ and $L$-smoothness, and there exist a constant $L_1 \text{and} L_2 \geq 0 $ such that for each $\textbf{w}_1$ and $\textbf{w}_2$, we have $\|\triangledown {V}_n(\textbf{w}_1)-{V}_n(\textbf{w}_2)\|\leq L_1 (\textbf{w}_1- \textbf{w}_2)$ and $\|\triangledown \tilde{V}_n(\textbf{w}_1)-  \tilde{V}_n(\textbf{w}_2)\|\leq L_2 (\textbf{w}_1- \textbf{w}_2)$, consequently.

\textbf{IV)} $V_n(\textbf{w})$ and $\tilde{V}_n(\textbf{w})$ $\mu$-(strongly) convex for $\mu \geq 0$, if 
    $$\triangledown {V}_n(\textbf{w}_1)-  {V}_n(\textbf{w}_2)+\frac{\mu}{2}\|\textbf{w}_1- \textbf{w}_2\|_2^2 \leq \nabla V_n(\textbf{w}_1)
    (\textbf{w}_1- \textbf{w}_2)$$
    $$\triangledown \tilde{V}_n(\textbf{w}_1)-  \tilde{V}_n(\textbf{w}_2)+\frac{\mu}{2}\|\textbf{w}_1- \textbf{w}_2\|_2^2 \leq \nabla \tilde{V}_n(\textbf{w}_1)
    (\textbf{w}_1- \textbf{w}_2)$$

\end{tcolorbox}


Let's first assume that all users are legitimate, for the IID data sets, the MoEA is the unbiased function and the convergence rate of MOE-FL can be studied as follows. 

\textbf{Lemma 1}: If all users are legitimate and $N \gg 1$ users, the convergence rate of the MoE-FL is reduced by the factor  $O(\frac{1}{\sqrt{N-N_{\text{MoE}}}})$ for non-convex $V_n$ and $O(\frac{1}{N-N_{\text{MoE}}})$ for convex $V_n$ where $N_{\text{MoE}}$ is the number of users with $\rho_n=0$ from \eqref{MOEoptimization}.

\textit{ Proof: }  See Appendix A. $\Box$

From Lemma 1, MoE-FL cannot improve the performance of the system when all users belong to  $\mathcal{N} \setminus \mathcal{E}$ and IID since there is a chance that a set of users are removed by \eqref{MOEsoft} or \eqref{MOEoptimization} in MoEA. In Appendix A, we also show that MoEA is a biased function over the non-IID data sets for the same scenario. It means that the performance of MoE-FL compared to the FedAvg depends on the number of users and their noise in the data sets and level of heterogeneity ( See \eqref{AppendixA-2} in Appendix A). When there are both legitimate users and attackers and $\mathcal{D}_0$ has data from both groups, the level of biased is increased in MoEA ( See \eqref{AppendixA-3} in Appendix A) and its a linear function of attackers' features, i.e., number of attackers ($E$), the values of $\sigma_n^2$, and $\eta^2_n$ for $n \in \mathcal{E}$ as well as the legitimate users' features, i.e., number of legitimate users ($N-E$), the values of $\sigma_n^1$, and $\eta^1_n$ for $n \in \mathcal{N}/\mathcal{E}$. First, this shows that the heterogeneity of users. We also show that if $\mathcal{D}_0$ contains only the data from the legitimate users, bias of MoE-FL is reduced considerably (See \eqref{AppendixA-4}) and there is a higher chance that MoEA can detect the attackers. In the following, we consider these two scenarios based on the data samples of the server in MoE-FL.

From the above discussions, we consider two cases: 
\begin{itemize}
    \item $\mathcal{D}_0$ only includes data sets of legitimate users which is called pure data sets or MoE-P, 
    \item $\mathcal{D}_0$ includes samples from both legitimate users and attackers which is called impure, impure or MoE-NP 
\end{itemize}
We aim to study under which conditions MoE can detect the attackers. Clearly, the MOE can detects the attackers if their models are far enough from $\textbf{w}_0$ which can be measured through the dissimilarity of data distributions. For study this, we assume that for all users $n$ and server, a total of $T$ stochastic gradients with $R$ times of communication among server and users are run and we have $K = T /R$ and assume that the optimal solution of the model is bounded and has value less than $B$, i.e., $\|\textbf{w}^*\| \leq B$ \cite{NEURIPS2020_45713f6f}. Now, we study the variance of noise of attackers through the following Lemma. 

\textbf{Lemma 2}. When the server and the users utilize the SGD with appropriate value of step size, and there is at least one legitimate user, we have: 
\begin{itemize}
    \item For pure $\mathcal{D}_0$,  MOE can detect attackers with 
    \begin{equation}\label{noise1}
        \sigma_n \geq 2(\varsigma+\nu), \quad \quad \quad  \forall n \in \mathcal{E},
    \end{equation} where $\varsigma=\frac{L_1}{t}+ \frac{\sigma^1}{\sqrt{NKt}}$ and $\nu=\frac{(L_1 (\eta^1)^2 B^4)^{1/3}}{R^{2/3}}+\frac{(L_1 (\sigma^1)^2 B^4)^{1/3}}{K^{1/3}R^{2/3}}$ in which $t=\frac{R}{B^2}$. 
    \item For non-pure $\mathcal{D}_0$ if the portions of samples from legitimate users and attackers in $\mathcal{D}_0$ is related to $E$ and $N-E$, MOE can detect attackers with 
\begin{equation}\label{noise2}
    \sigma_n \geq \varsigma+\eta+\frac{N-E}{N}(\eta_1 +\nu_1)+\frac{E}{N}(\eta_2 +\nu_2),  , \quad \quad \quad  \forall n \in \mathcal{E},
\end{equation} where $\varsigma_1=\frac{L_1}{t}+ \frac{\sigma^1}{\sqrt{(N-E)Kt}}$ and $\nu_1=\frac{(L_1 (\eta^1)2 B^4)^{1/3}}{R^{2/3}}+\frac{(L_1 (\sigma^1)^2 B^4)^{1/3}}{K^{1/3}R^{2/3}}$ $\varsigma_2=\frac{L_2}{t}+ \frac{\sigma_2}{\sqrt{EKt}}$ and $\nu_2=\frac{(L_2 (\eta^2)^2 B^4)^{1/3}}{R^{2/3}}+\frac{(L_2 (\sigma^2)^2 B^4)^{1/3}}{K^{1/3}R^{2/3}}$ .
\end{itemize}
\textit{ Proof: See Appendix B}. $\Box$   
 
In \eqref{noise1}, $\varsigma$ comes form the update of the model and $\nu$ shows the effect of heterogeneous data sets. Similarly, in \eqref{noise2}, in case that the data sets of non-attackers are more heterogeneous, the attackers with higher level of noise can be detected compared to the case that non-attackers users are heterogeneous. Comparing \eqref{noise1} and \eqref{noise2} shows $\sigma_n$ of the detectable attackers are greater when $\mathcal{D}_0$ is impure, which means impure $\mathcal{D}_0$ misleads the values of $\rho_n$ in MoE-FL and attackers with higher level of noise in their data set are detectable. This bound is also directly influenced by the number of attackers and when the number of attackers is larger than the legitimate users, the variance, noise and dissimilarities of attackers are dominant in the server model, i.e., the last part of \eqref{noise2}. This means the parameters of the server model tends to be close to the attackers' parameters. Then the weighted average calculation is completely misled by the values of $\rho_n$ in MoEA. 

The above discussions shows the importance of $\mathcal{D}_0$ to assign effective $\rho_n$. This issue cannot be handled in MoE-FL unless the server can trust some of the users to build or fine-tune its own model. We propose some mechanisms to assist the server for this purpose. Assume that server acquaintances with "trustful users" (e.g., $n \in {\mathcal{T}}$), and the rest of users are unknown to the server (untruthful users ($n \in {\mathcal{U}}$)) where $\mathcal{N}=\mathcal{T} \cup \mathcal{U}$. The server can use the  following mechanisms:  
\begin{itemize}
    \item Online algorithm where the server starts with $\textbf{w}_0^{t=0}$ from users in $\mathcal{T}$, and at each $t$, it adds data of one user in $\mathcal{U}$ to its own data sets. For $d_0 > 0$, if $\|\textbf{w}_0^{t-1}- \textbf{w}_0^{t}\| \geq d_0$, the server discards the data samples and inserts "outlier labels" for user $n$. 
    \item Stochastic algorithm: In this case, the server collects all the public samples of all users. It starts from $\textbf{w}_0^{t=0}$ and then, each iteration, randomly select a number of samples from all users in $\mathcal{U}$ and then it updates $\textbf{w}_0^{t}$. If $\|\textbf{w}_0^{t-1}- \textbf{w}_0^{t}\| \geq d_0$, then it discards the update of that iteration and remove all samples at iteration $t$ from its own data sets. Otherwise the weights of server is updates. 
    \item Data sorting approach: In this case, the server sorts the data for its learning process where the data sets of $\mathcal{U}$ are used first and then the data set's of $\mathcal{T}$ are picked for better tuning. 
\end{itemize}
For the objective function in \eqref{MOEoptimization}, we can have more general form such as $\min_{\boldsymbol{\rho}=[\rho_1, \cdots, \rho_N]} \sum_{n=1}^N  u_n\left(\rho_n , \textbf{w}_n^t,  \textbf{w}_0^t\right)$
where $u_n(.)$ can be linear function ($u(\textbf{z})=\textbf{z}$), exponential function ($u(\textbf{z})=\log{\textbf{z}}$), or logarithmic function ($u(\textbf{z})=\exp{\textbf{z}}$). This can be considered as a hyper parameter of MoE-FL in Algorithm 1 which should be selected according to the applications' specifications. 

\section{Evaluation Results}
To study the performance of the proposed algorithm, we evaluate MoE-FL to predict the handwritten digits based on a deep neural network (DNN) using the MNIST data set \cite{cohen2017emnist}. The MNIST data set can be used to model both the IID and non-IID partitioning of data. In the IID setting, the data is first shuffled, then divided into 200 shards and finally partitioned into 100 clients. In the non-IID setting, the data is first sorted by its labels and then shuffled, divided and partitioned into clients \cite{McMahan2016CommunicationEfficientLO}. We use PySyft library \cite{ryffel2018generic} to ensure decoupling the private data from the model training in the federated environment within the PyTorch deep learning framework and the code is available on GitHub \footnote{\url{https://github.com/etesami/MOE-FL}}.

For MoE-FL, we simulate \eqref{MOEoptimization} where there are 100 users for MNIST data sets and each round $30$ users ($N=30$) are randomly selected by server to update the model based on batch SGD with batch size 20. We apply the approach in \cite{DBLP:journals/corr/McMahanMRA16} to derive the IID and non-IID sets. Our initial study includes DNN with two $5\times5$ convolution layers (the first with
20 channels, the second with 50, each followed by $2\times2$ 
max pooling layer). We set the log-likelihood loss function ($nl_{loss}$ function in Pytorch), $T=500$, $\zeta=0.01$, and the learning rate is $0.01$. Note that the results of \eqref{MOEsoft} are similar with the results of \eqref{MOEoptimization}. 

The performance of MoE-FL is compared to that of FedAvg in terms of the accuracy of test phase and the training loss of the server (the average and weighted average of the training losses over all users in FedAvg and MoE-FL) over training phase. We perform the experiment by changing the number of attackers where $E=[25, 50, 75]$. At the beginning of each round $t$, $30$ users are randomly selected and the server sends the model to each of these users. Each client then performs the local computation and sends back its updates to the server. Finally, the server applies the MoE-FL and returns the model to all users. 

\begin{figure} [ht!]
\begin{center}
\begin{subfigure}{.48\textwidth}
  \includegraphics[width=0.98\textwidth]{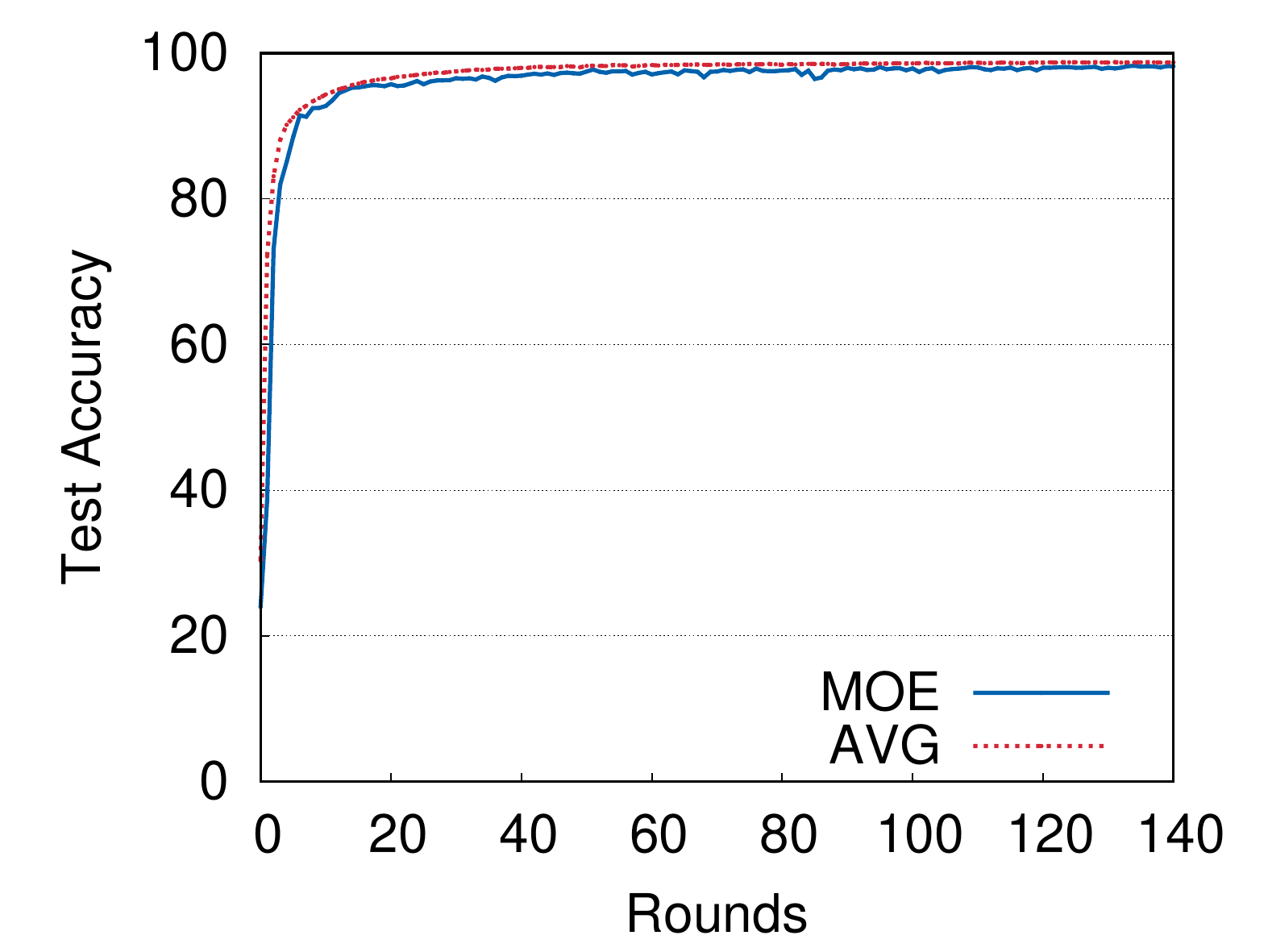}
  \caption{IID Data sets}
  \label{fig:niid_accuracy_base:iid}
\end{subfigure}
\begin{subfigure}{.48\textwidth}
  \includegraphics[width=0.98\textwidth]{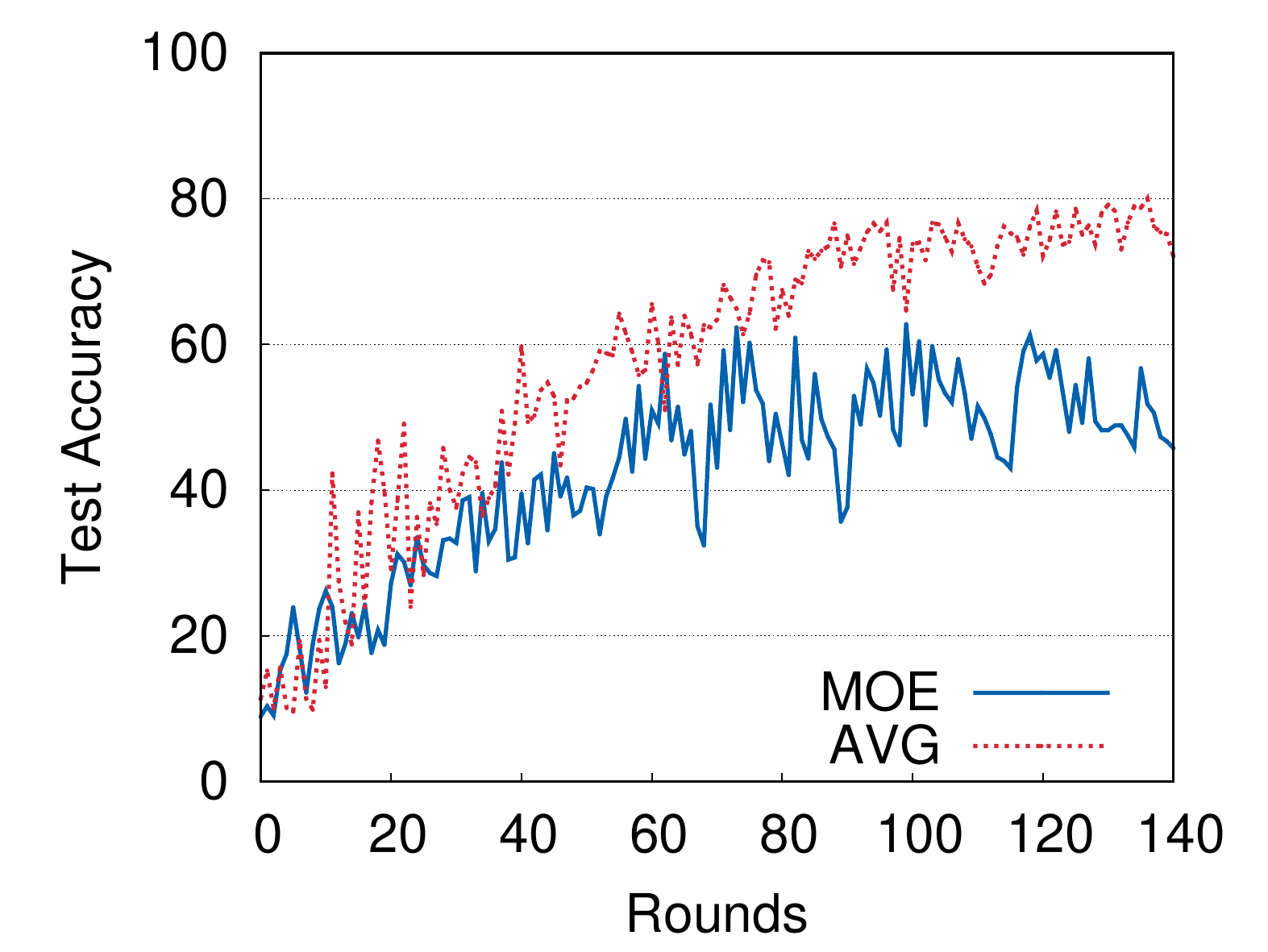}
  \caption{Non-IID Data sets}
  \label{fig:niid_accuracy_base:niid}
\end{subfigure}
\end{center}
\caption{Accuracy of FedAVG vs. MOE-FL for IID (\ref{fig:niid_accuracy_base:iid}) and Non-IID (\ref{fig:niid_accuracy_base:niid}) data sets with $E=0$ i.e. no Attackers (baseline)}
\label{fig:niid_accuracy_base}
\end{figure}

To have a benchmark for the results of FedAvg and MoE-FL under attack, we first show their accuracy versus number of rounds when there is no attackers in Fig. \ref{fig:niid_accuracy_base}. As expected from Lemma 1, the FedAvg outperforms MoE-FL in terms of accuracy and convergence due to a larger number of users in regular average method compared to MoEA (wighted sum in \eqref{updatesss}) for both IID and non-IID data sets. The highlight is that the accuracy of non-IID data sets is considerably less than that IID data sets as supported by \cite{Li2020FairRA} and it converges around 80\% after 140 round in FedAvg with non-IID data sets, while the maximum accuracy for IID data set converges to 100\%.

In the followings, we focus on two scenarios for MOE-FL: First, when the server is provided with pure data sets from legitimate users; Second, when the server's data set is manipulated by attackers (impure data sets) which are shown by MoE-P and MoE-NP in the legends of experiments, respectively. Each user shares 15\% of its samples with the server. In MoE-NP, the data sets of attackers are generated by shuffling all pixels of each sample in the data set. We consider both the IID and non-IID data sets for our experiments. We assume attackers simply choose their models weights randomly to poison the updated model in the server. We also evaluate the effects of more sophisticated approach taken by attackers in section \ref{effect_of_attacks}

\subsection{IID Local Data Sets} 

\begin{figure}
\begin{center}
\begin{subfigure}{.32\textwidth}
  \centering
  \includegraphics[width=.98\linewidth]{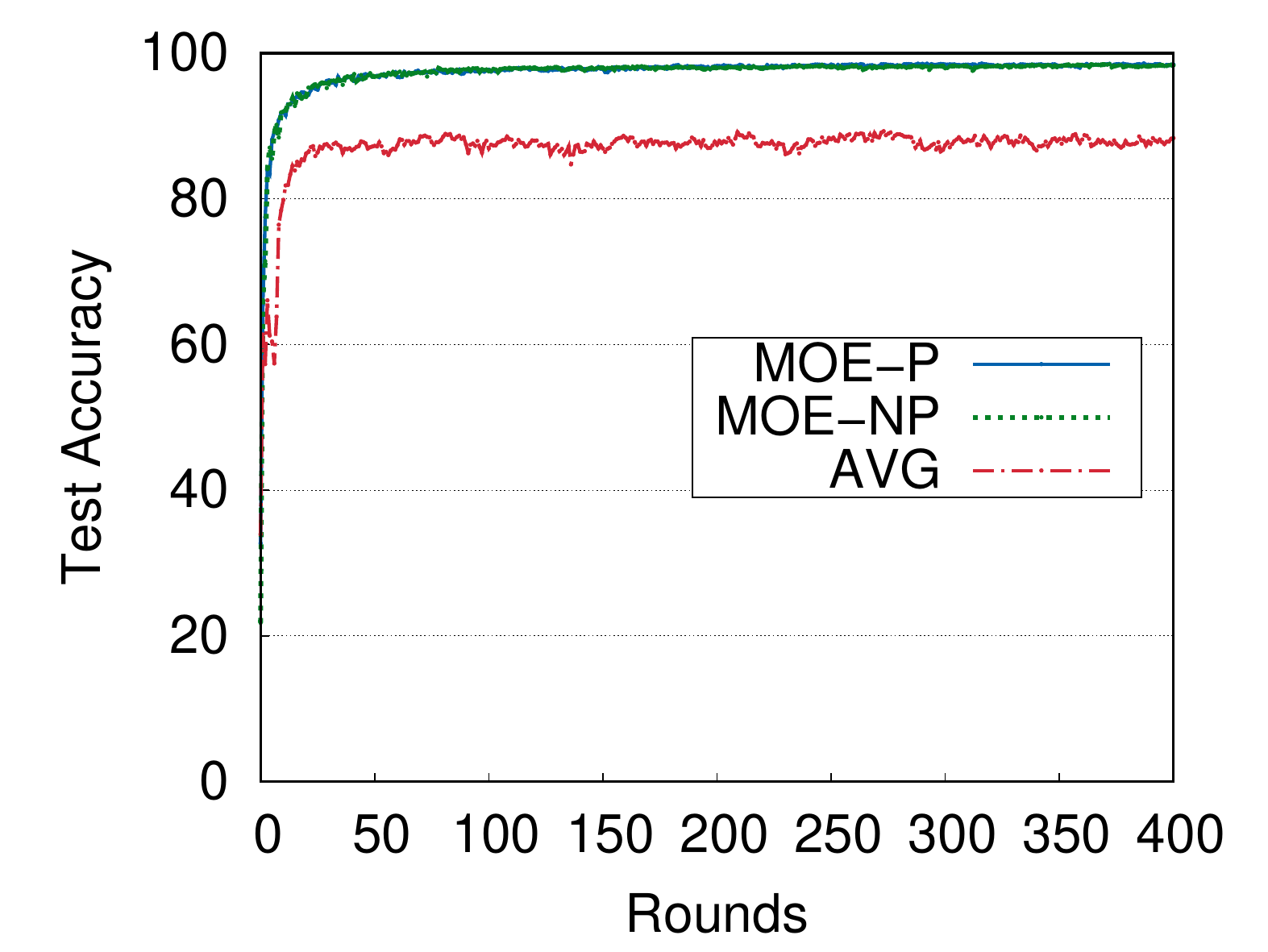}
  \caption{25 attackers}
  \label{fig:iid_accuracy_25}
\end{subfigure}%
\begin{subfigure}{.32\textwidth}
  \centering
  \includegraphics[width=.98\linewidth]{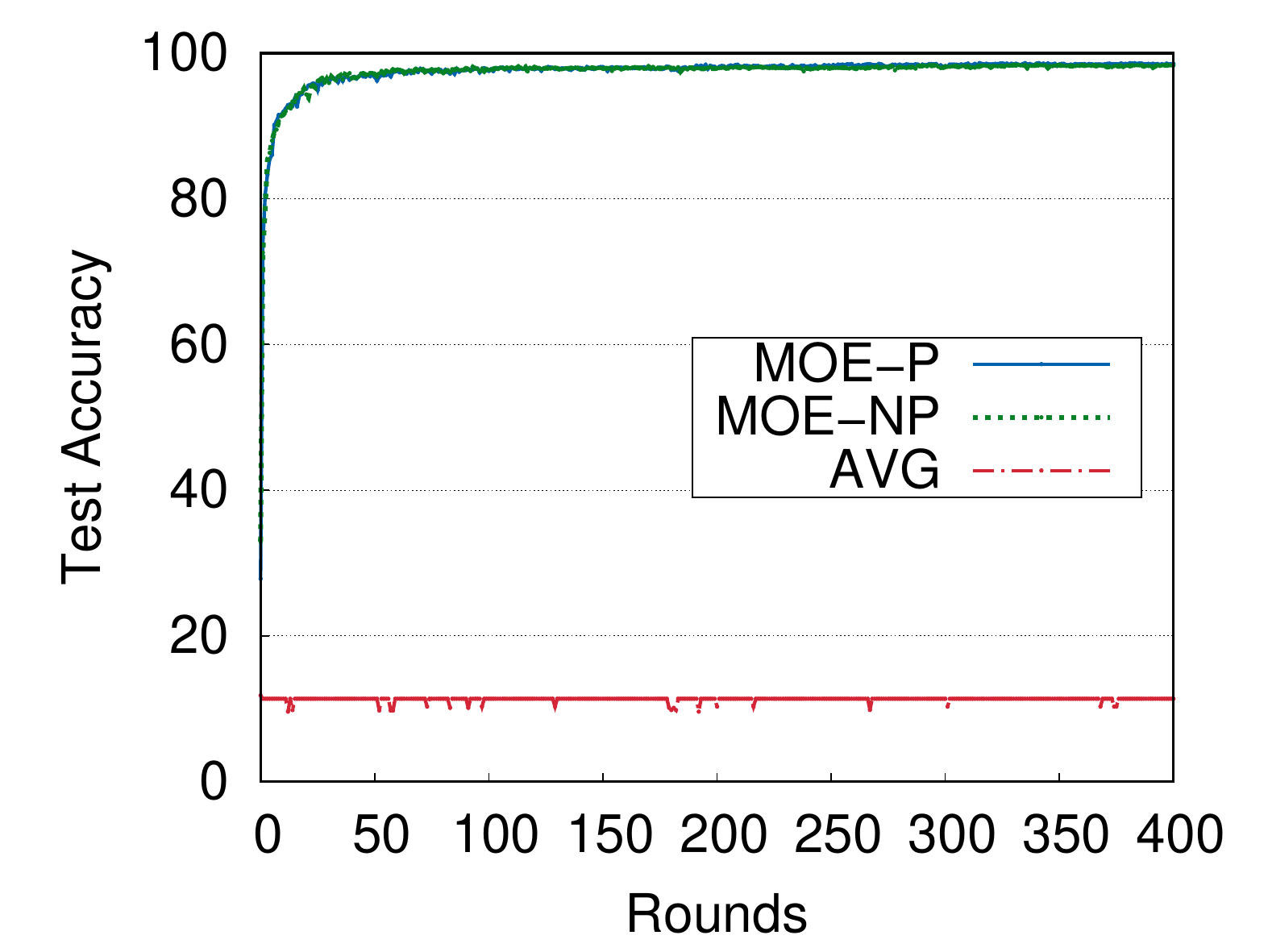}
  \caption{50 attackers}
  \label{fig:iid_accuracy_50}
\end{subfigure}
\begin{subfigure}{.32\textwidth}
  \centering
  \includegraphics[width=.98\linewidth]{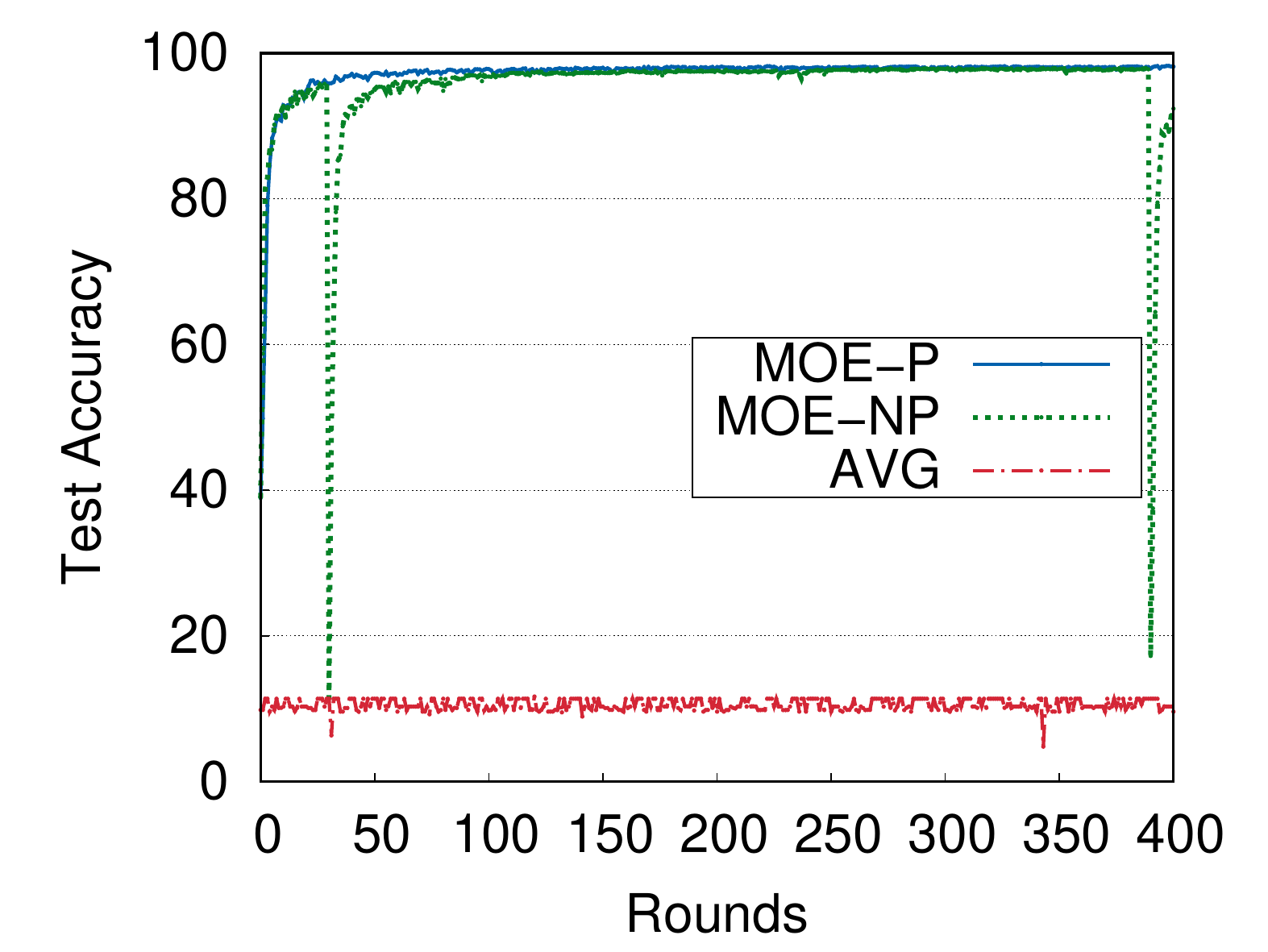}
  \caption{75 attackers}
  \label{fig:iid_accuracy_75}
\end{subfigure}

\begin{subfigure}{.32\textwidth}
  \centering
  \includegraphics[width=.98\linewidth]{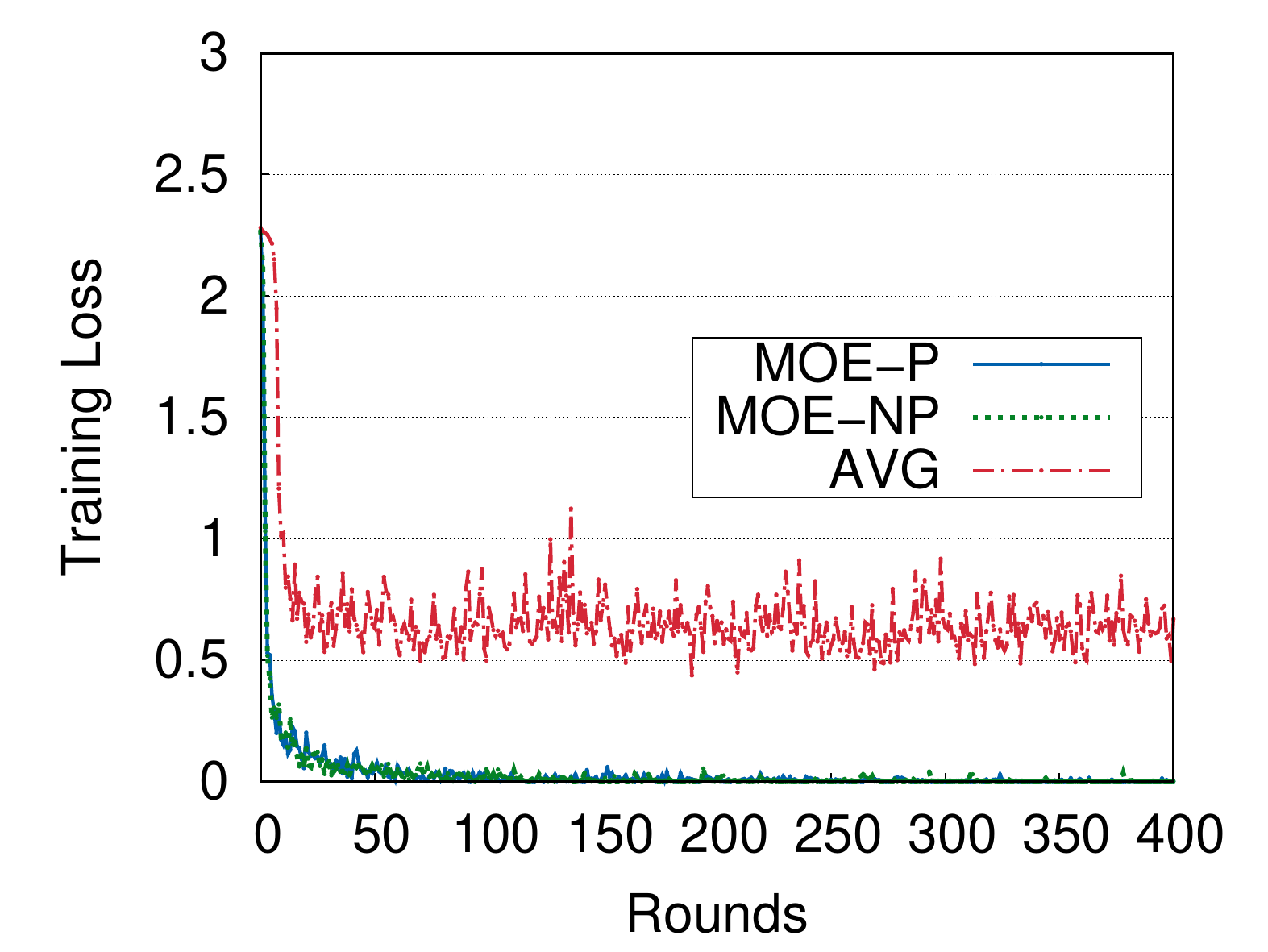}
  \caption{25 attackers}
  \label{fig:iid_loss_25}
\end{subfigure}%
\begin{subfigure}{.32\textwidth}
  \centering
  \includegraphics[width=.98\linewidth]{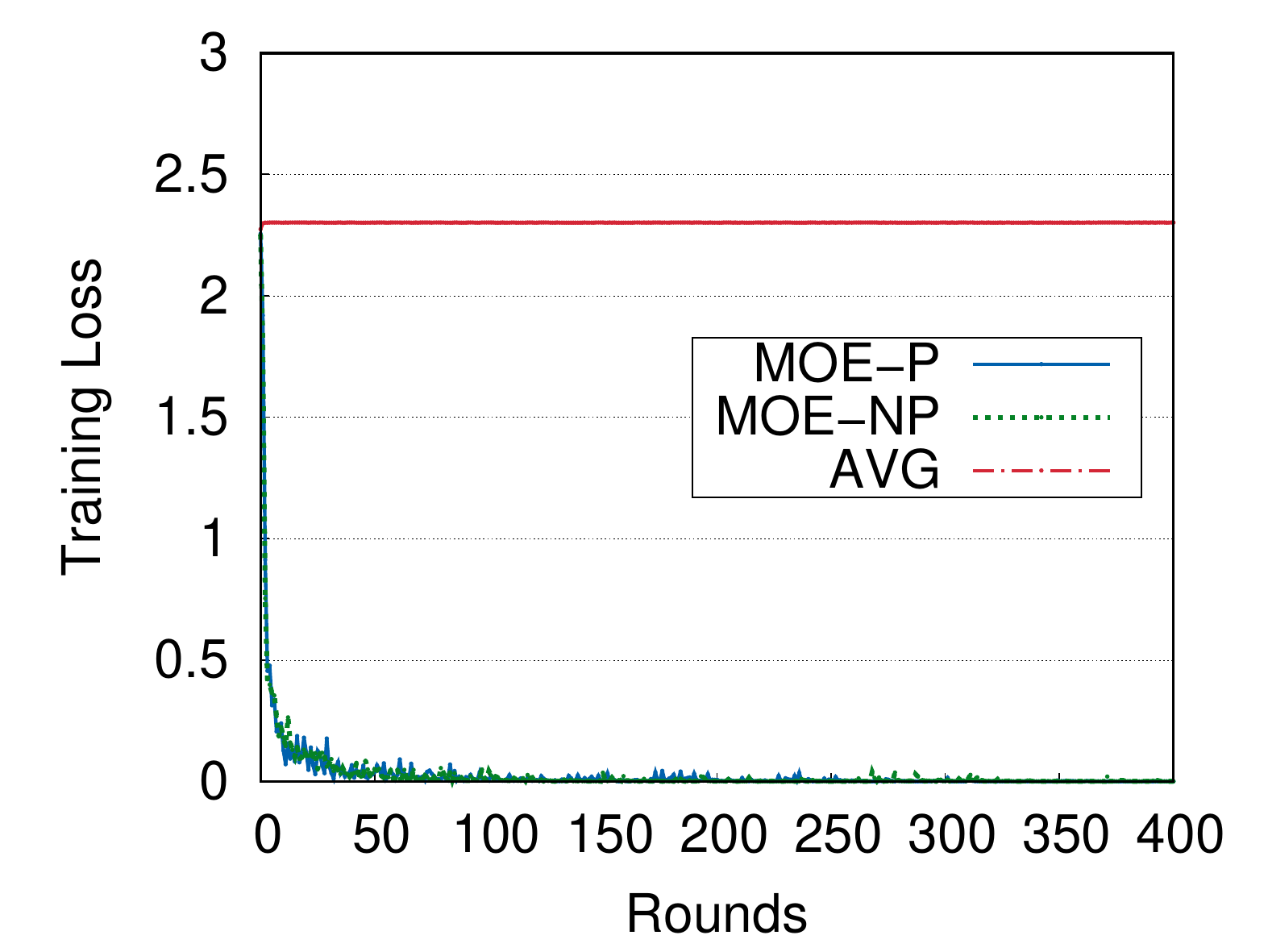}
  \caption{50 attackers}
  \label{fig:iid_loss_50}
\end{subfigure}
\begin{subfigure}{.32\textwidth}
  \centering
  \includegraphics[width=.98\linewidth]{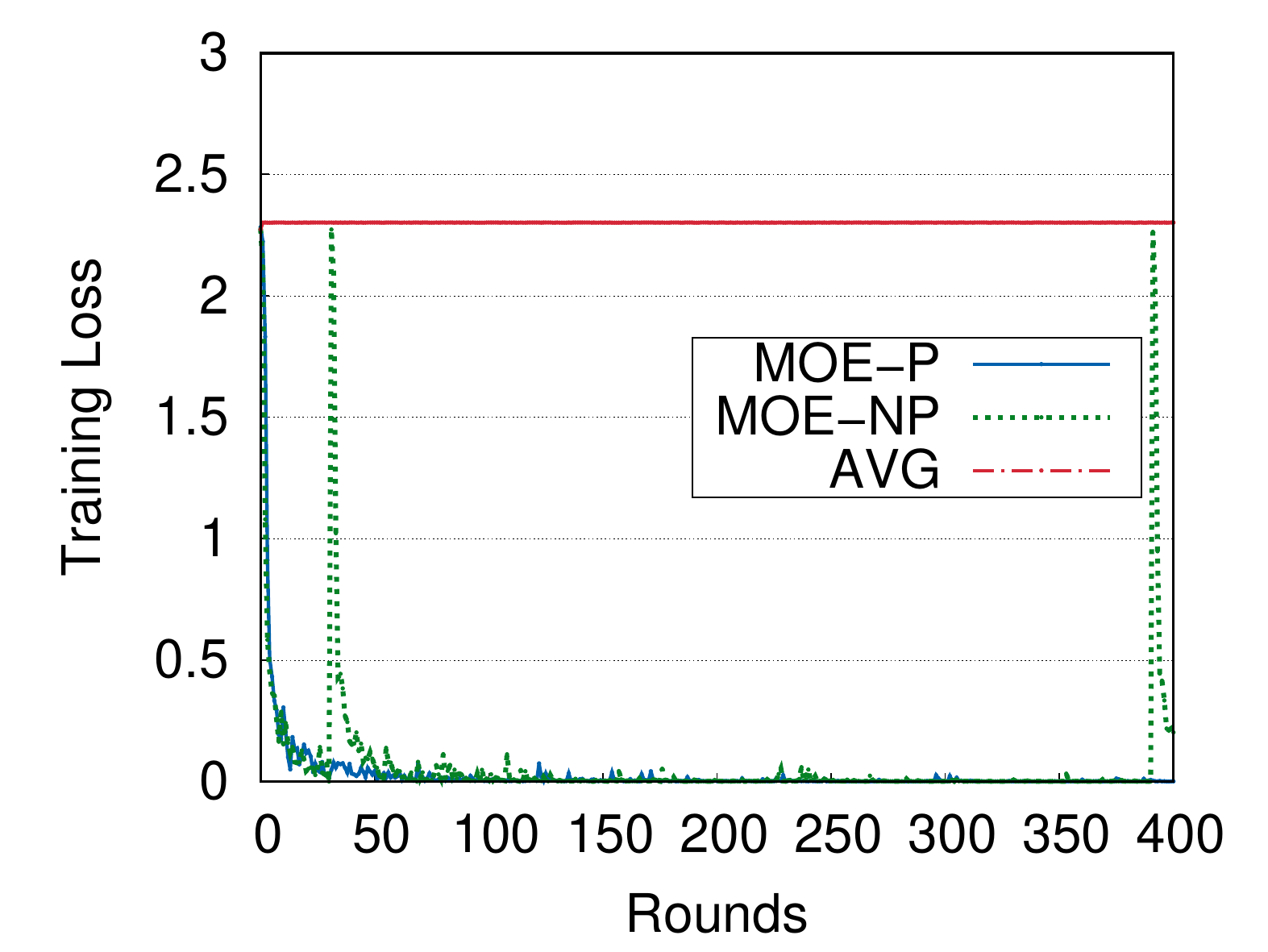}
  \caption{75 attackers}
  \label{fig:iid_loss_75}
\end{subfigure}
\end{center}
\caption{Accuracy and train Loss in server versus number of rounds and number of attackers for IID data sets}
\label{fig:iid_accuracy_loss}
\end{figure}

For this case, all $\mathcal{D}_n$s are IID. In Fig. \ref{fig:iid_accuracy_loss}, the accuracy and loss are shown versus rounds and $E$. Clearly, from Figs. \ref{fig:iid_accuracy_loss} (a) and (d), when the number of attackers is not considerable, i.e., $E=25$, MoE outperforms FedAvg by 10\% and and the loss is close to zero for MoE-FL. By increasing the number of attackers, the MoE considerably maintain the performance for both loss and accuracy while FedAvg is mislead by the attackers, as it is shown in Figs. \ref{fig:iid_accuracy_loss} (e) and (f). The important point is that for $E=25$ and $E=50$, pure and impure $\mathcal{D}_0$ do not have considerable effect on the performance of MoE-FL for both the accuracy and training loss. However, for a larger number of attackers i.e., $E=75$, MoE-FL in some parts of algorithms is mislead by the impure data sets due to assigning $\rho_n \neq 0$ to the attackers. It is worth mentioning that in each round, 30 users are chosen randomly to participate in the training process and the larger number of attackers in the system affects the performance as in rounds 50 and 400. However, the performance can be improved in the next rounds while the FedAVg is not robust enough against the larger number of attackers (e.g. E>=50).

\subsection{Non-IID Local Data Sets for All users} 
\begin{figure}
\begin{center}
\begin{subfigure}{.32\textwidth}
  \centering
  \includegraphics[width=.98\linewidth]{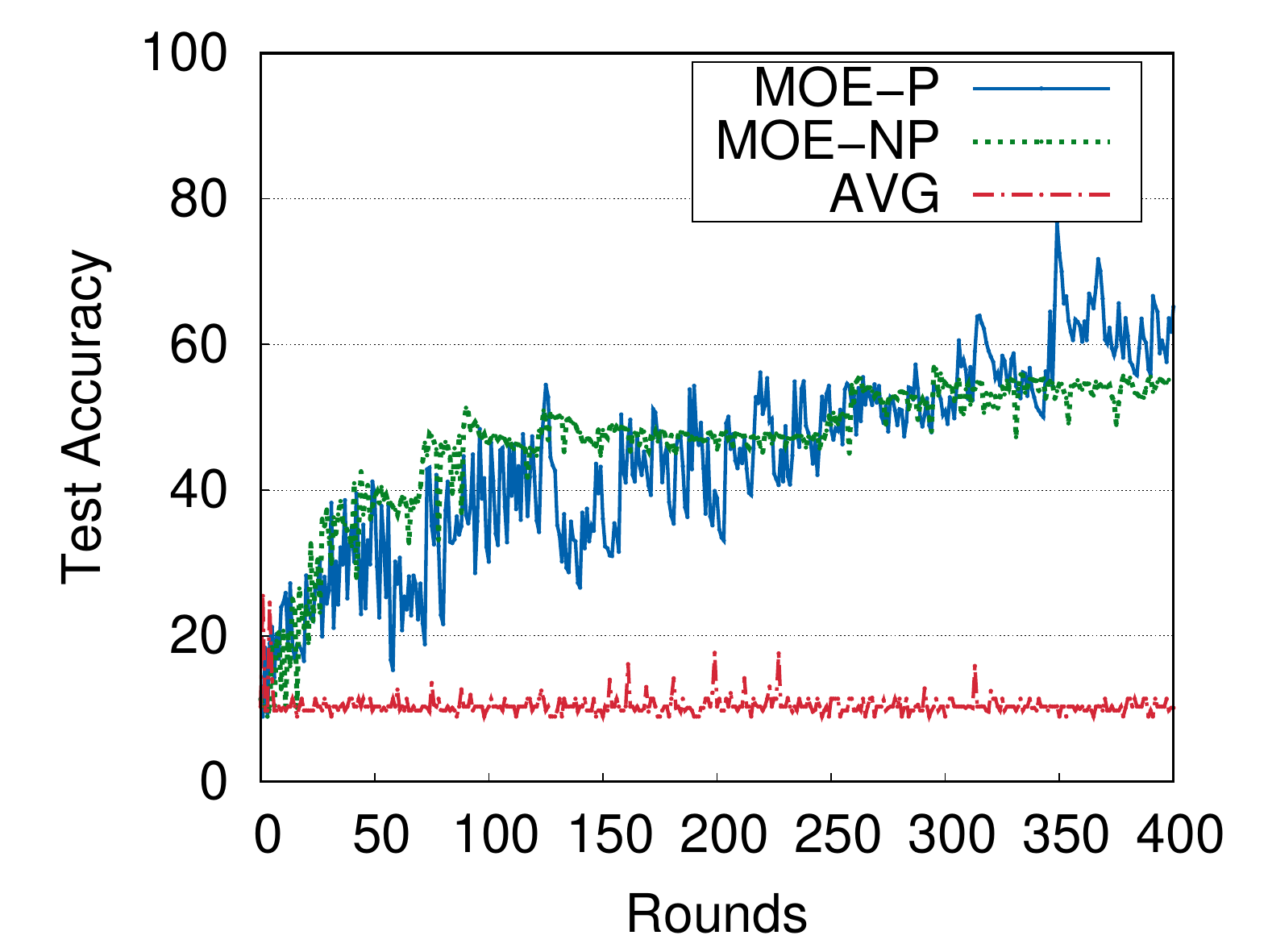}
  \caption{25 attackers}
  \label{fig:niid_accuracy_25}
\end{subfigure}%
\begin{subfigure}{.32\textwidth}
  \centering
  \includegraphics[width=.98\linewidth]{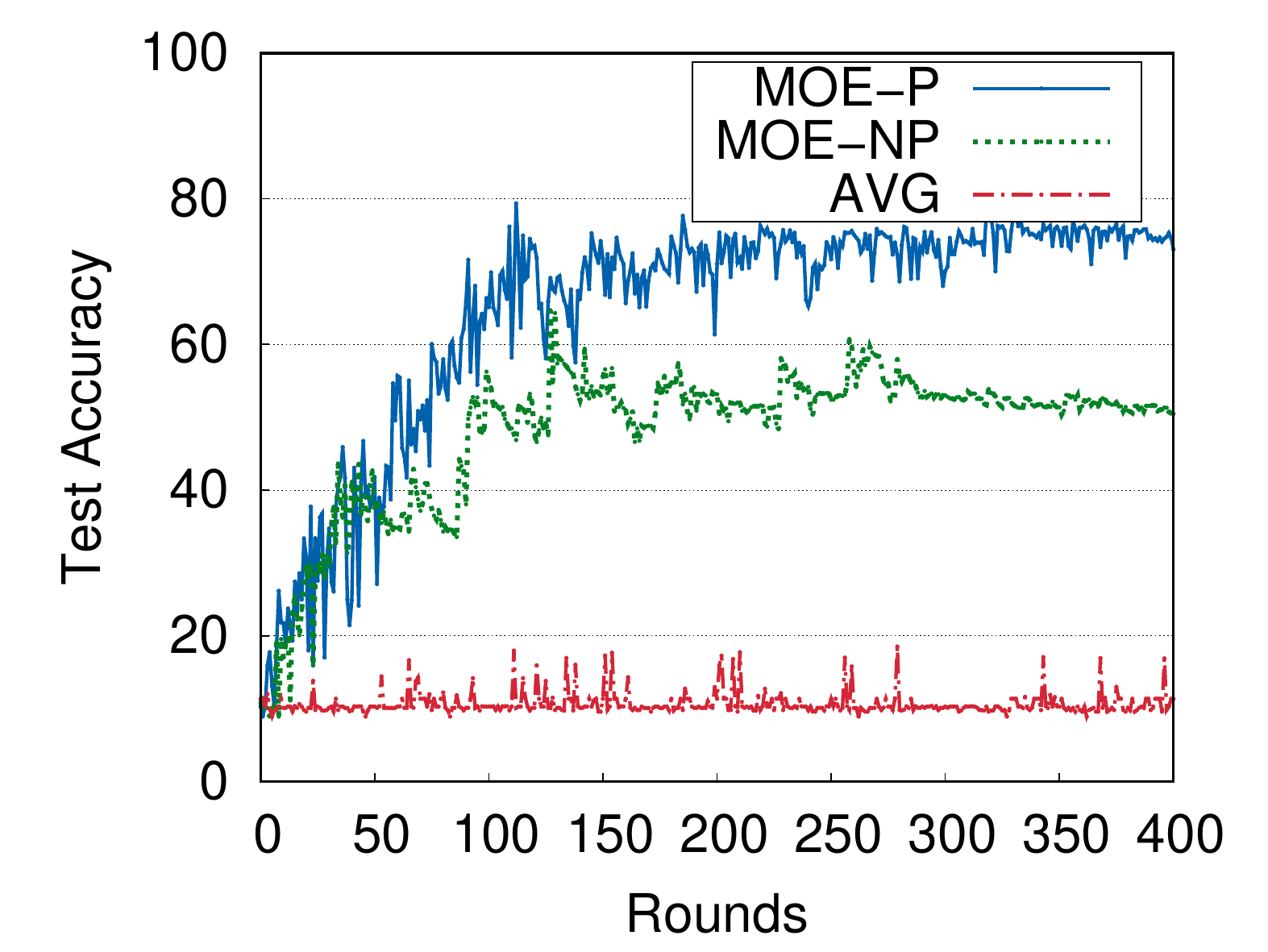}
  \caption{50 attackers}
  \label{fig:niid_accuracy_50}
\end{subfigure}
\begin{subfigure}{.32\textwidth}
  \centering
  \includegraphics[width=.98\linewidth]{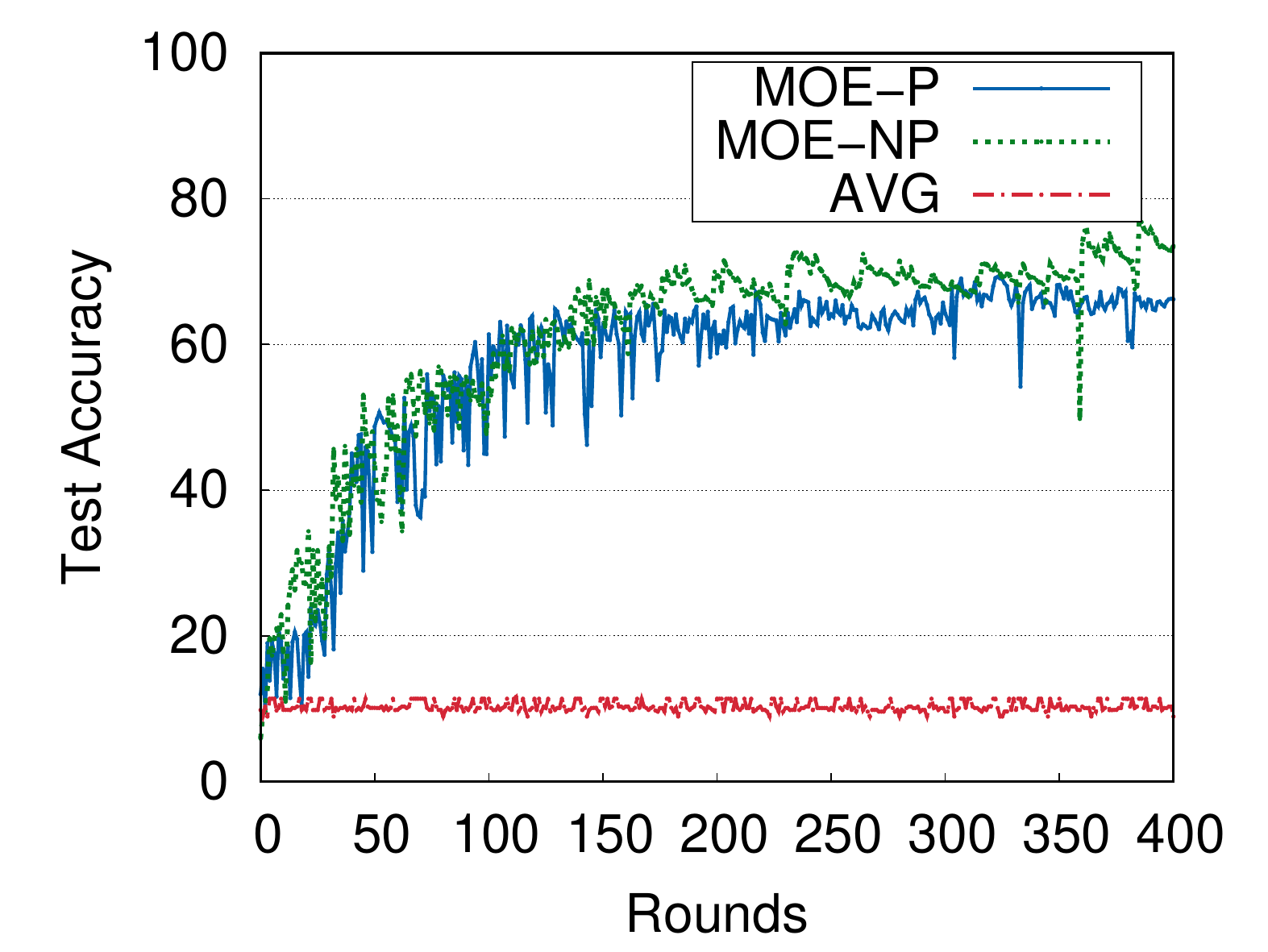}
  \caption{75 attackers}
  \label{fig:niid_accuracy_75}
\end{subfigure}

\begin{subfigure}{.32\textwidth}
  \centering
  \includegraphics[width=.98\linewidth]{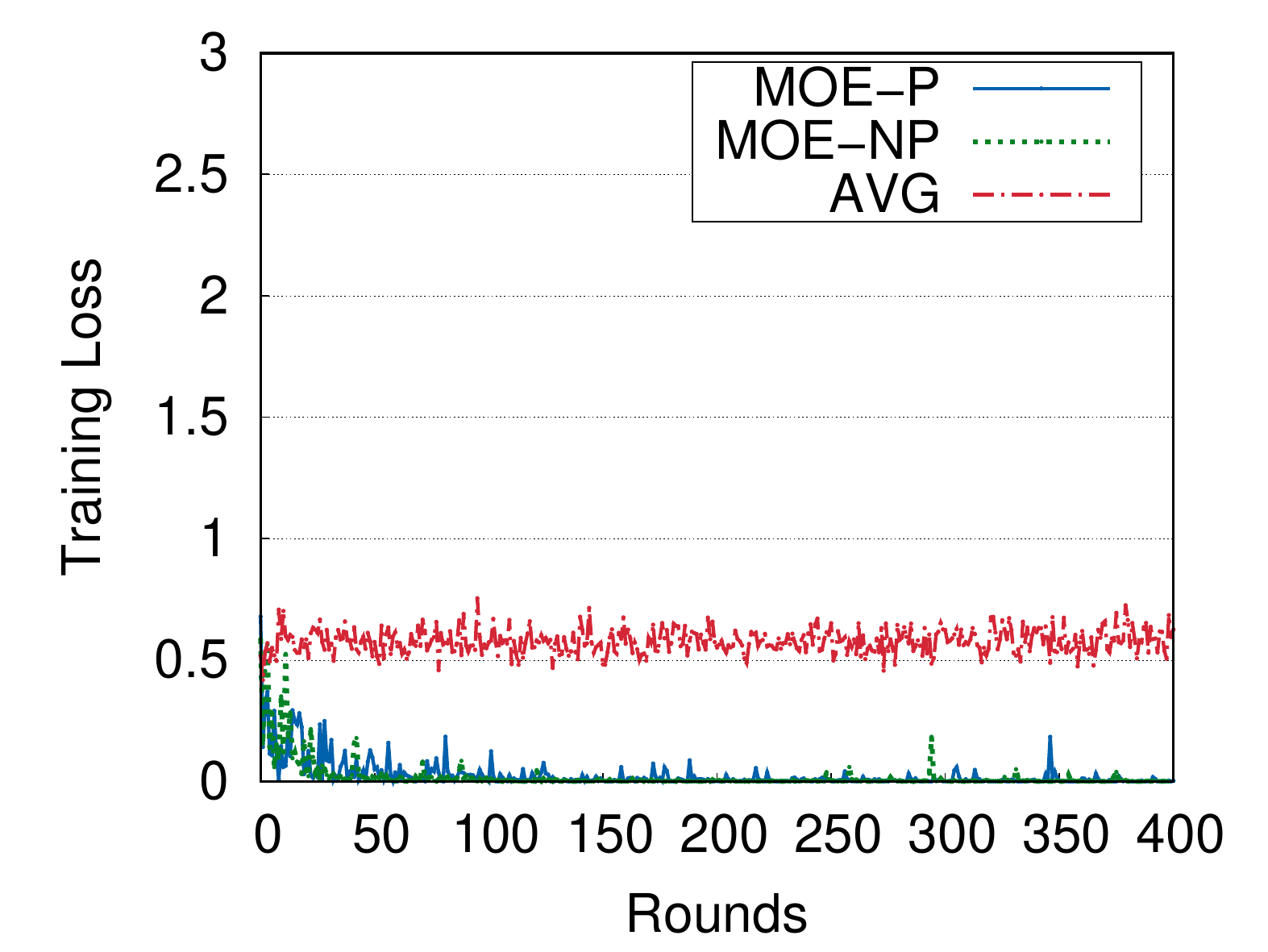}
  \caption{25 attackers}
  \label{fig:niid_loss_25}
\end{subfigure}%
\begin{subfigure}{.32\textwidth}
  \centering
  \includegraphics[width=.98\linewidth]{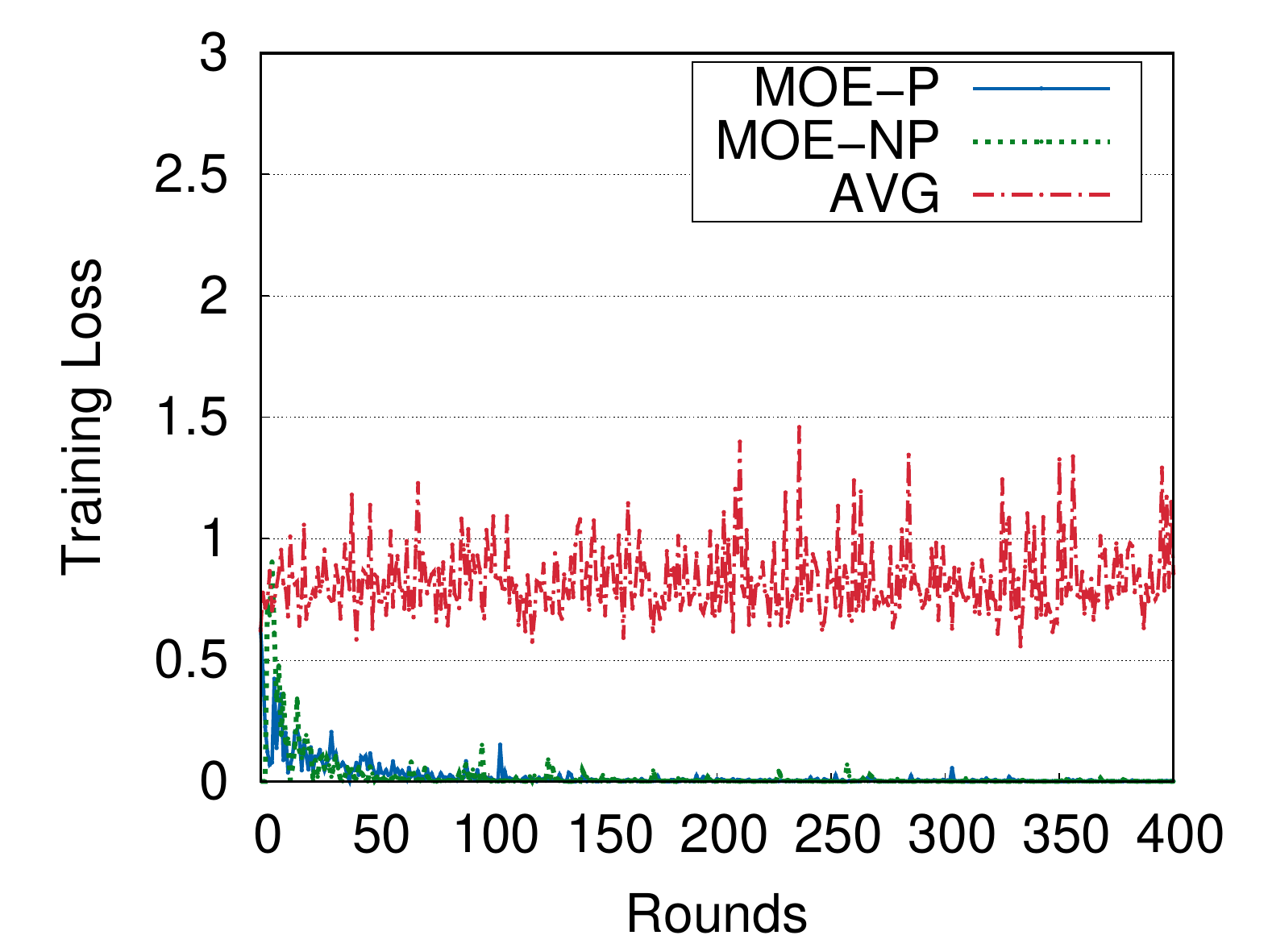}
  \caption{50 attackers}
  \label{fig:niid_loss_50}
\end{subfigure}
\begin{subfigure}{.32\textwidth}
  \centering
  \includegraphics[width=.98\linewidth]{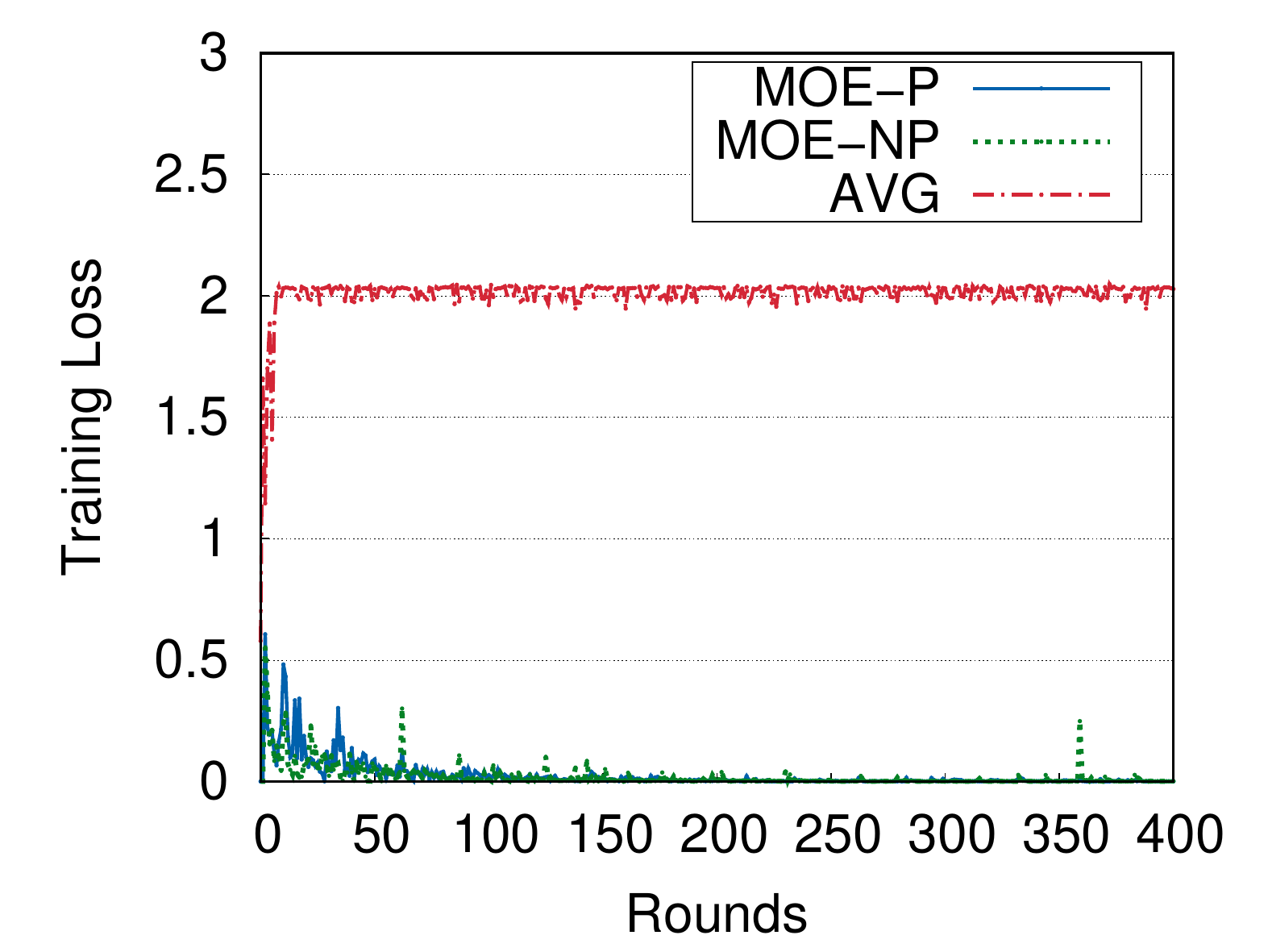}
  \caption{75 attackers}
  \label{fig:niid_loss_75}
\end{subfigure}
\end{center}
\caption{Accuracy and train Loss versus number of rounds and number of attackers for Non-IID data sets}
\label{fig:niid_accuracy_loss}
\end{figure}

In this case, all the parameters are similar to above examples excepts that the data sets are non-IIDs. Fig. \ref{fig:niid_accuracy_loss} highlights that the performance of FedAvg and demonstrates poor results even for the lower number of attackers in FL, e.g., Figs. \ref{fig:niid_accuracy_loss} (a) and (d). However, MoE-FL improves its performance gradually while the simulation progresses Figs. \ref{fig:niid_accuracy_loss} (b) and (e). As expected from Fig. \ref{fig:niid_accuracy_base:niid}, the accuracy does not surpass 80\%. However, hen $\mathcal{D}_0$ is pure, the MoE-FL reaches the same level of accuracy. In general for the non-IID data sets, MoE-FL is more sensitive to the poisoned samples in $\mathcal{D}_0$, e.g., accuracy of MOE-NP in Fig. \ref{fig:niid_accuracy_loss} (b) is 25\% less than that of MoE-P. For $E\geq 75$, the accuracy of both MoE-P and MoE-NP cannot attain to 80\%. We still can see the spike in training loss in non-IID setting for MoE-NP compared to the MoE-P which highlights that MoE-NP is more sensitive to the number of attackers compared to the MoE-P. In general, larger $E$ degrades both MOE-P and MOE-NP.

\subsection{The Effects of Various Types of Attack}\label{effect_of_attacks}
\begin{figure*}
\begin{center}
\begin{subfigure}{.38\textwidth}
  \centering
  \includegraphics[width=.98\linewidth]{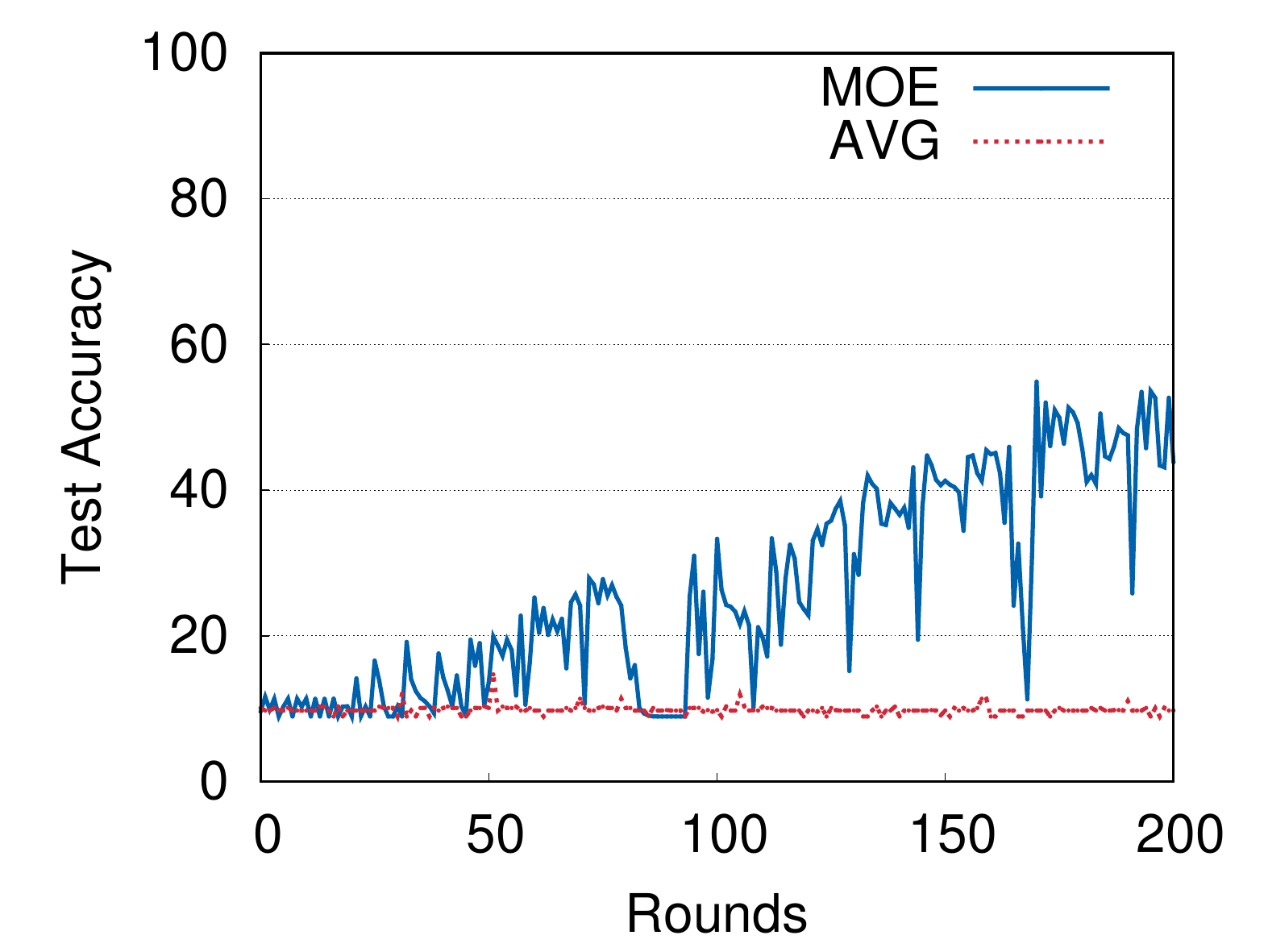}
  \caption{Random Weight Attack}
  \label{fig:niid_accuracy_atk1}
\end{subfigure}
\begin{subfigure}{.38\textwidth}
  \centering
  \includegraphics[width=0.98\textwidth]{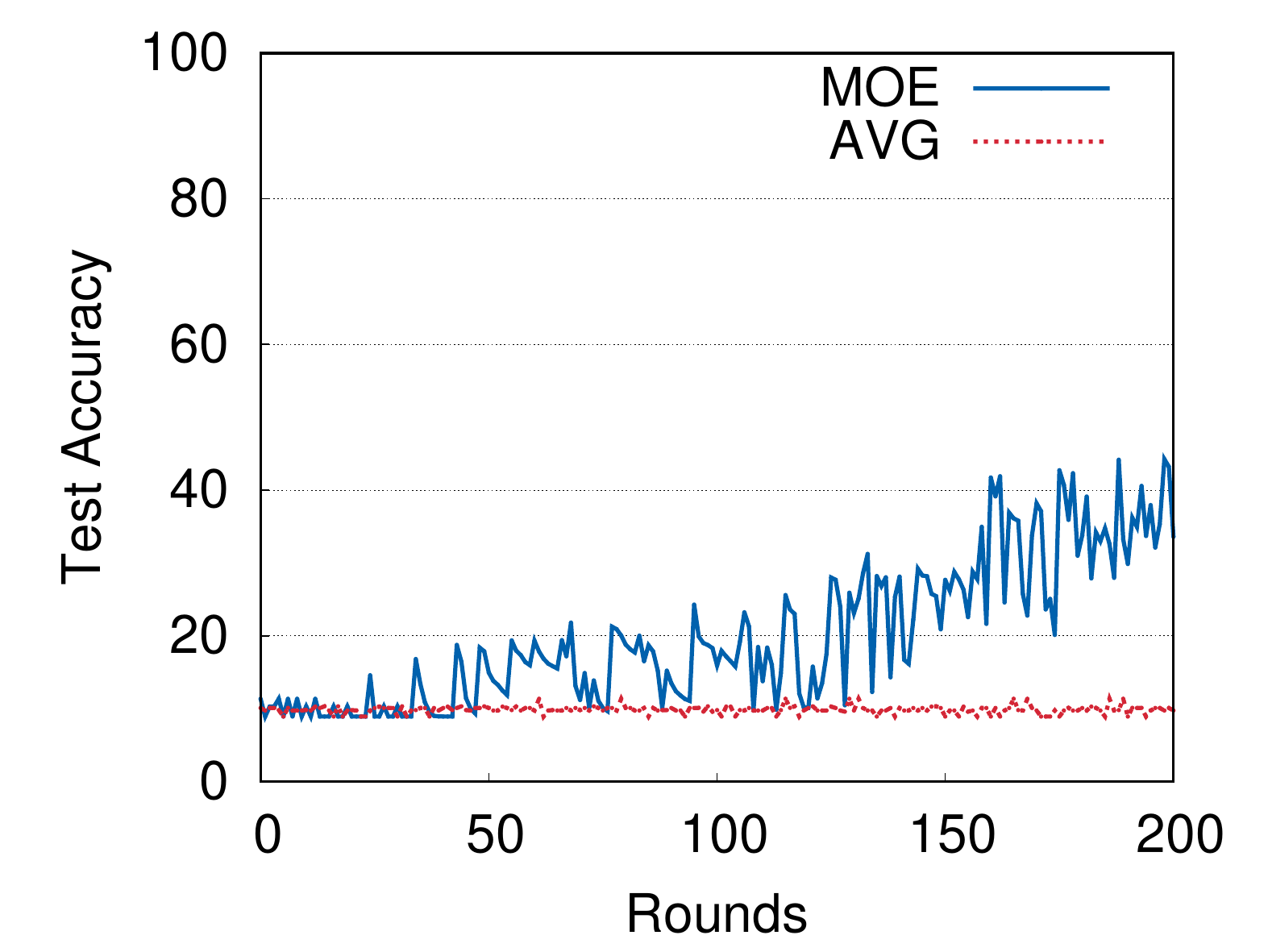}
  \caption{Negative Weight Attack}
  \label{fig:niid_accuracy_atk2}
\end{subfigure}
\end{center}
\caption{Accuracy versus number of rounds for Non-IID data sets for random weight attack and negative weight attack }
\label{fig:niid_accuracy_attacks}
\end{figure*}

In this section, we investigate the effects of two different of attack to show how $\widetilde{\mathcal{D}}_n$ can affect the performance of MoE-FL and Fed-Avg. First, the random weight attacks where each attacker adds noise to their derived models which directly simulates the scenario that there is a large noise over $\widetilde{\mathcal{D}}_n$. Second, the negative weight attacks, where each attacker sends the negative values of the server model parameters which can mimic the scenario that $\widetilde{\mathcal{D}}_n$ of $n \in \mathcal{E}$ is changed according to $\mathcal{D}_n$ of $n \in \mathcal{N}/\mathcal{E}$. For both scenarios, the data sets in the server is pure. Results in Figs. \ref{fig:niid_accuracy_attacks} (a) and (b) for 50 attackers demonstrate that the negative weight attack is more harmful that the random weight attack for MoE-FL, e.g., accuracy of 55\% versus 40\% in 200 rounds. It can be concluded that if the attackers are smart enough to perform an attack that minimize the distance of their model parameters with the server model, then MoE-FL performance degrades considerably. Still the MoE-FL can maintain the performance and recover from the attack after $150$ rounds for both cases while for the negative weight attack, the convergence rate is slower. FedAvg for both cases is not robust and has an unacceptable performance e.g., 10\% accuracy.

\subsection{Illustrative Examples} 
\begin{figure}[ht]
\begin{center}
\begin{subfigure}{.49\textwidth}
  \centering
  \includegraphics[width=.49\linewidth]{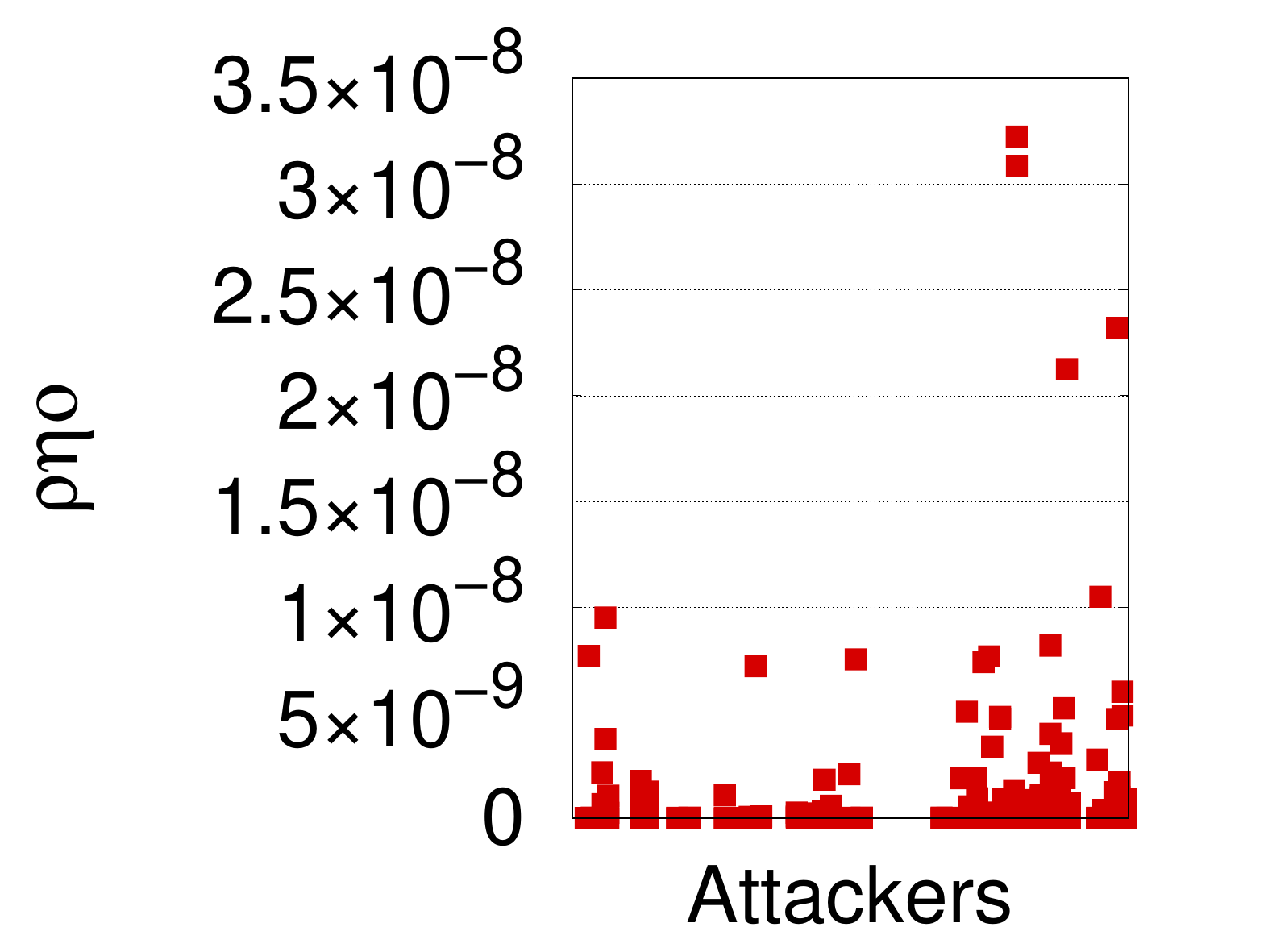}
  \includegraphics[width=.49\linewidth]{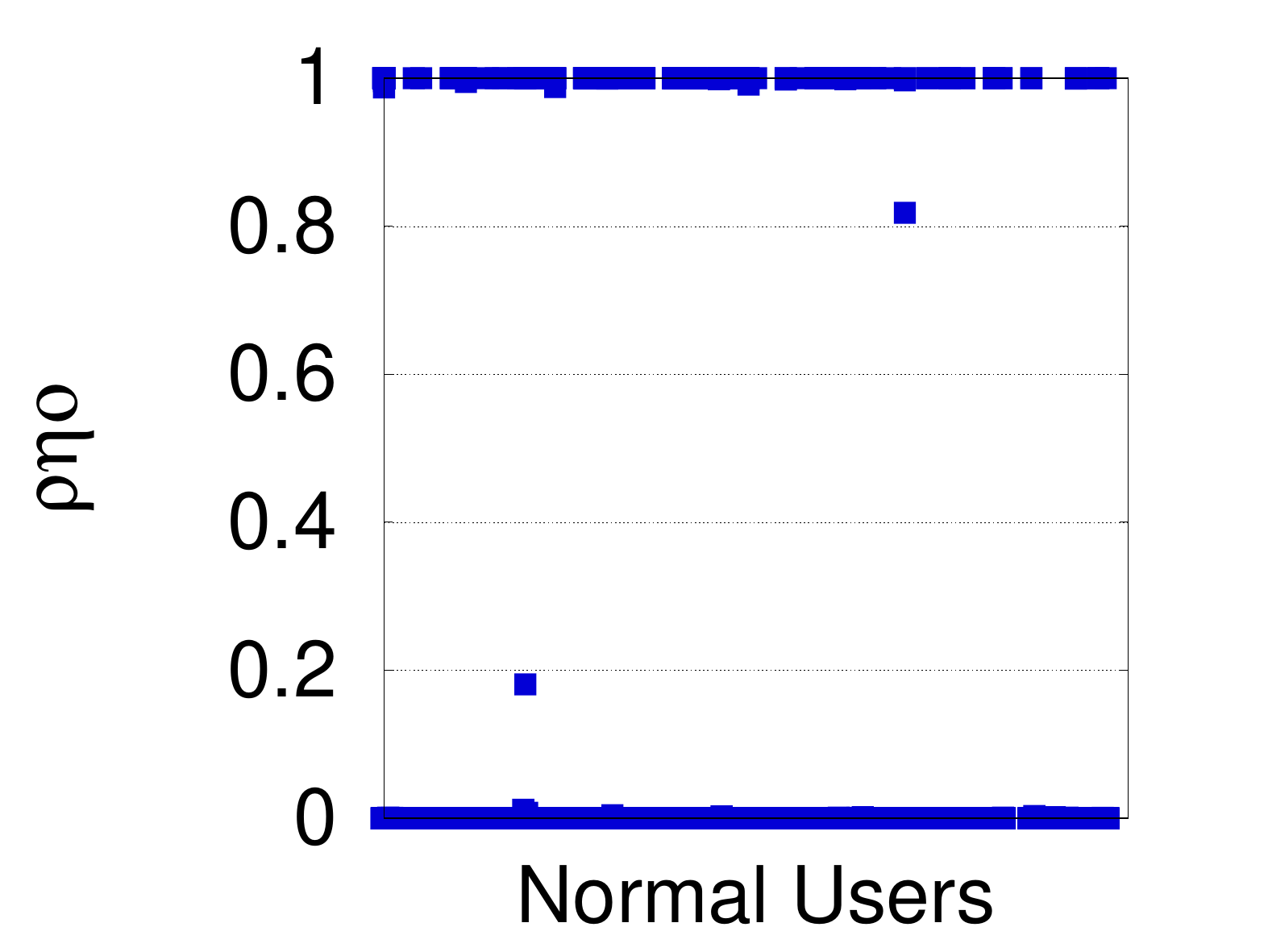}
  \caption{25 attackers}
  \label{fig:iid_weight_dist_25}
\end{subfigure}%
\begin{subfigure}{.49\textwidth}
  \centering
  \includegraphics[width=.49\linewidth]{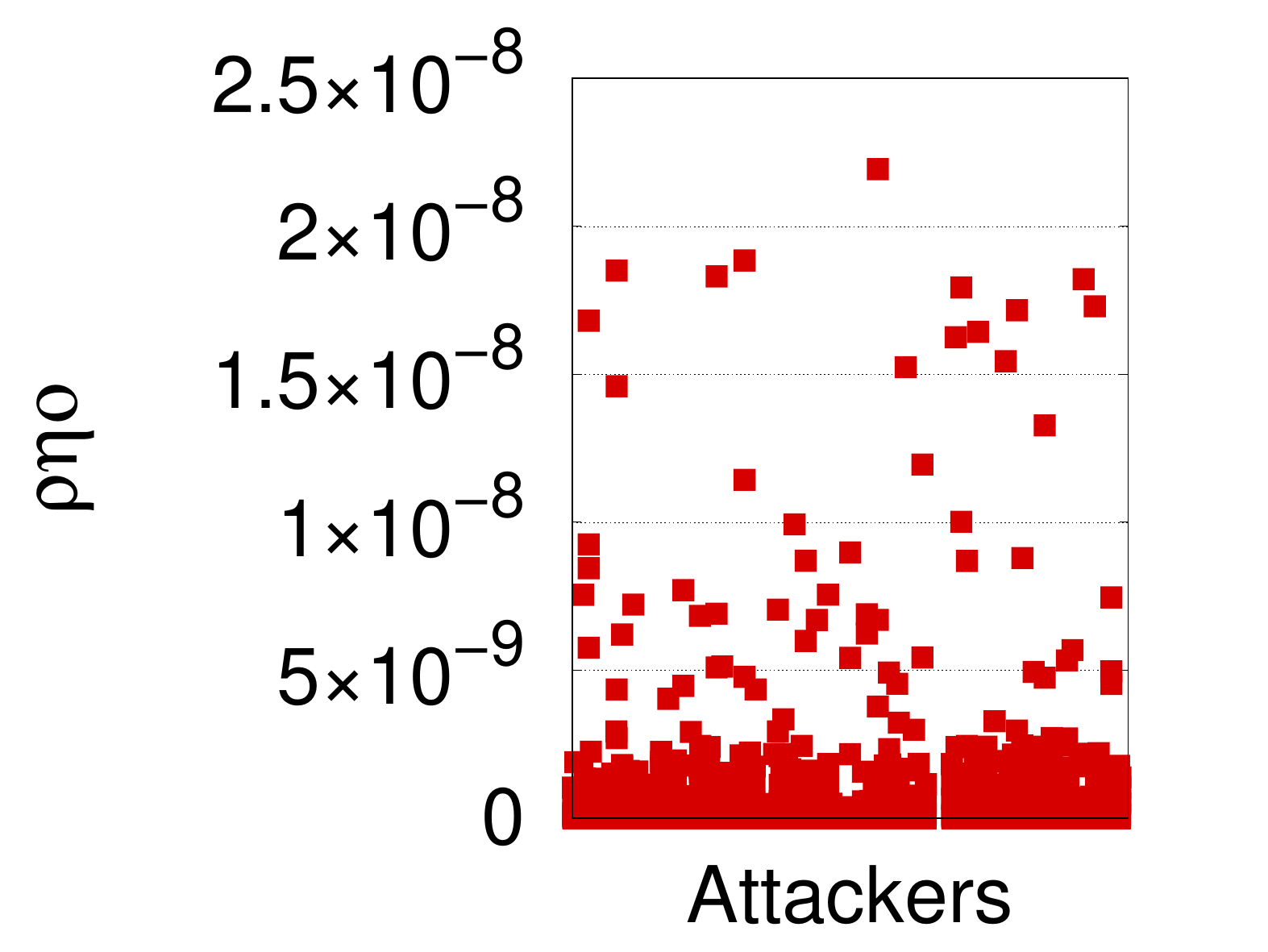}
  \includegraphics[width=.49\linewidth]{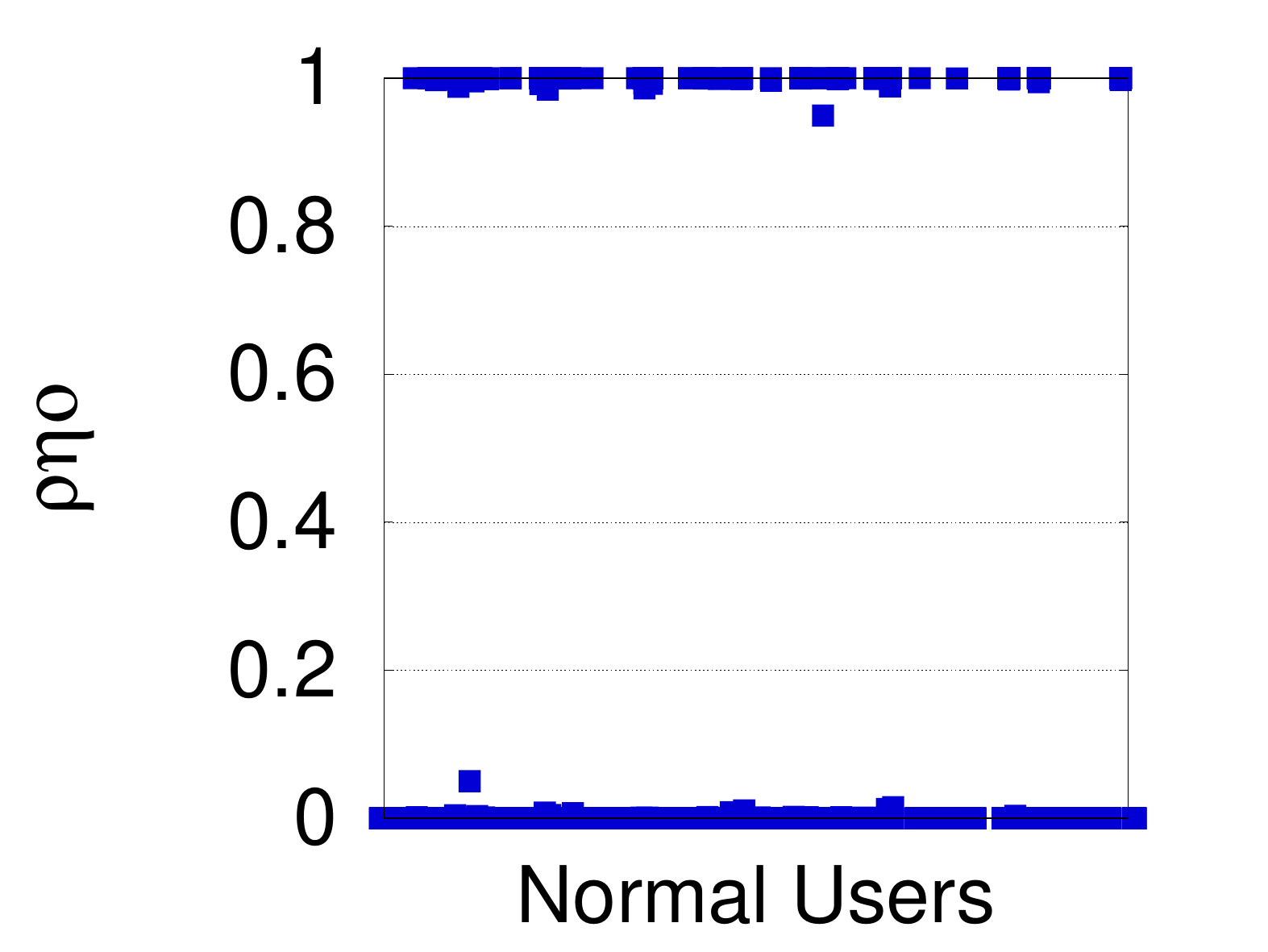}
  \caption{50 attackers}
  \label{fig:iid_weight_dist_50}
\end{subfigure}\\
\begin{subfigure}{.49\textwidth}
  \centering
  \includegraphics[width=.49\linewidth]{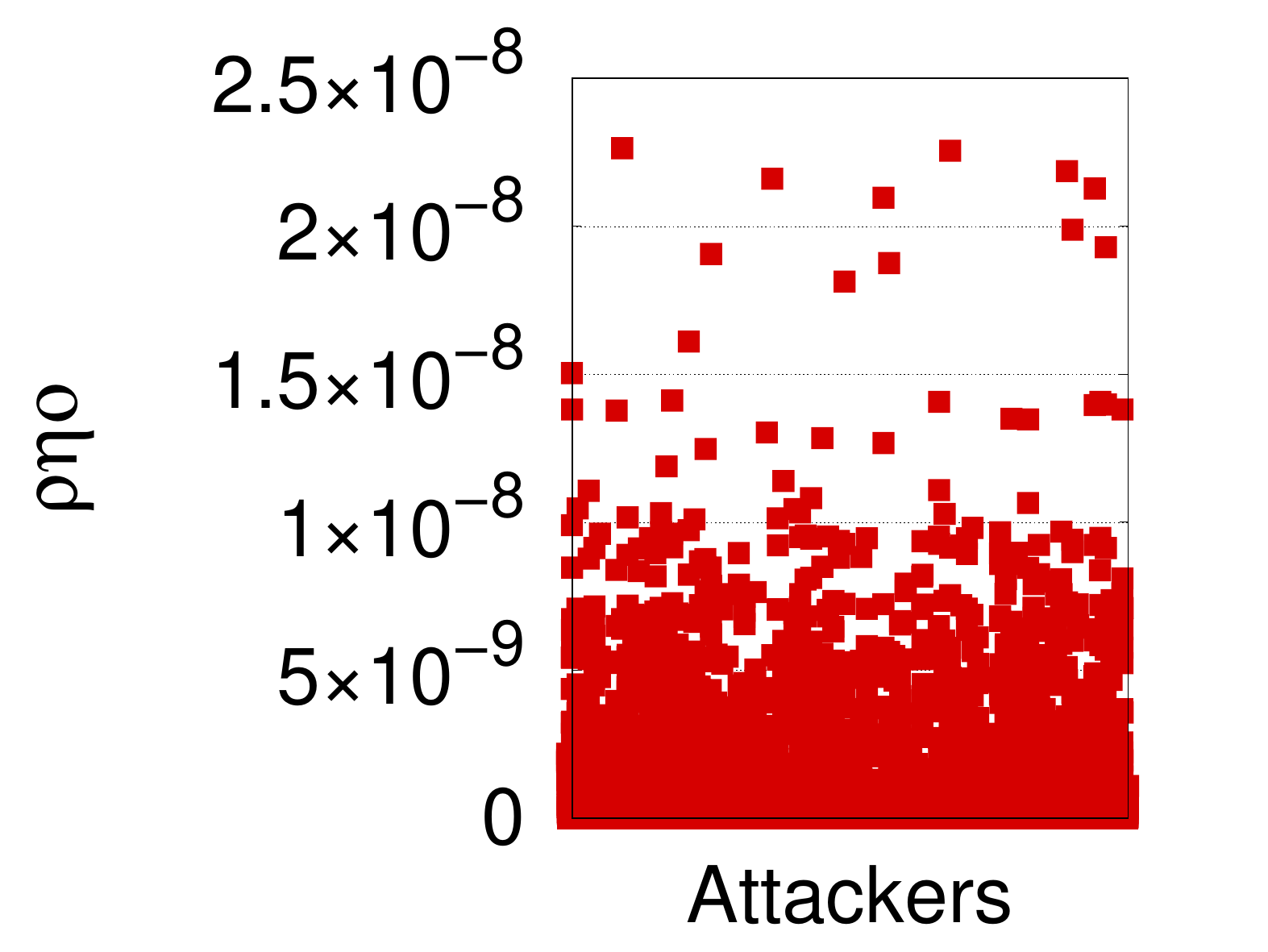}
  \includegraphics[width=.49\linewidth]{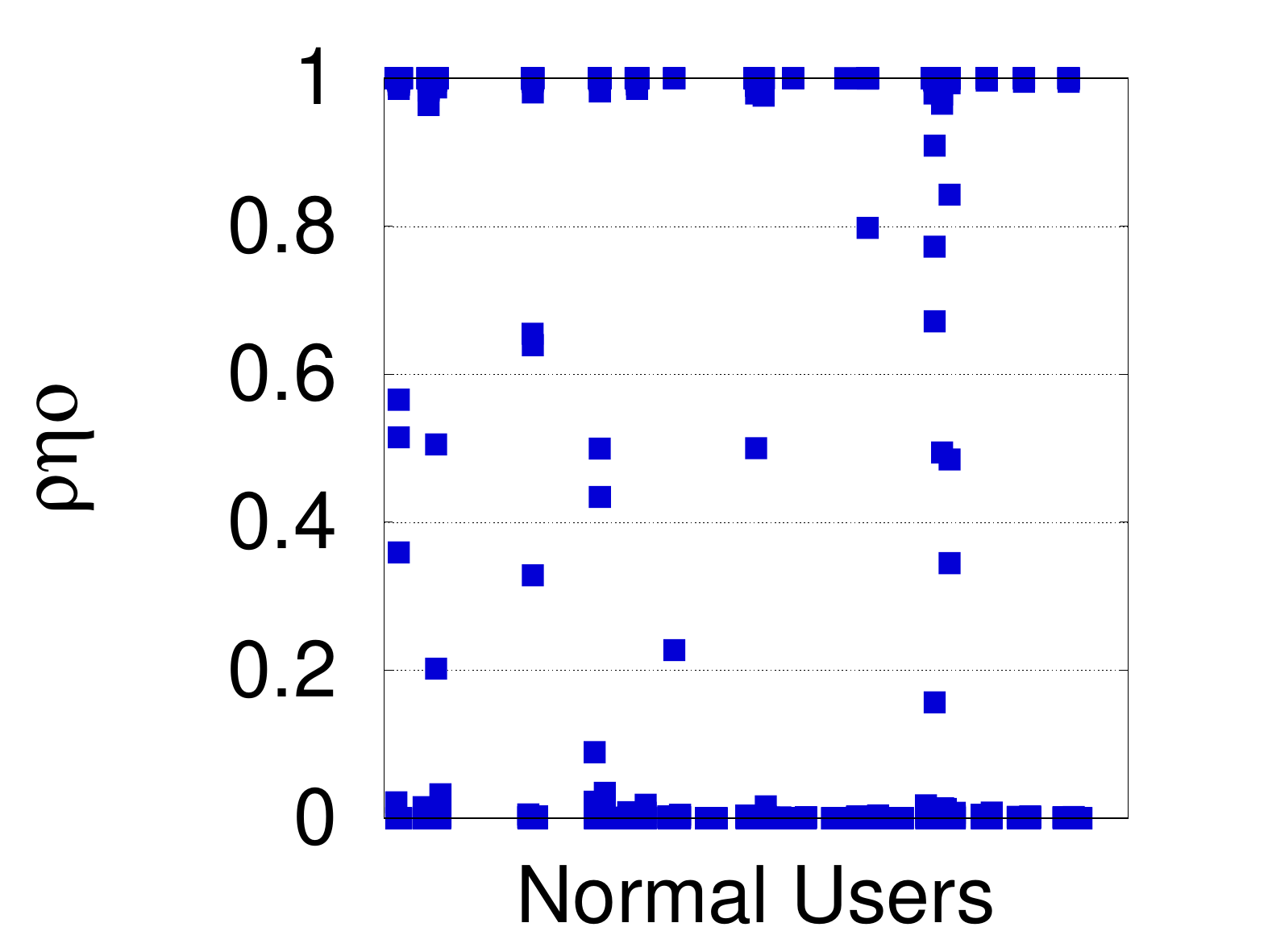}
  \caption{75 attackers}
  \label{fig:iid_weight_dist_75}
\end{subfigure}
\end{center}
\caption{Weight Distribution for IID data set with pure data in server (MoE-P)}
\label{fig:iid_weight_dist}
\end{figure}

\begin{figure} 
\begin{center}
\begin{subfigure}{.49\textwidth}
  \centering
  \includegraphics[width=.49\linewidth]{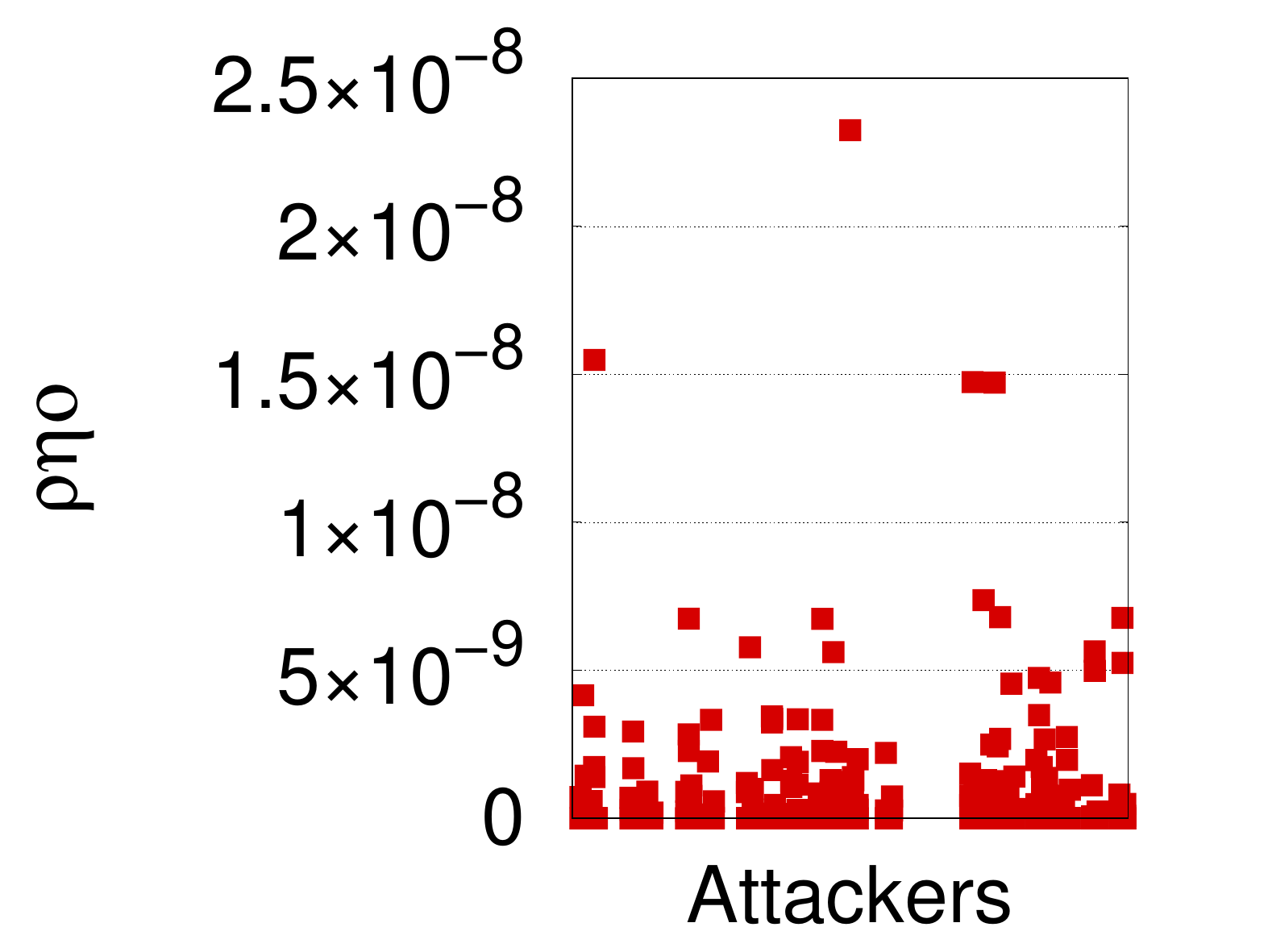}
  \includegraphics[width=.49\linewidth]{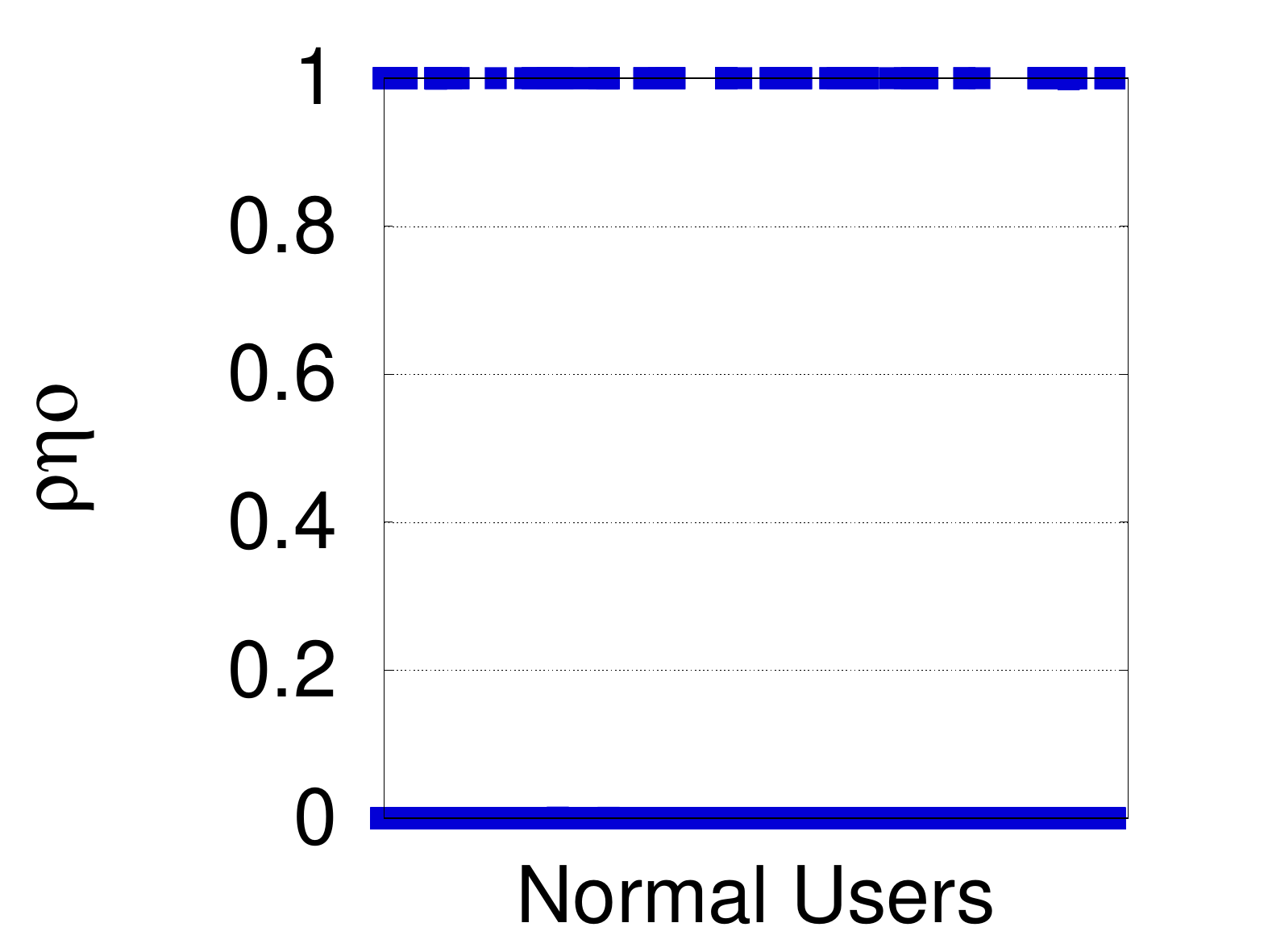}
  \caption{25 attackers}
  \label{fig:iid_np_weight_dist_25}
\end{subfigure}%
\begin{subfigure}{.49\textwidth}
  \centering
  \includegraphics[width=.49\linewidth]{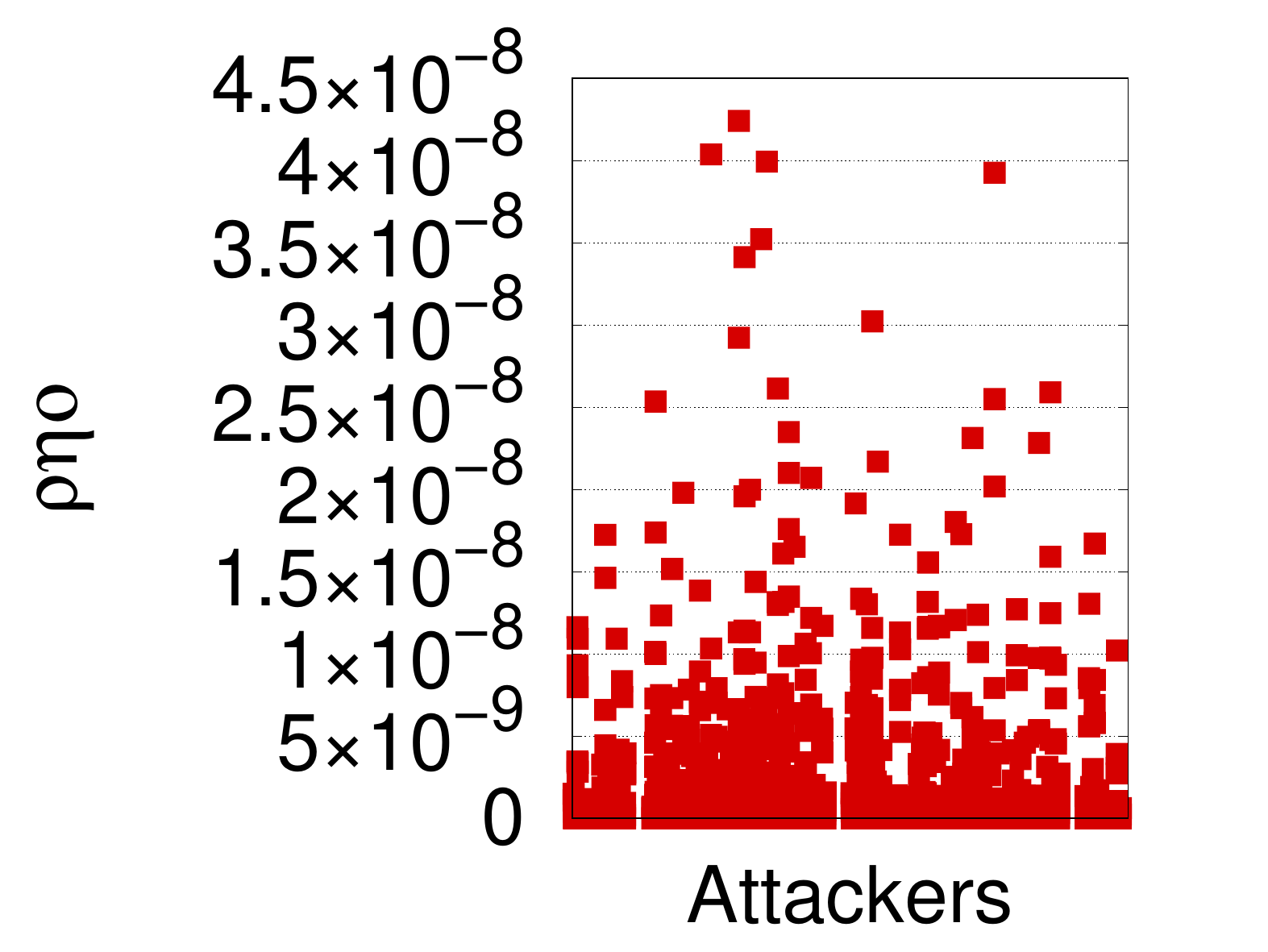}
  \includegraphics[width=.49\linewidth]{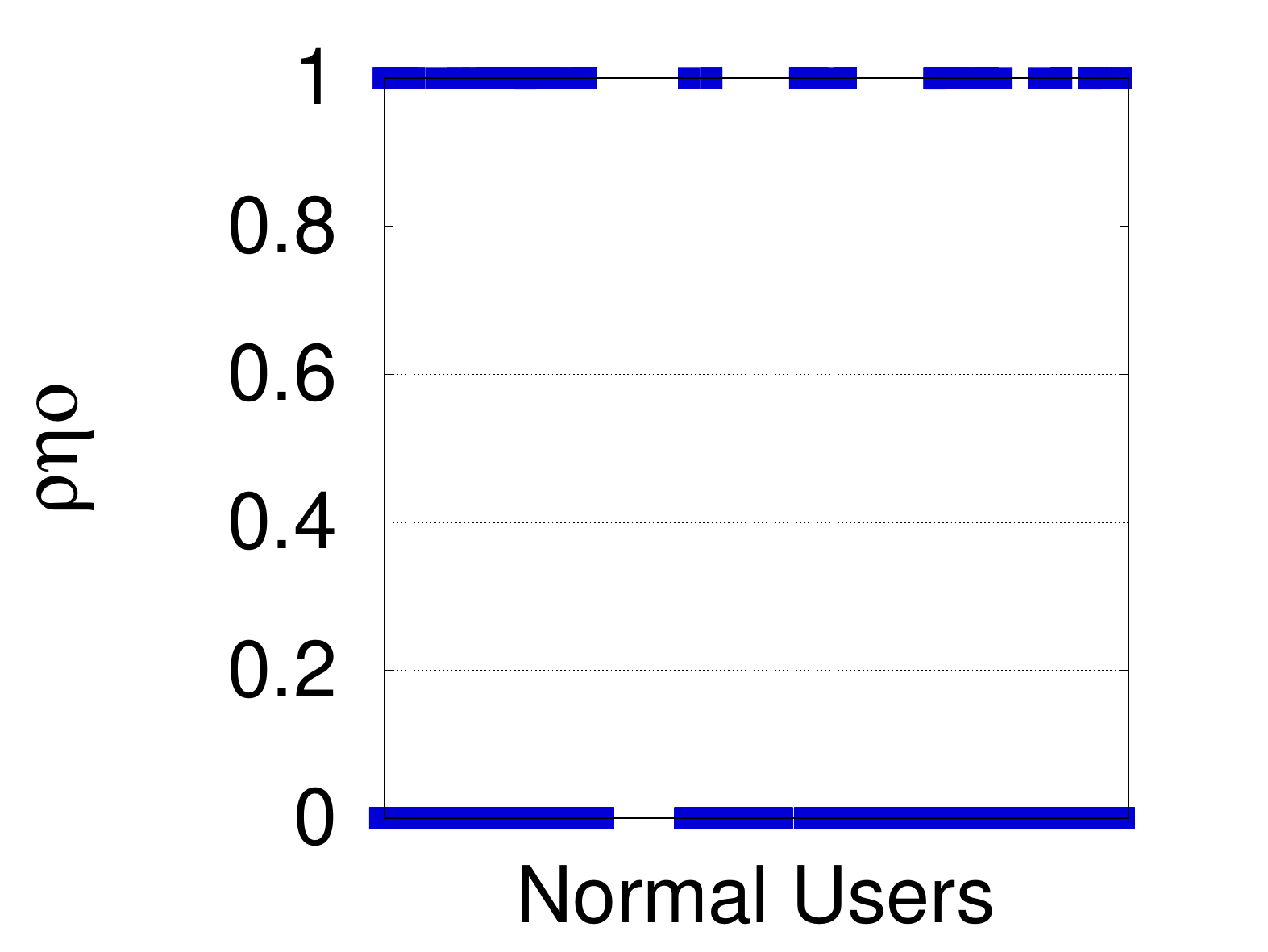}
  \caption{50 attackers}
  \label{fig:iid_np_weight_dist_50}
\end{subfigure}\\
\begin{subfigure}{.49\textwidth}
  \centering
  \includegraphics[width=.49\linewidth]{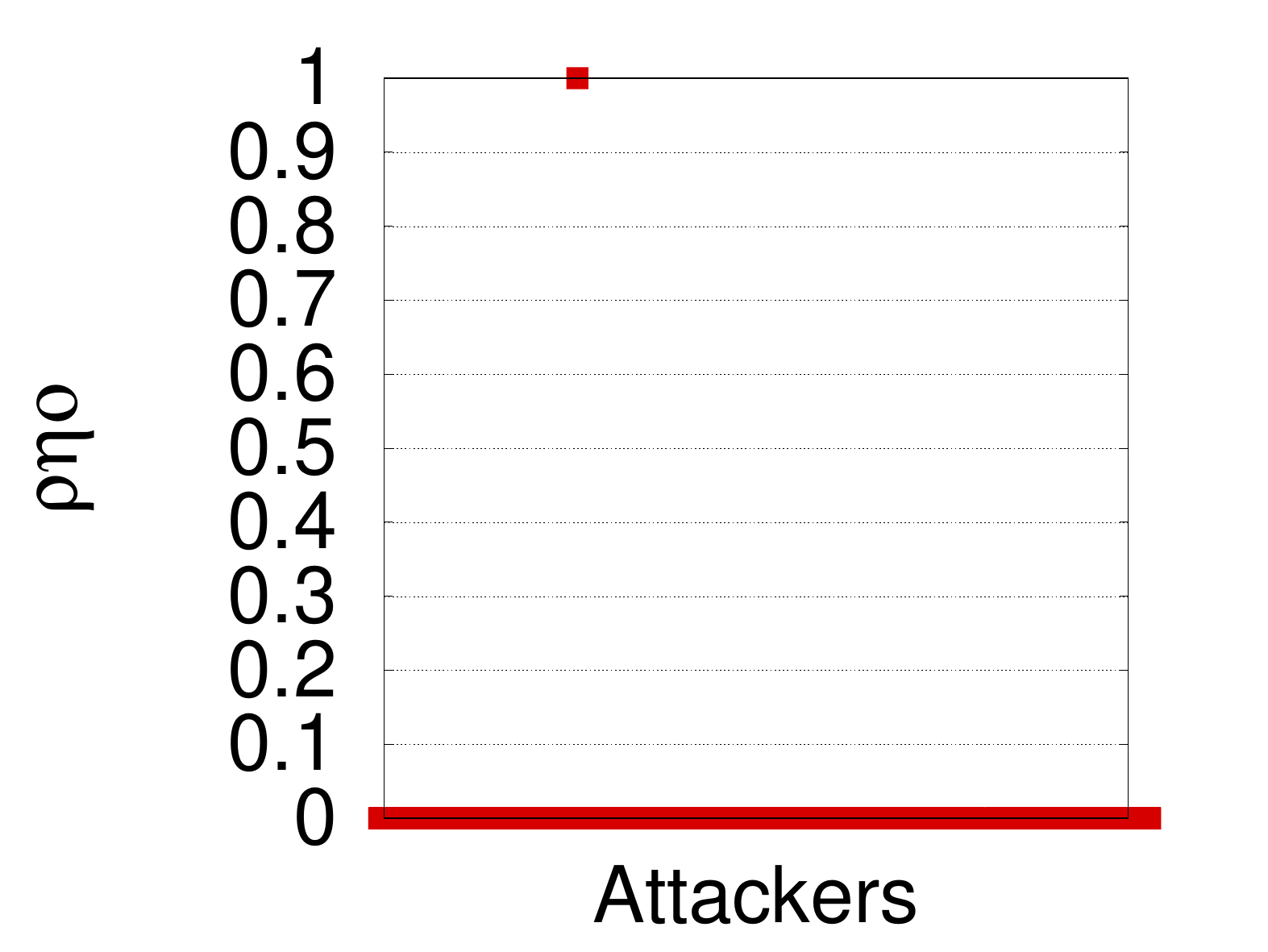}
  \includegraphics[width=.49\linewidth]{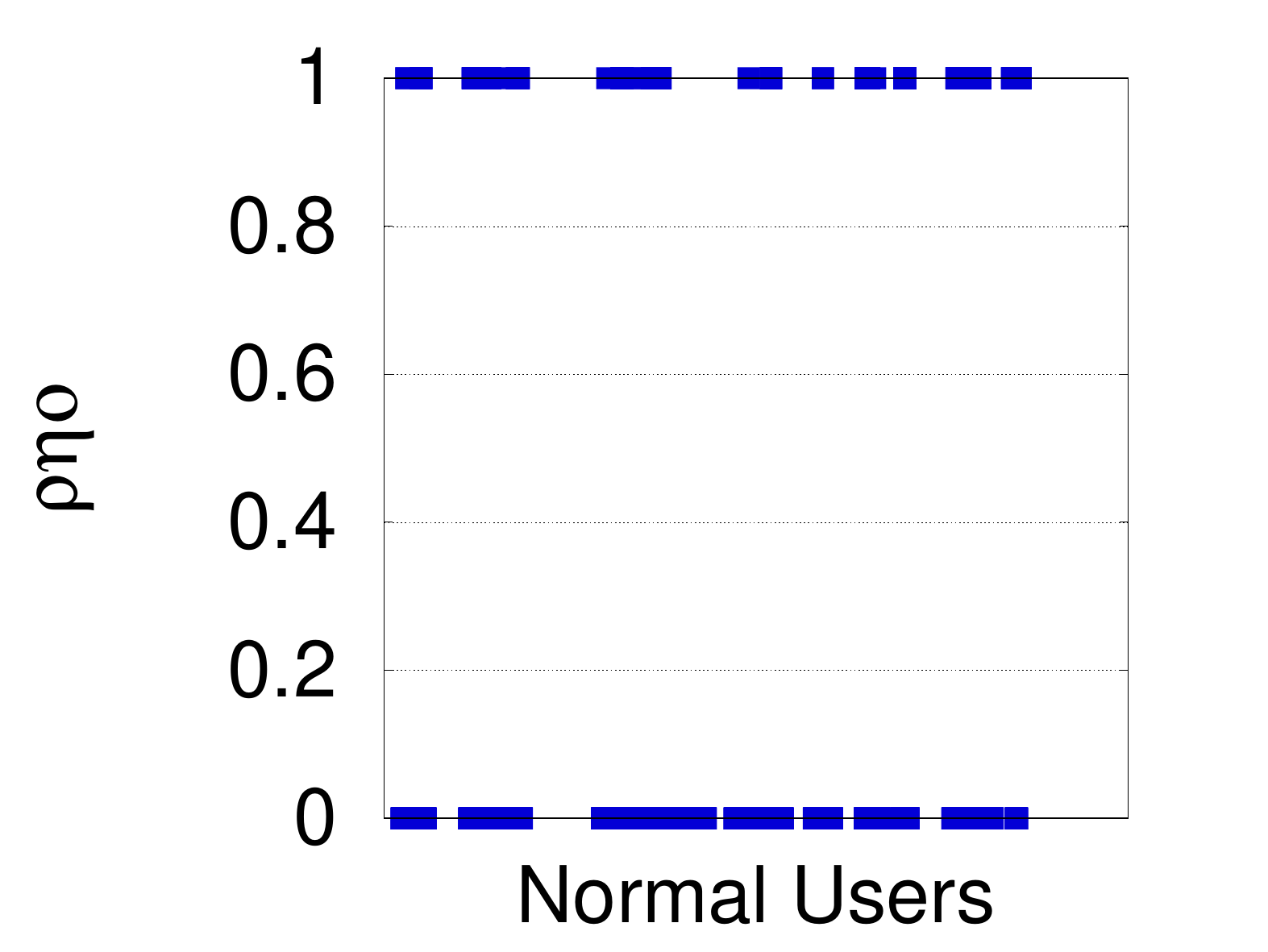}
  \caption{75 attackers}
  \label{fig:iid_np_weight_dist_75}
\end{subfigure}
\end{center}
\caption{Weight Distribution for IID data set with impure $\mathcal{D}_0$ (MoE-NP)}
\label{fig:iid_np_weight_dist}
\end{figure}

\begin{figure} 
\begin{center}
\begin{subfigure}{.49\textwidth}
  \centering
  \includegraphics[width=.49\linewidth]{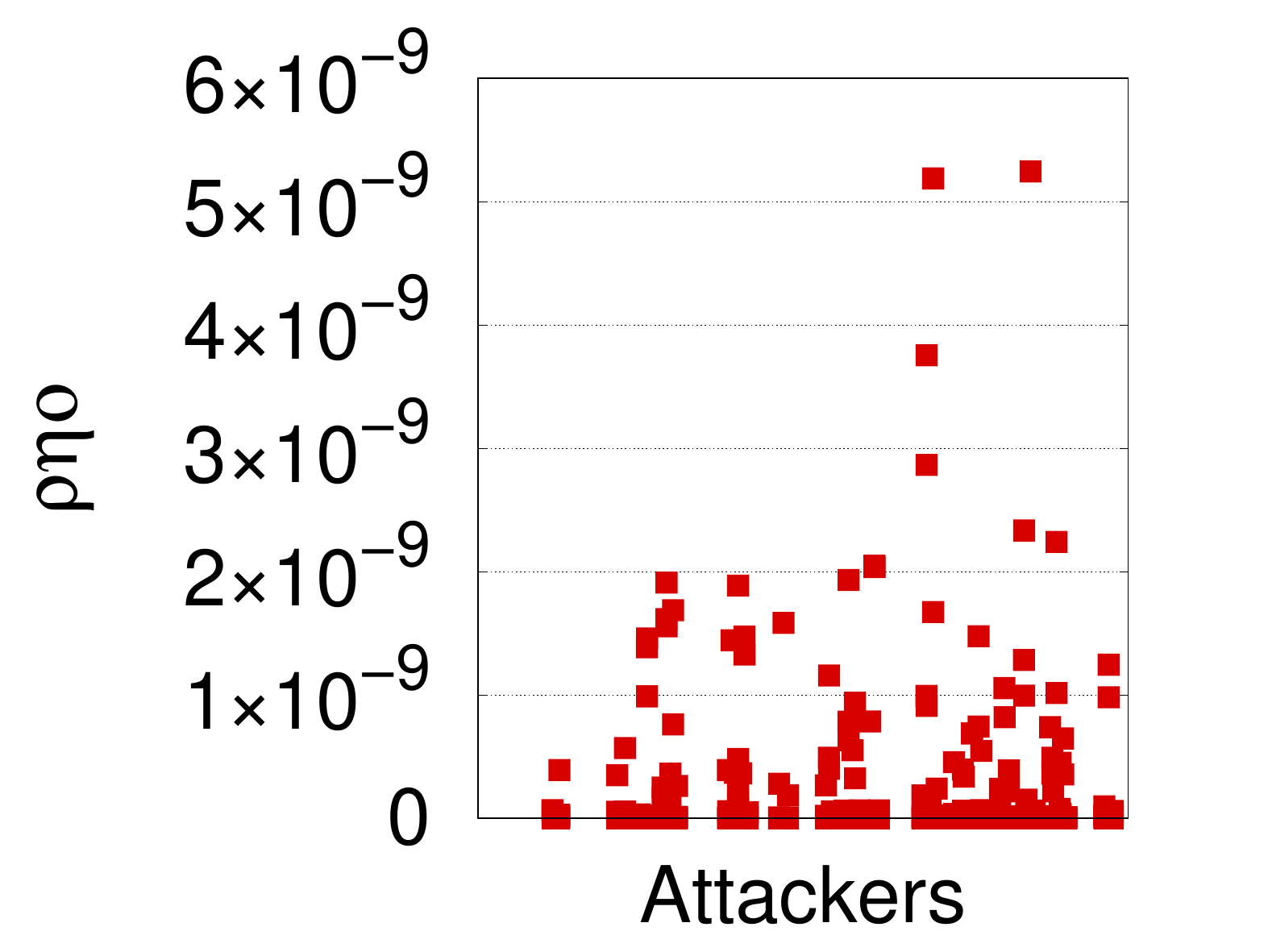}
  \includegraphics[width=.49\linewidth]{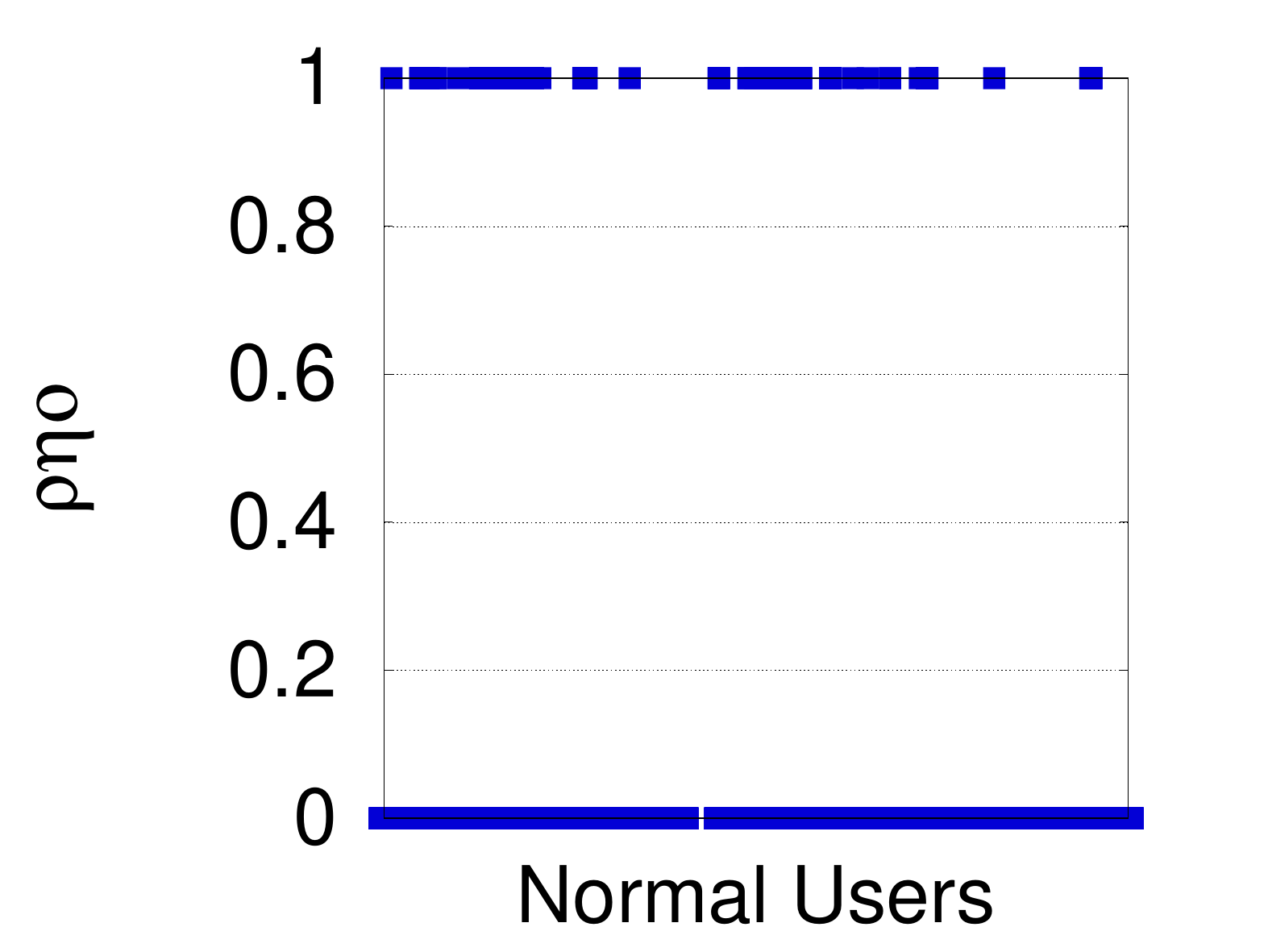}
  \caption{25 attackers}
  \label{fig:niid_weight_dist_25}
\end{subfigure}%
\begin{subfigure}{.49\textwidth}
  \centering
  \includegraphics[width=.49\linewidth]{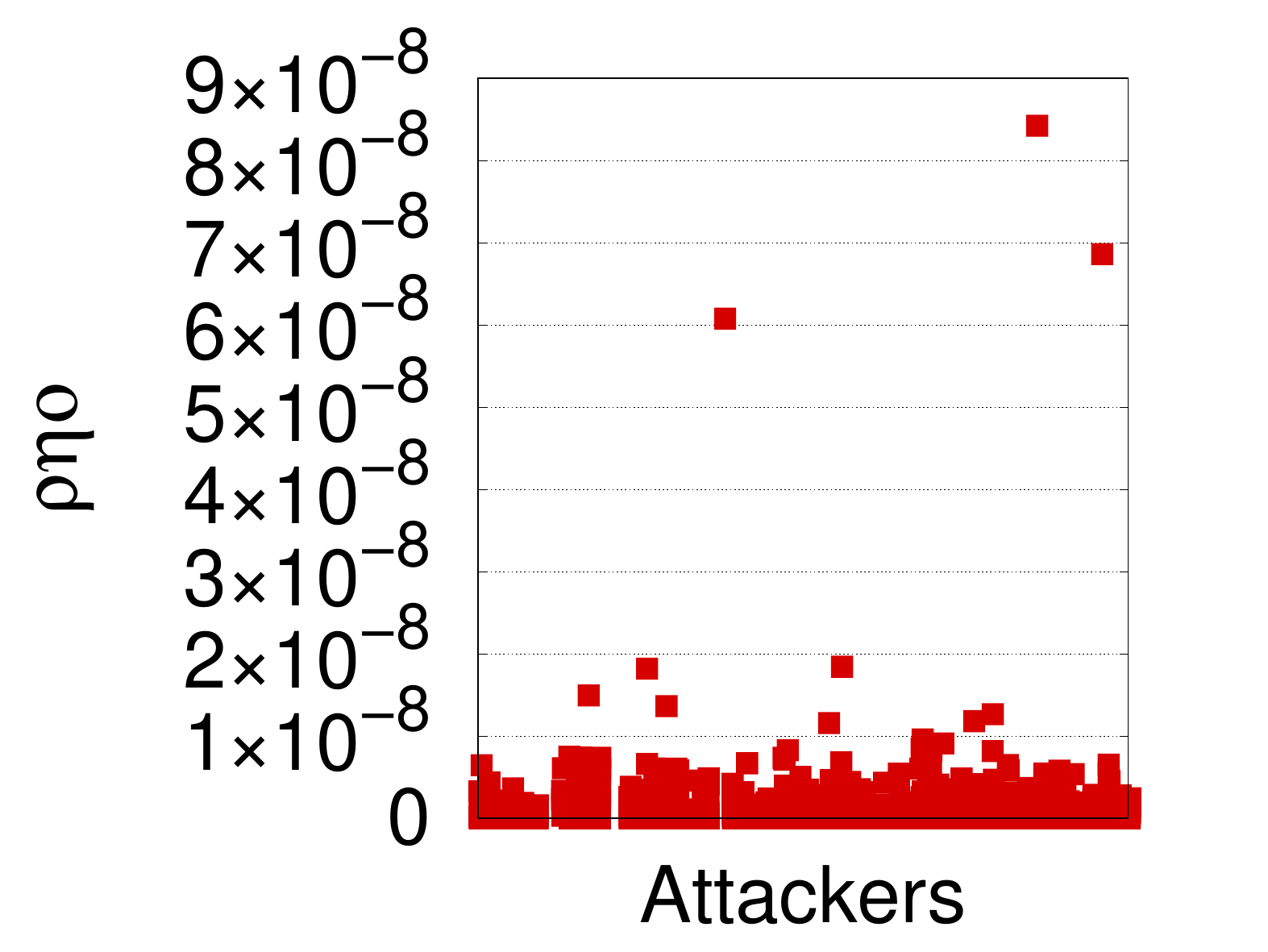}
  \includegraphics[width=.49\linewidth]{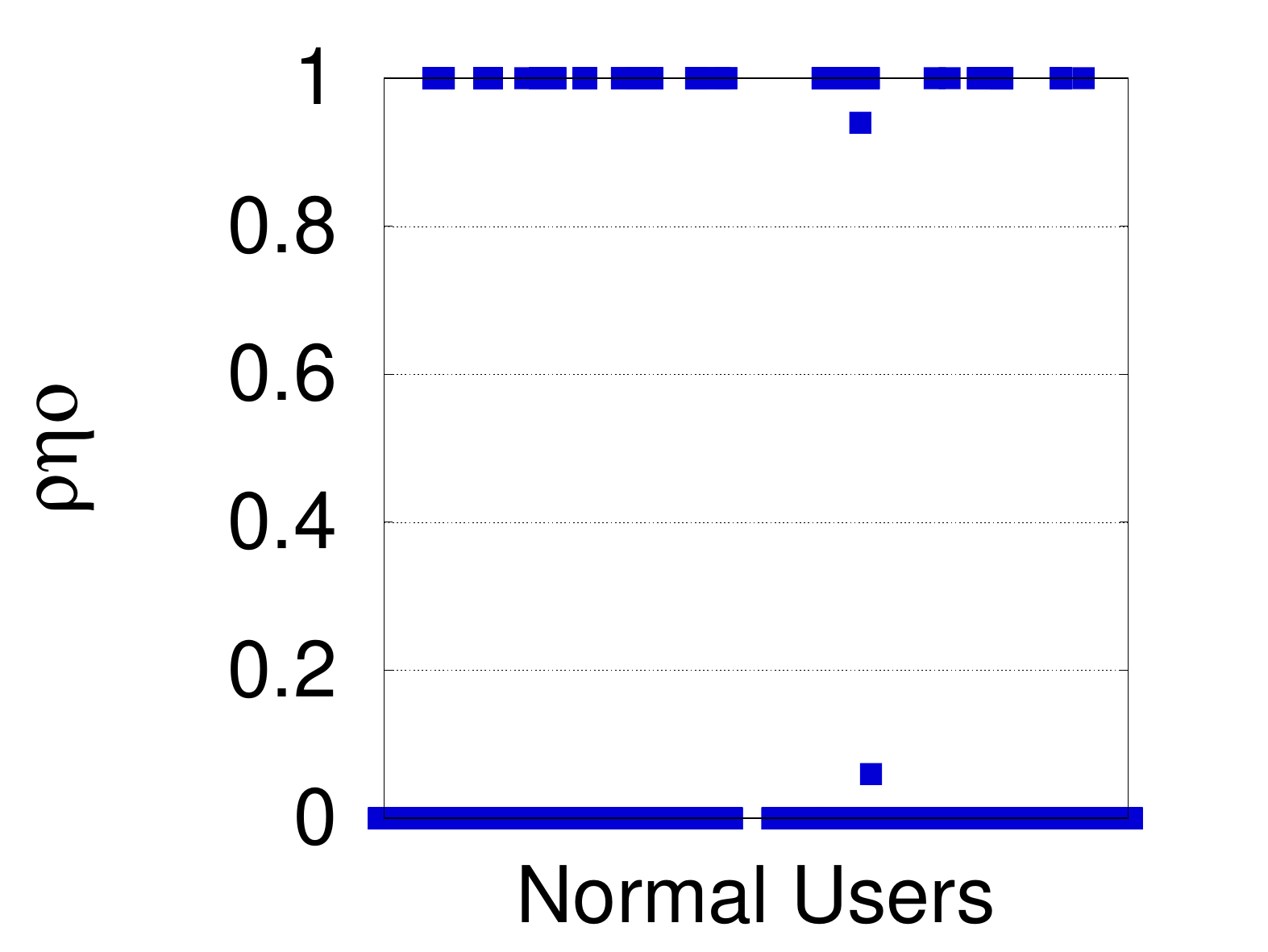}
  \caption{50 attackers}
  \label{fig:niid_weight_dist_50}
\end{subfigure}\\
\begin{subfigure}{.49\textwidth}
  \centering
  \includegraphics[width=.49\linewidth]{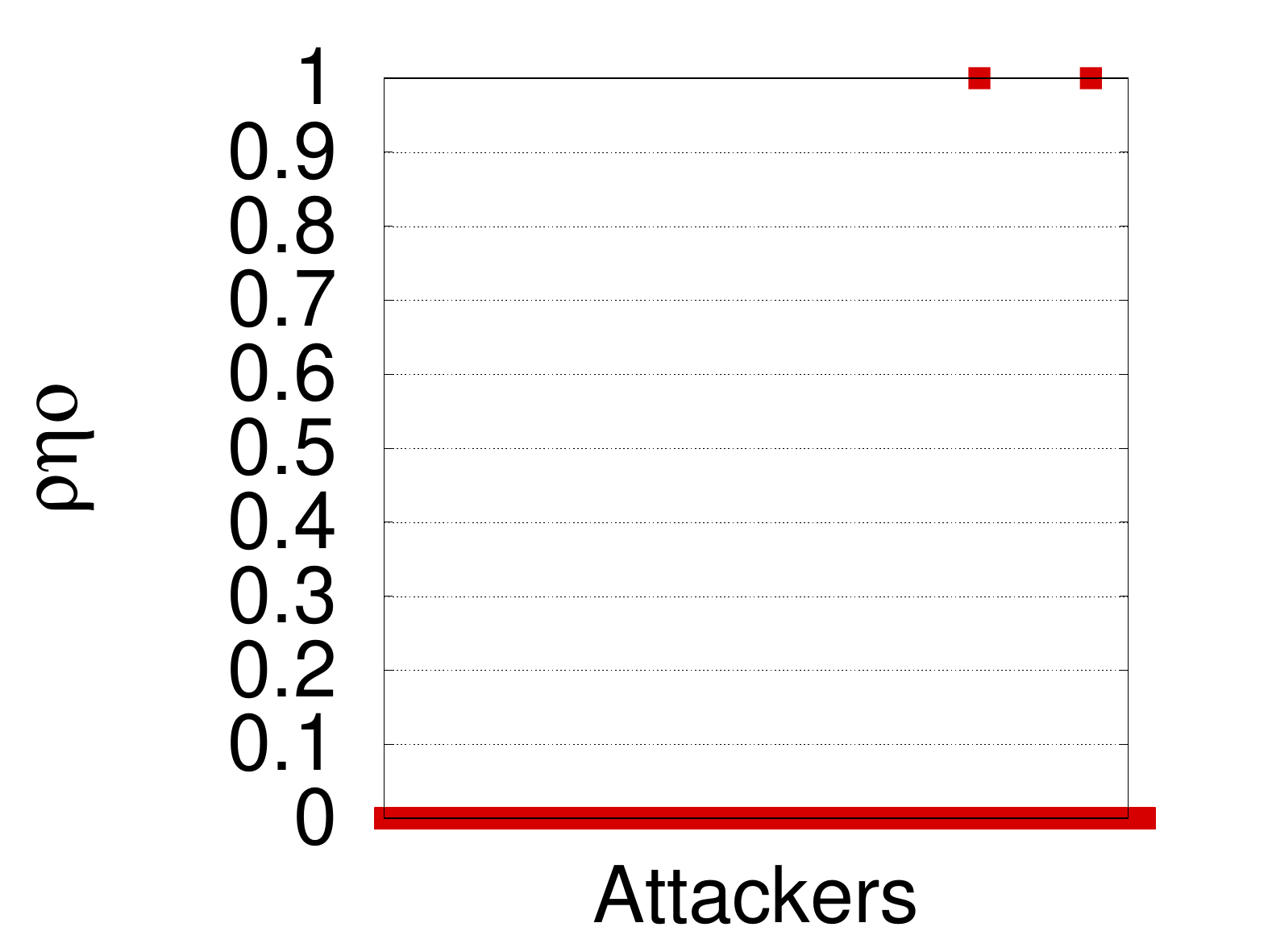}
  \includegraphics[width=.49\linewidth]{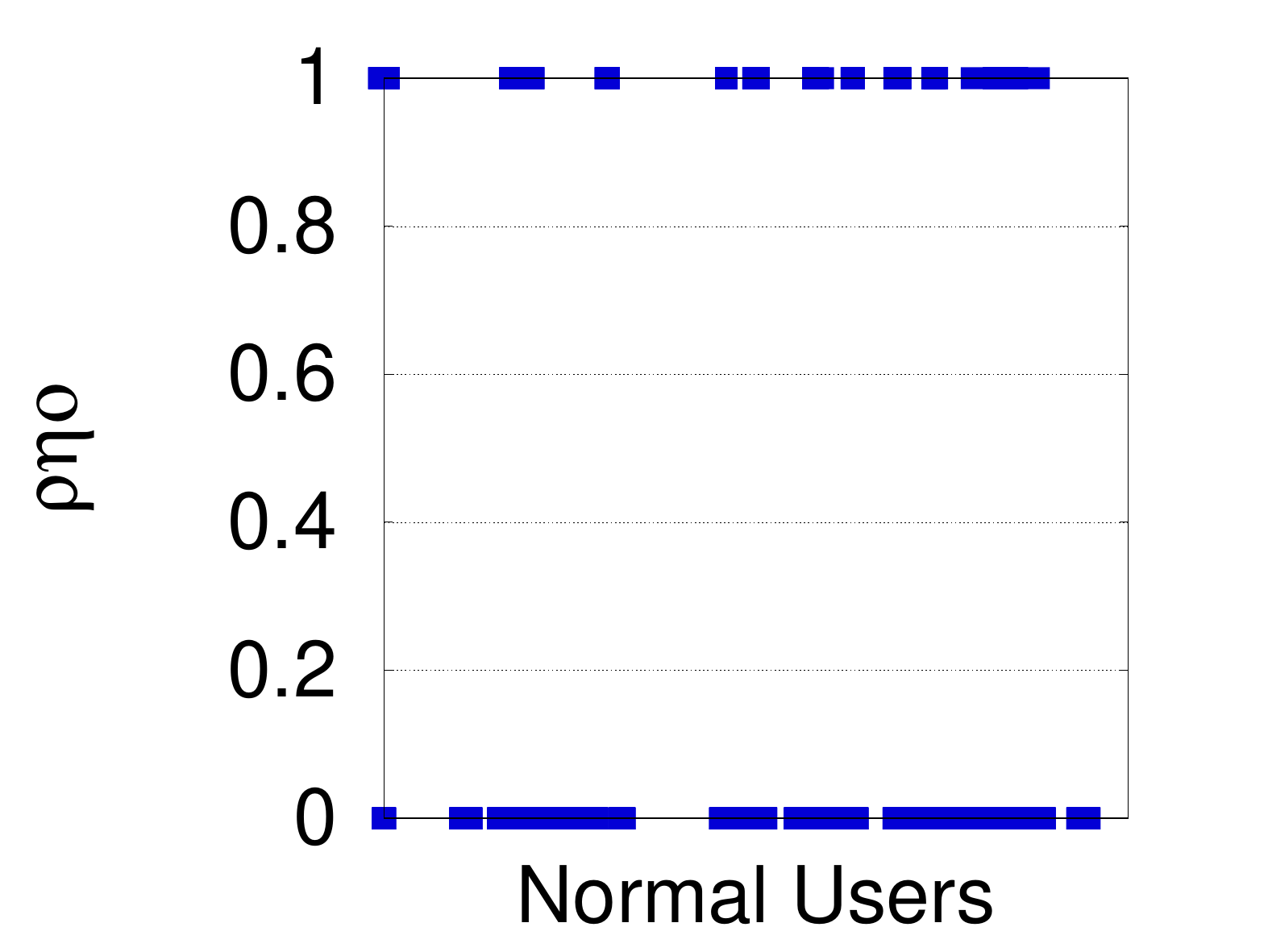}
  \caption{75 attackers}
  \label{fig:niid_weight_dist_75}
\end{subfigure} 
\end{center}
\caption{Weight Distribution for Non-IID data set with pure data in server}
\label{fig:niid_weight_dist}
\end{figure}

\begin{figure} 
\begin{center}
\begin{subfigure}{.49\textwidth}
  \centering
  \includegraphics[width=.49\linewidth]{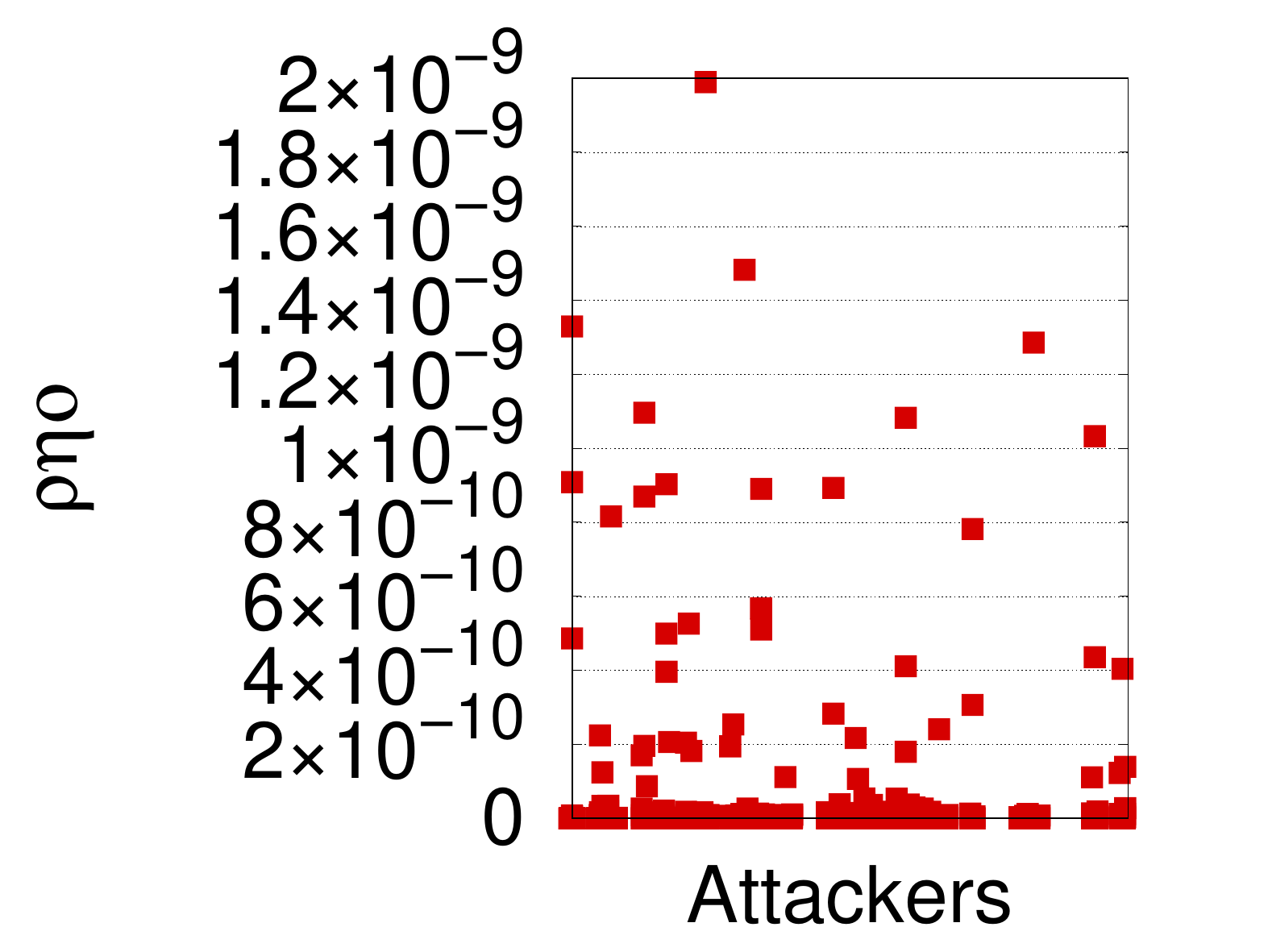}
  \includegraphics[width=.49\linewidth]{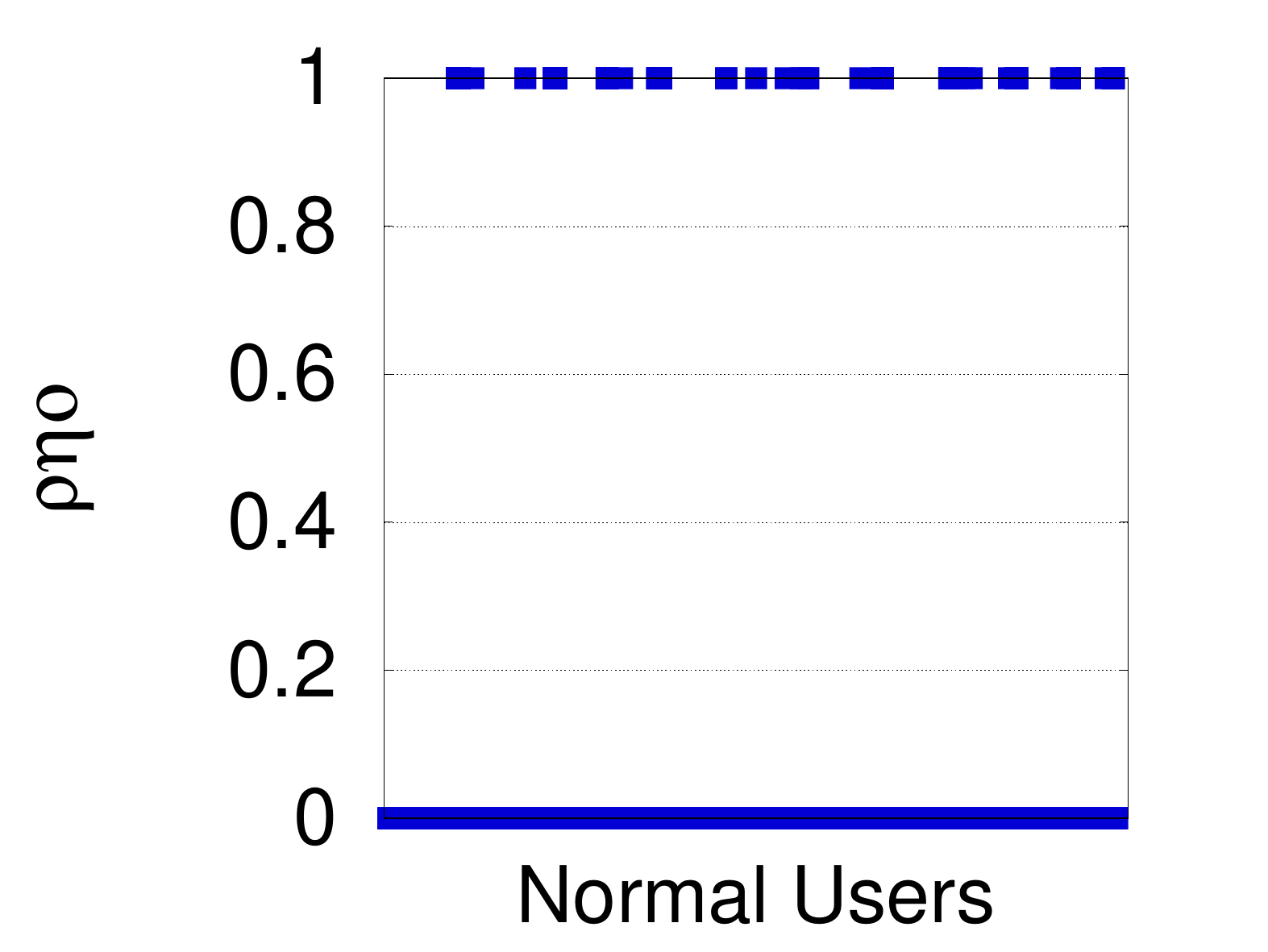}
  \caption{25 attackers}
  \label{fig:niid_np_weight_dist_25}
\end{subfigure}%
\begin{subfigure}{.49\textwidth}
  \centering
  \includegraphics[width=.49\linewidth]{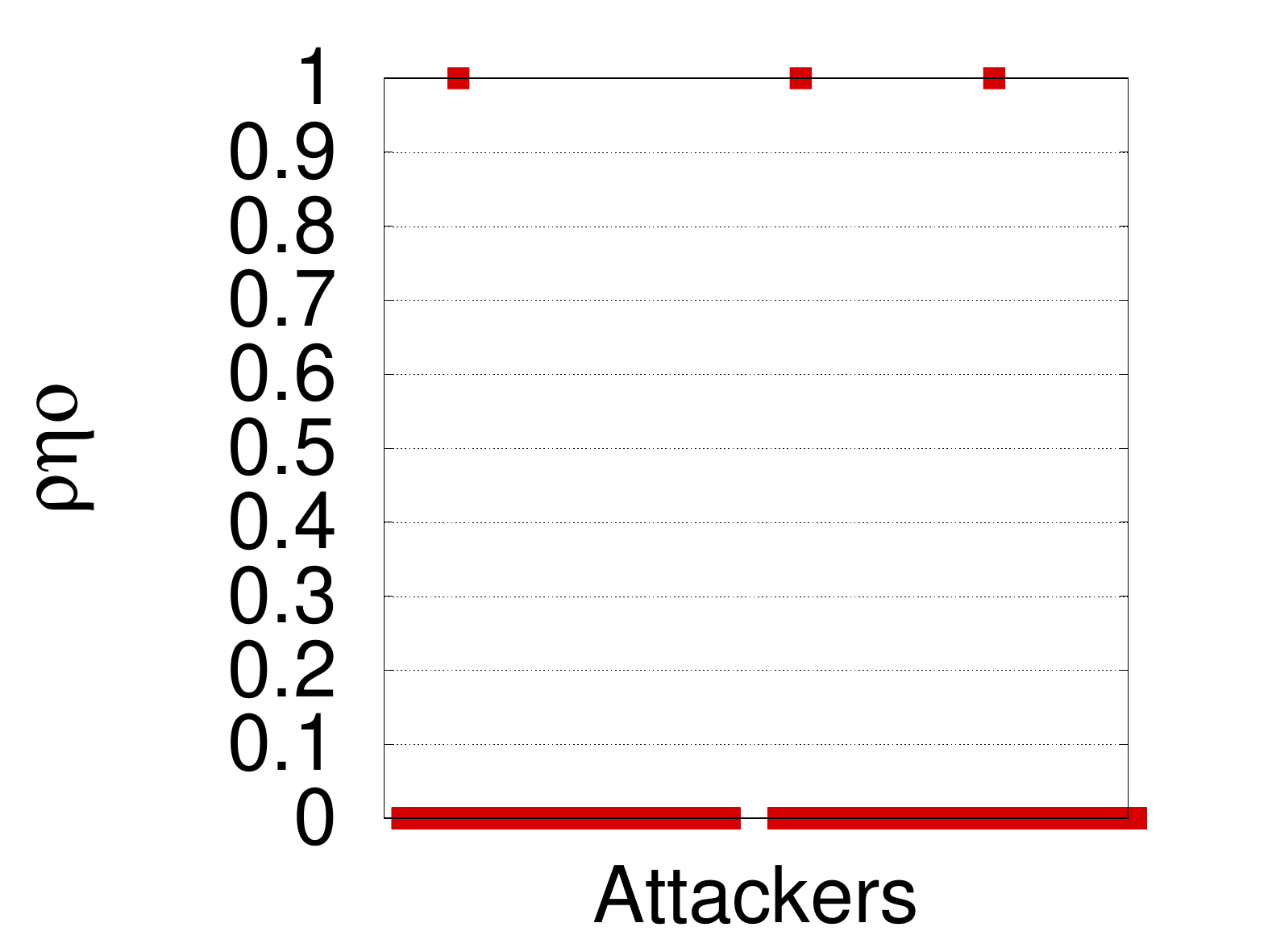}
  \includegraphics[width=.49\linewidth]{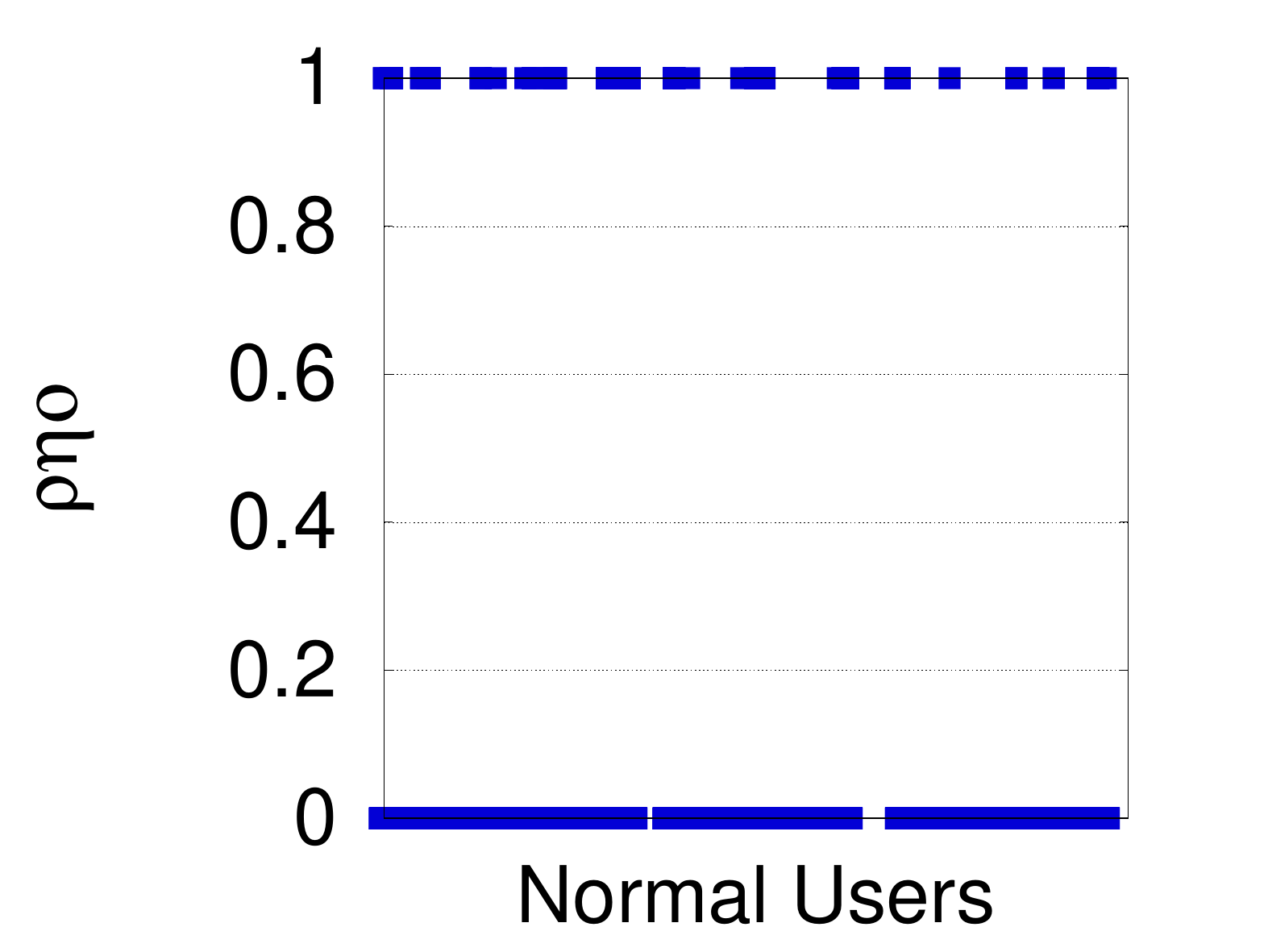}
  \caption{50 attackers}
  \label{fig:niid_np_weight_dist_50}
\end{subfigure}\\
\begin{subfigure}{.49\textwidth}
  \centering
  \includegraphics[width=.49\linewidth]{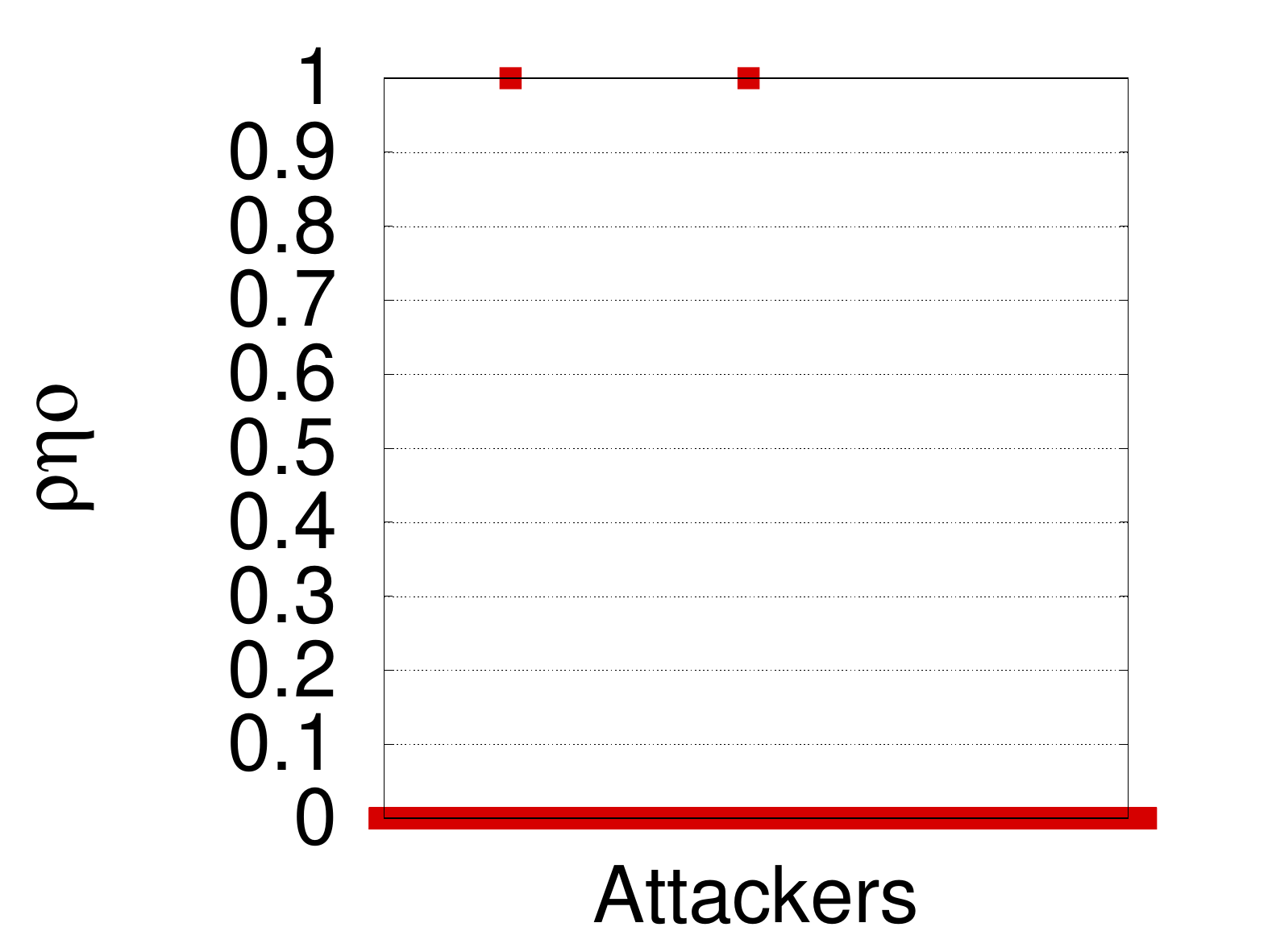}
  \includegraphics[width=.49\linewidth]{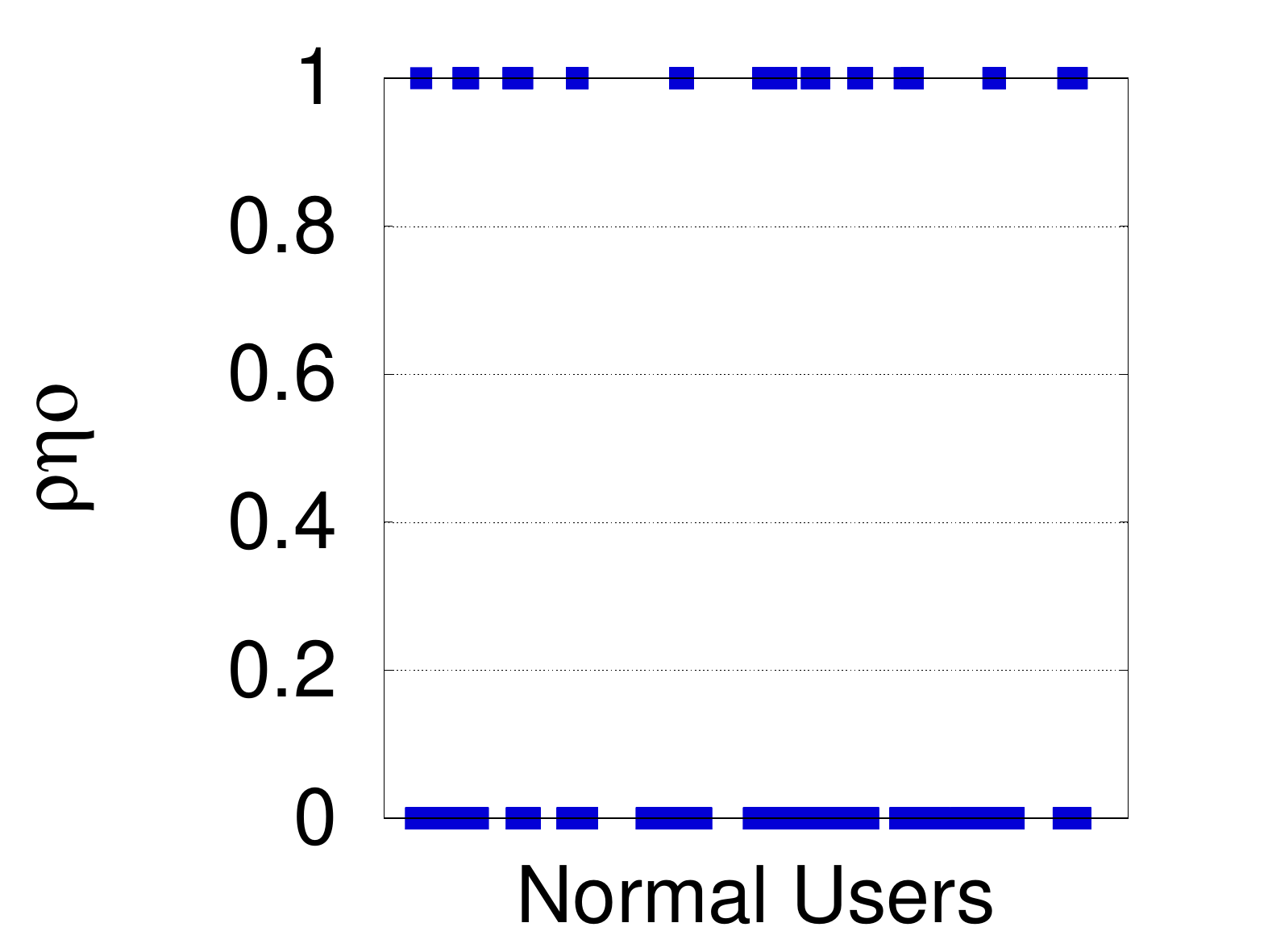}
  \caption{75 attackers}
  \label{fig:niid_np_weight_dist_75}
\end{subfigure}
\end{center}
\caption{Weight Distribution for Non-IID data set with impure data in server}
\label{fig:niid_np_weight_dist}
\end{figure}

In this subsection, we provide visualization of the weights and the values of $\rho_n$ for attackers and legitimate  users in MoE-FL versus FedAvg.   \Cref{fig:iid_weight_dist,fig:iid_np_weight_dist,fig:niid_weight_dist,fig:niid_np_weight_dist} shows the values of $\rho_n$ for attackers versus number of attackers for both IID and non-IID data sets and pure and impure $\mathcal{D}_0$, i.e., MoE-P and MoE-NP. For pure $\mathcal{D}_0$, in IID setting, MOE-FL can detect all the attackers and there is no $\rho_n \geq 2.5 \times 10^{-8}$ even in the worst case scenario with a large number of attackers (i.e., $E=75$). 

In IID setting, for $E=75$ with impure $\mathcal{D}_0$ (MoE-NP), for the first time, the MOE-FL is mislead by attackers and assigns $\rho_n=1$ for $n \in \mathcal{E}$. However, still for $E<75$, the values of $\rho_n$ for attackers is considerably less than that for the legitimate users, e.g., compare $2.5 \times 10^{-8}$ with 1 in Fig. \Cref{fig:iid_weight_dist,fig:iid_np_weight_dist,fig:niid_weight_dist,fig:niid_np_weight_dist} (a).

Comparing IID and non-IID data sets, we can conclude that the values of $\rho_n$ for attackers in non-IID are more than those for IID data sets relatively. However, MOE-FL is more vulnerable against attacks for non-IID data sets as we discussed in Lemma 2 and supported by the results in Figs. \ref{fig:niid_weight_dist} (c) and \ref{fig:niid_np_weight_dist} (c), and consequently $\rho_n$ obtains larger values for attackers. However, still having $\rho_n=1$ for legitimate users is more likely in all conditions compared to that for attackers. As aa result, MoE-FL can maintain its own performance even for a larger number of attackers in the pure and impure $\mathcal{D}_0$ for IID and non-IID data sets. 

\begin{figure}%
\begin{center}
\begin{subfigure}{.20\textwidth}
  \centering
  \includegraphics[width=.99\linewidth]{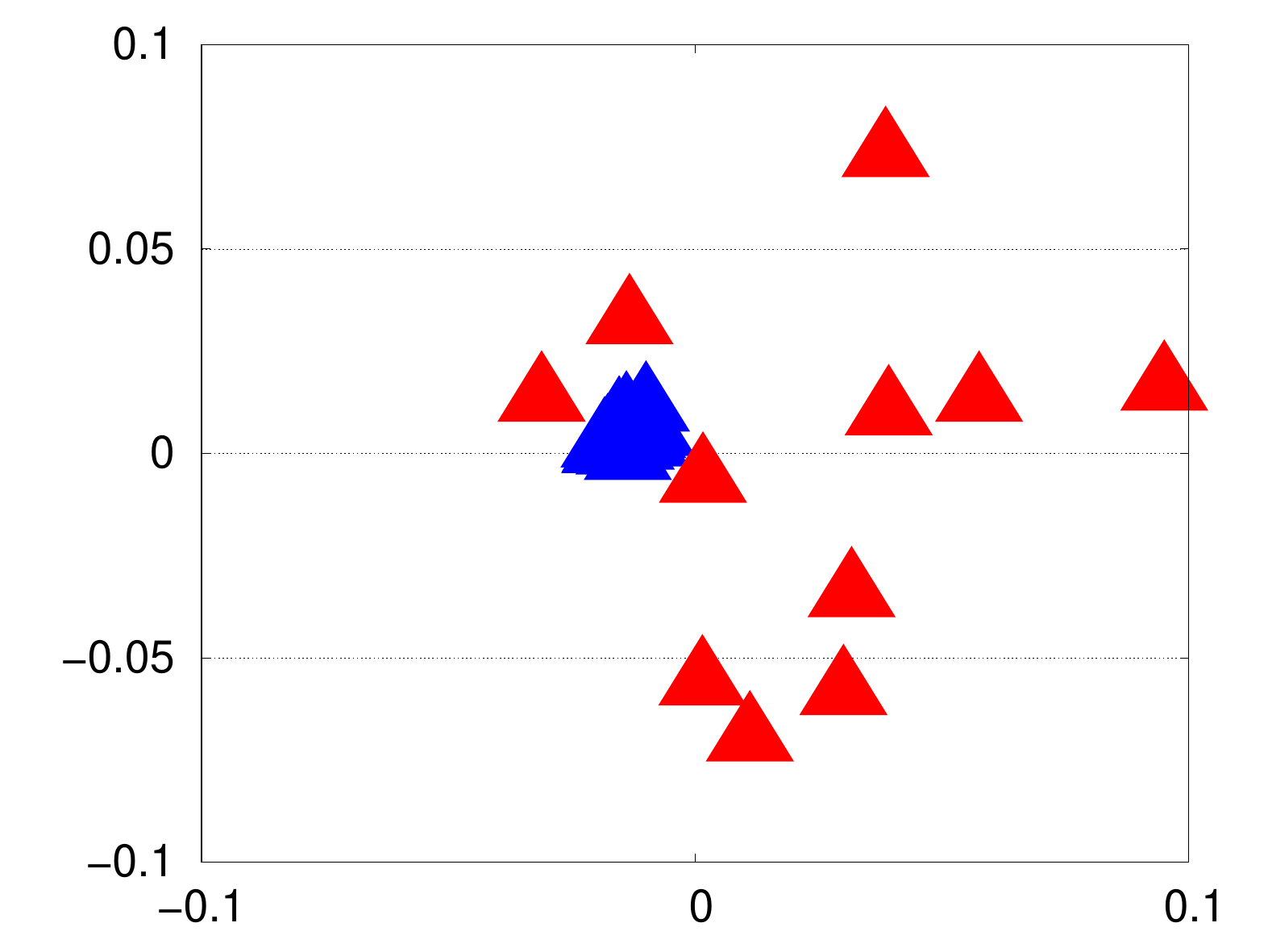}
  \caption{Round 0}
  \label{fig:pca:iid_avg_0}
\end{subfigure}%
\begin{subfigure}{.2\textwidth}
  \centering
  \includegraphics[width=.99\linewidth]{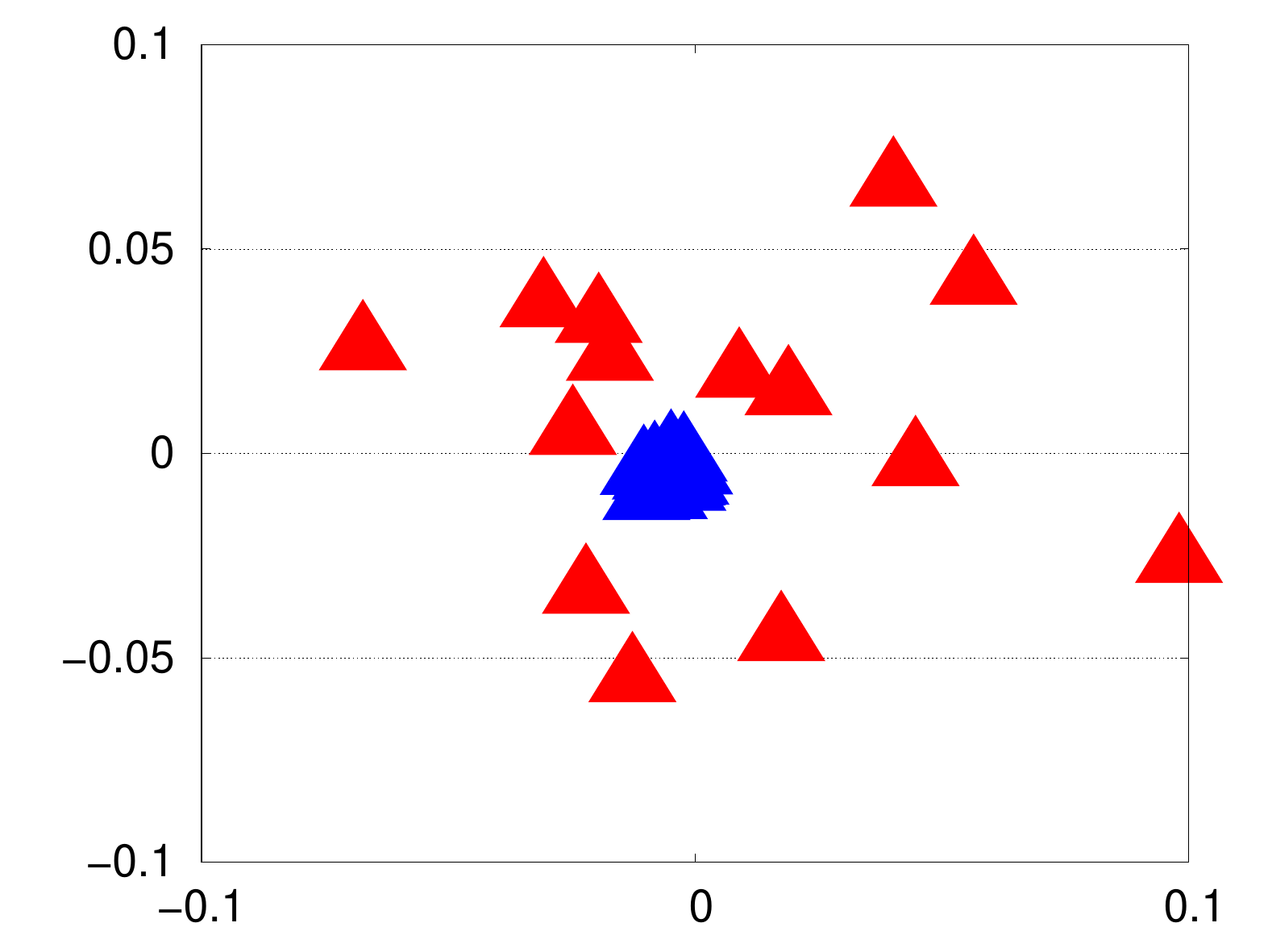}
  \caption{Round 50}
  \label{fig:pca:iid_avg_50}
\end{subfigure}%
\begin{subfigure}{.2\textwidth}
  \centering
  \includegraphics[width=.99\linewidth]{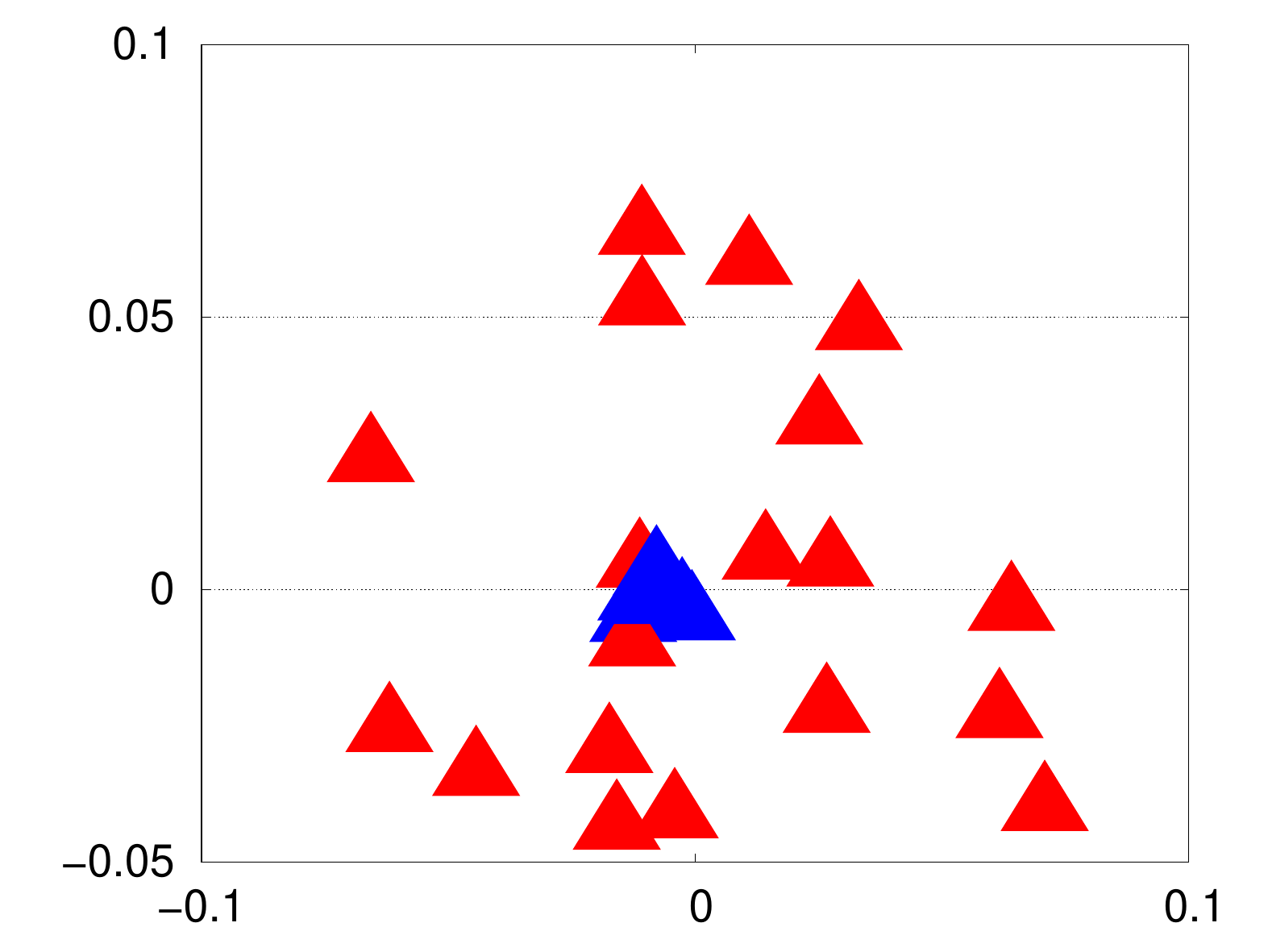}
  \caption{Round 100}
  \label{fig:pca:iid_avg_100}
\end{subfigure}%
\begin{subfigure}{.2\textwidth}
  \centering
  \includegraphics[width=.99\linewidth]{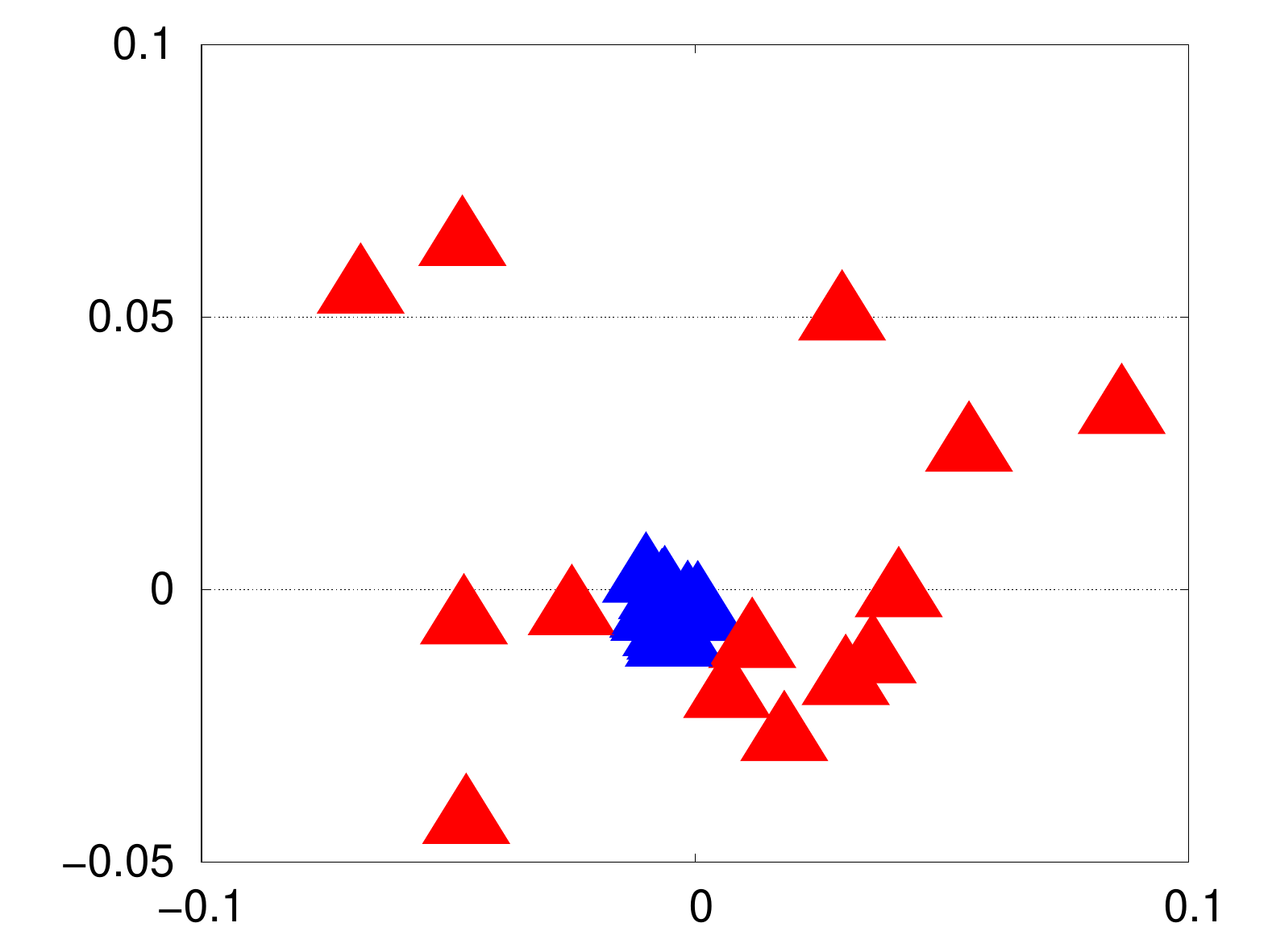}
  \caption{Round 150}
  \label{fig:pca:iid_avg_150}
\end{subfigure}%
\begin{subfigure}{.2\textwidth}
  \centering
  \includegraphics[width=.99\linewidth]{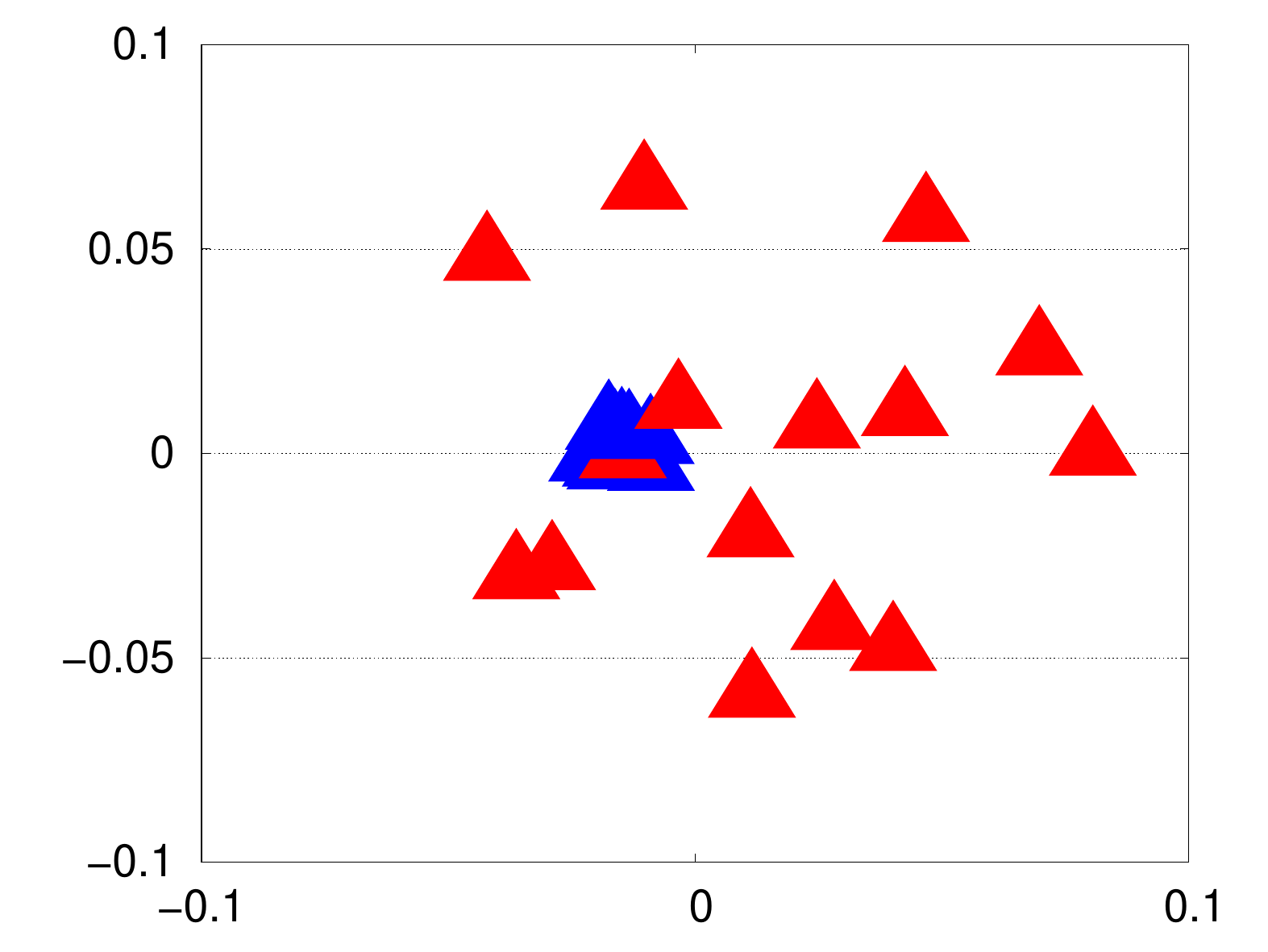}
  \caption{Round 200}
  \label{fig:pca:iid_avg_200}
\end{subfigure}\\
\begin{subfigure}{.20\textwidth}
  \centering
  \includegraphics[width=.99\linewidth]{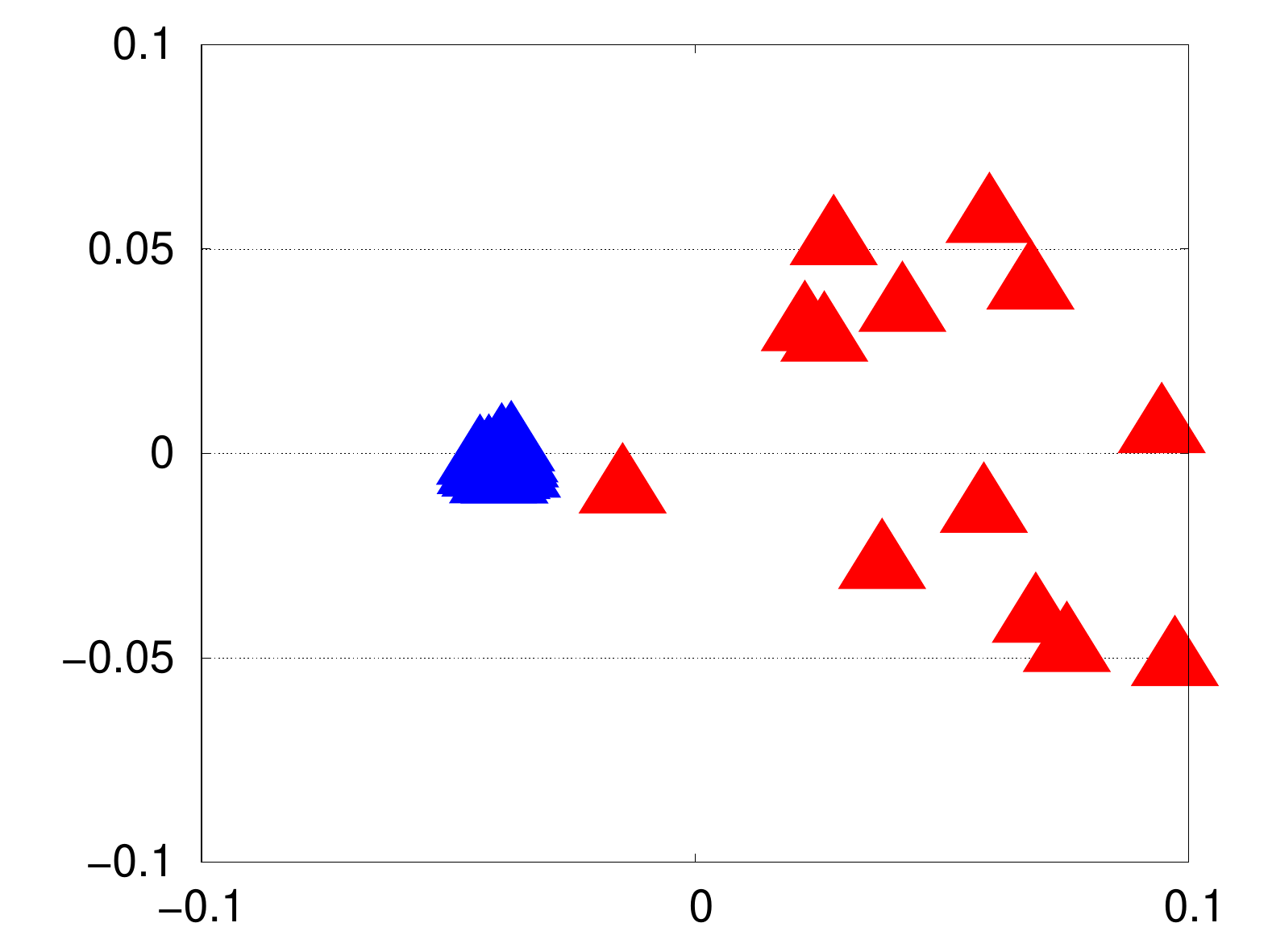}
  \caption{Round 0}
  \label{fig:pca:iid_moe_0}
\end{subfigure}%
\begin{subfigure}{.2\textwidth}
  \centering
  \includegraphics[width=.99\linewidth]{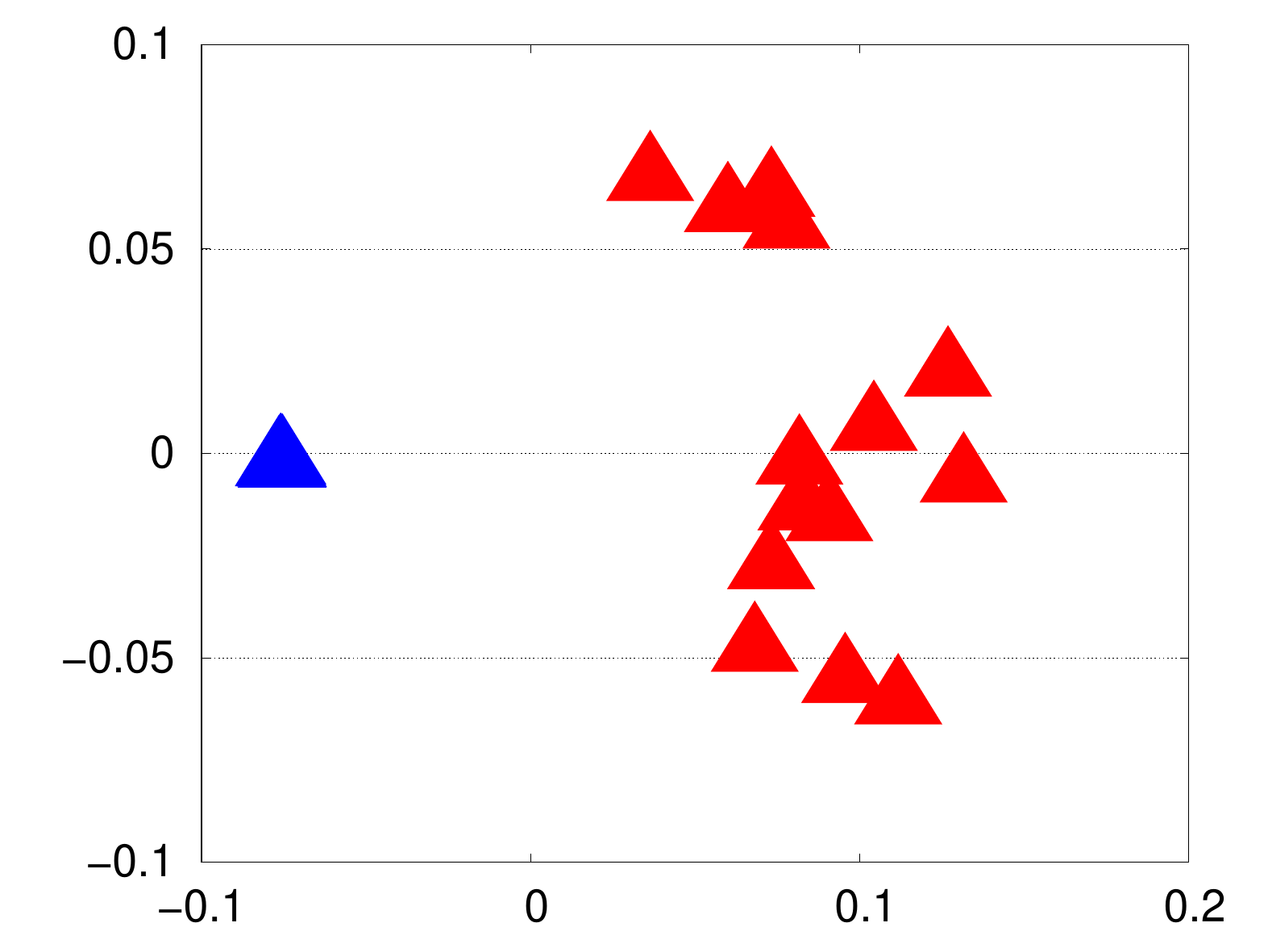}
  \caption{Round 50}
  \label{fig:pca:iid_moe_50}
\end{subfigure}%
\begin{subfigure}{.2\textwidth}
  \centering
  \includegraphics[width=.99\linewidth]{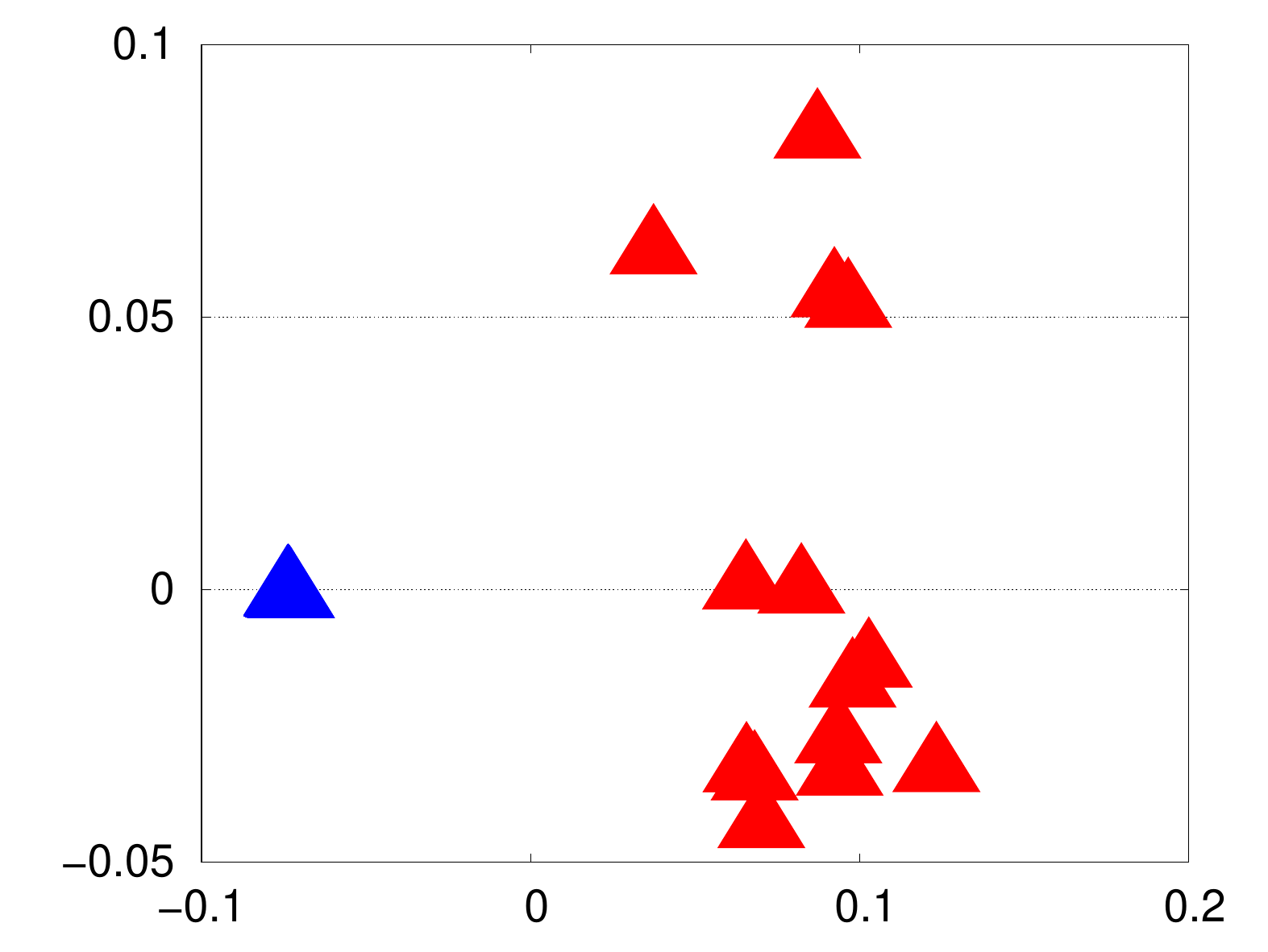}
  \caption{Round 100}
  \label{fig:pca:iid_moe_100}
\end{subfigure}%
\begin{subfigure}{.2\textwidth}
  \centering
  \includegraphics[width=.99\linewidth]{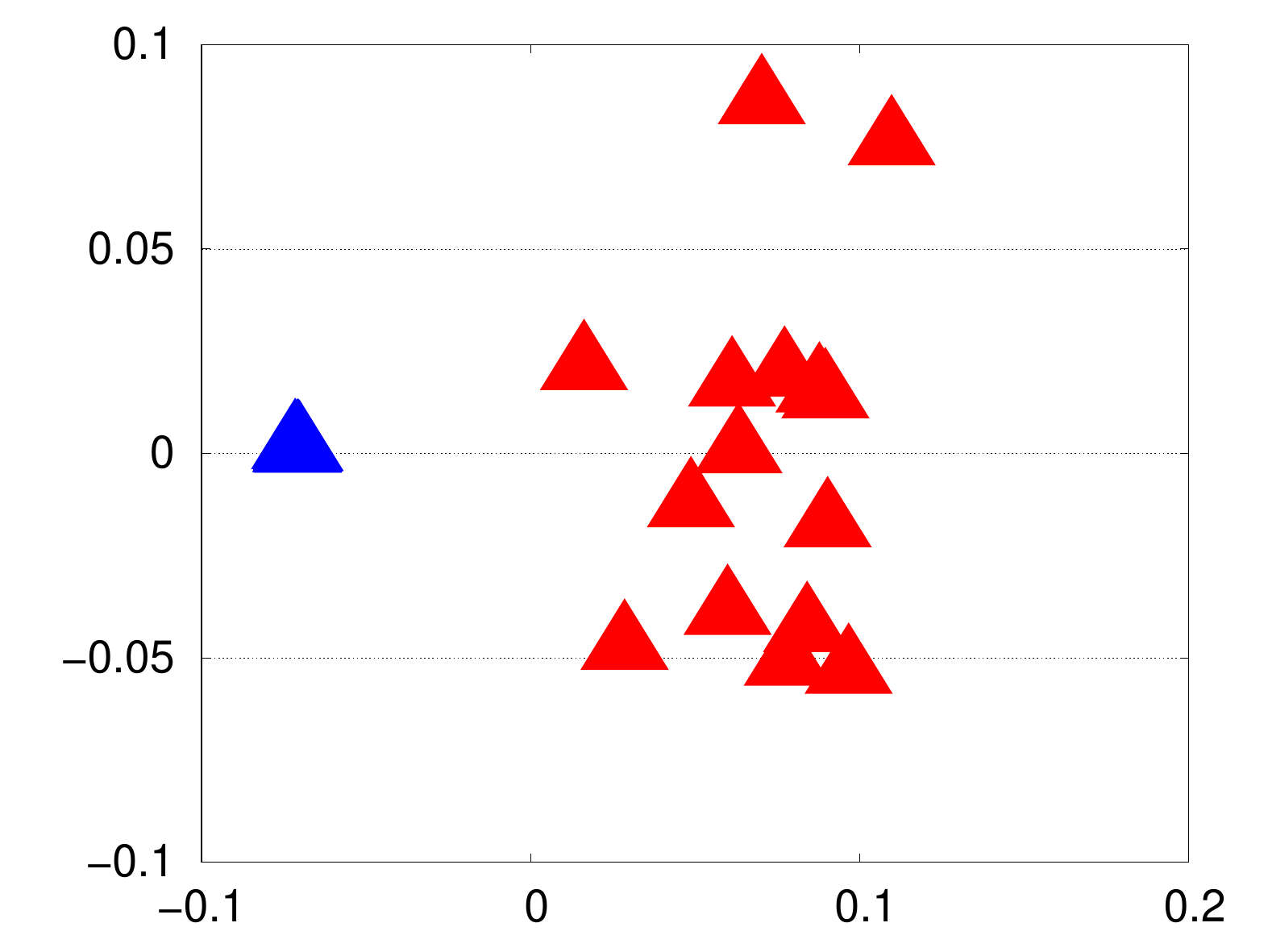}
  \caption{Round 150}
  \label{fig:pca:iid_moe_150}
\end{subfigure}%
\begin{subfigure}{.2\textwidth}
  \centering
  \includegraphics[width=.99\linewidth]{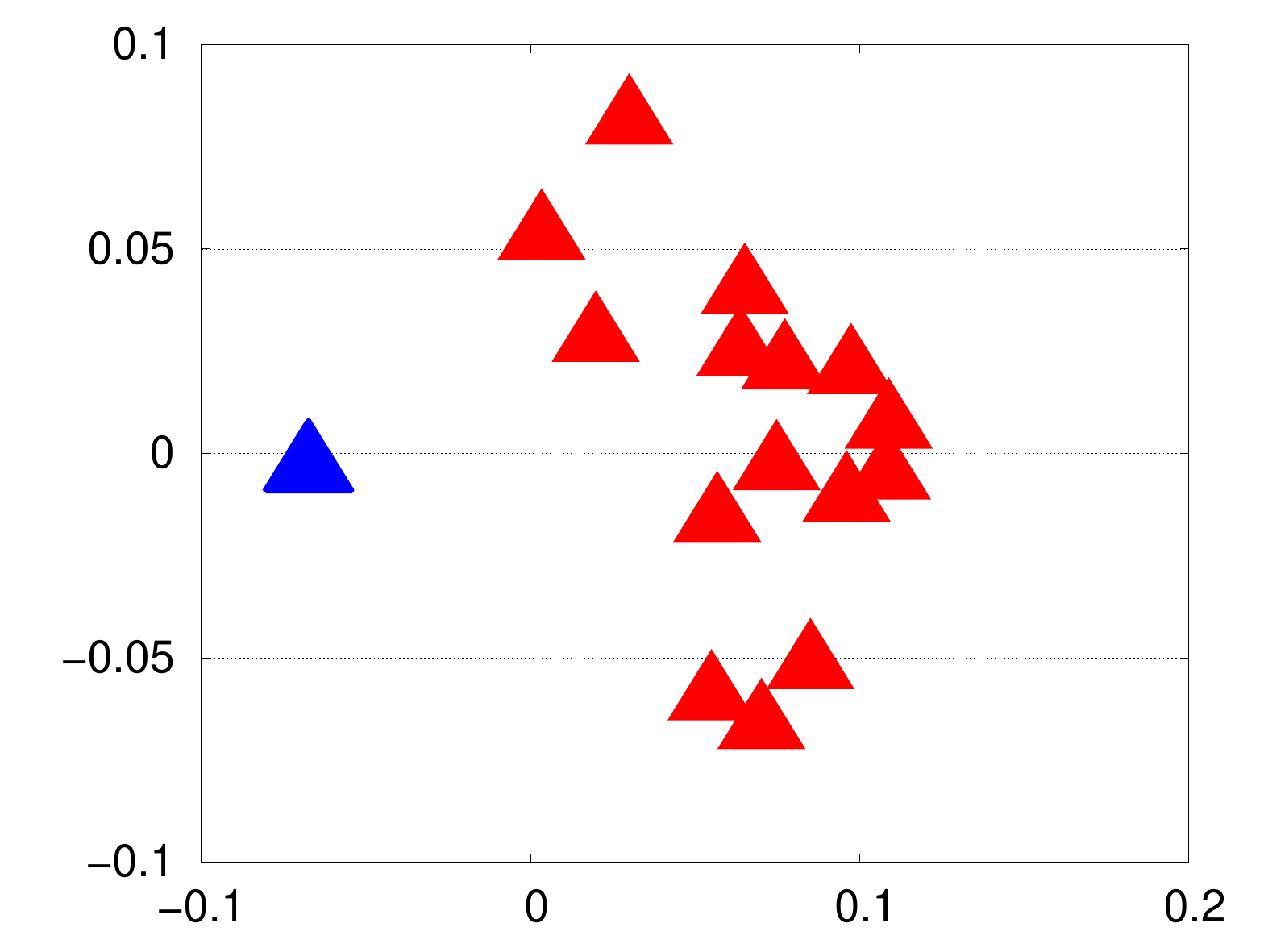}
  \caption{Round 200}
  \label{fig:pca:iid_moe_200}
\end{subfigure}%
\end{center}
\caption{PCA projection on $\mathbb{R}2$ of model parameters for IID data set for FedAVG 
(\cref{fig:pca:iid_avg_0,fig:pca:iid_avg_50,fig:pca:iid_avg_100,fig:pca:iid_avg_150,fig:pca:iid_avg_200}) and MOE-FL 
(\cref{fig:pca:iid_moe_0,fig:pca:iid_moe_50,fig:pca:iid_moe_100,fig:pca:iid_moe_150,fig:pca:iid_moe_200}) with 50 attackers
}
\label{fig:pca_iid_50}
\end{figure}

\begin{figure}[ht!]
\begin{center}
\begin{subfigure}{.20\textwidth}
  \centering
  \includegraphics[width=.99\linewidth]{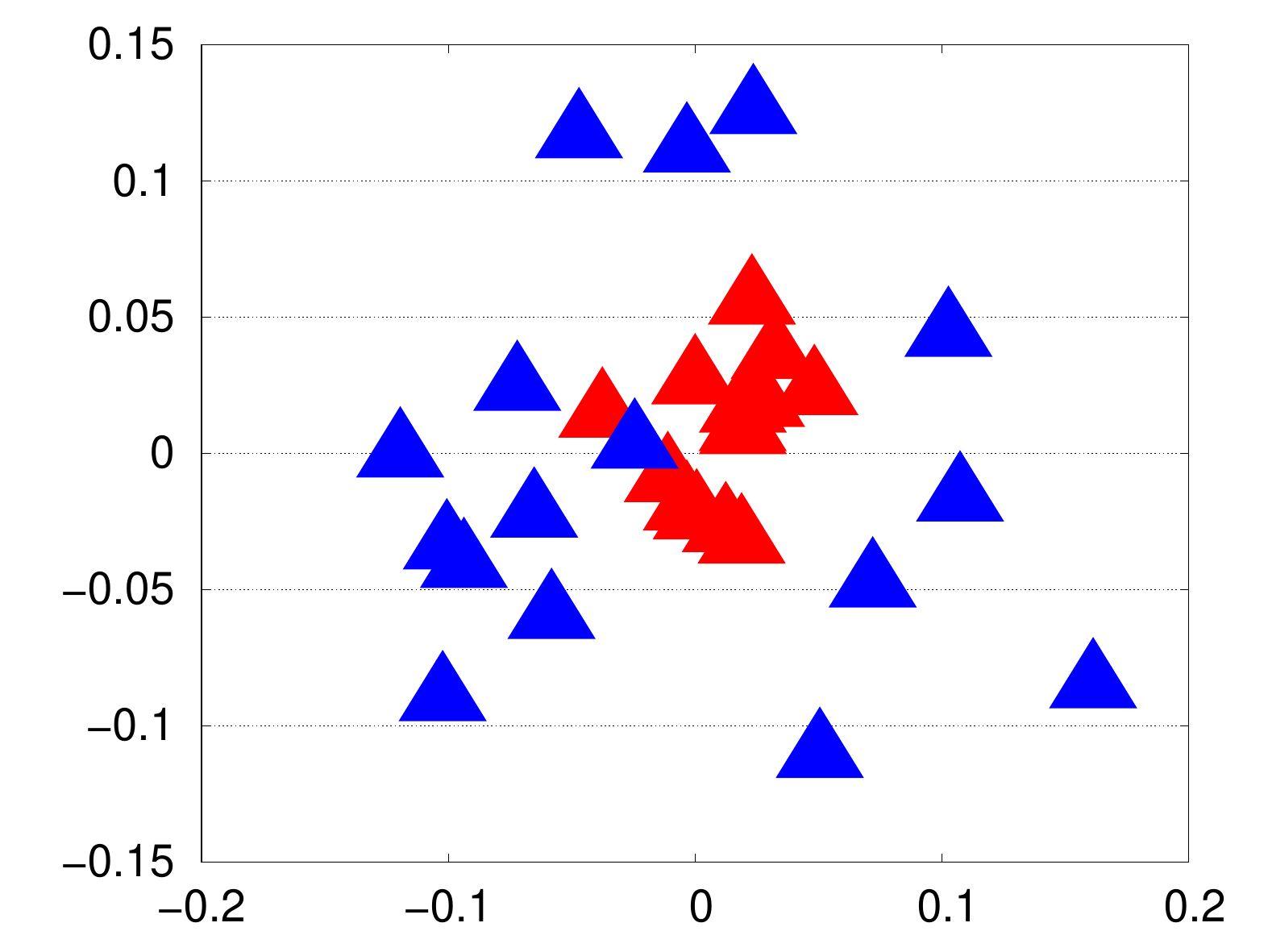}
  \caption{Round 0}
  \label{fig:pca:niid_avg_0}
\end{subfigure}%
\begin{subfigure}{.2\textwidth}
  \centering
  \includegraphics[width=.99\linewidth]{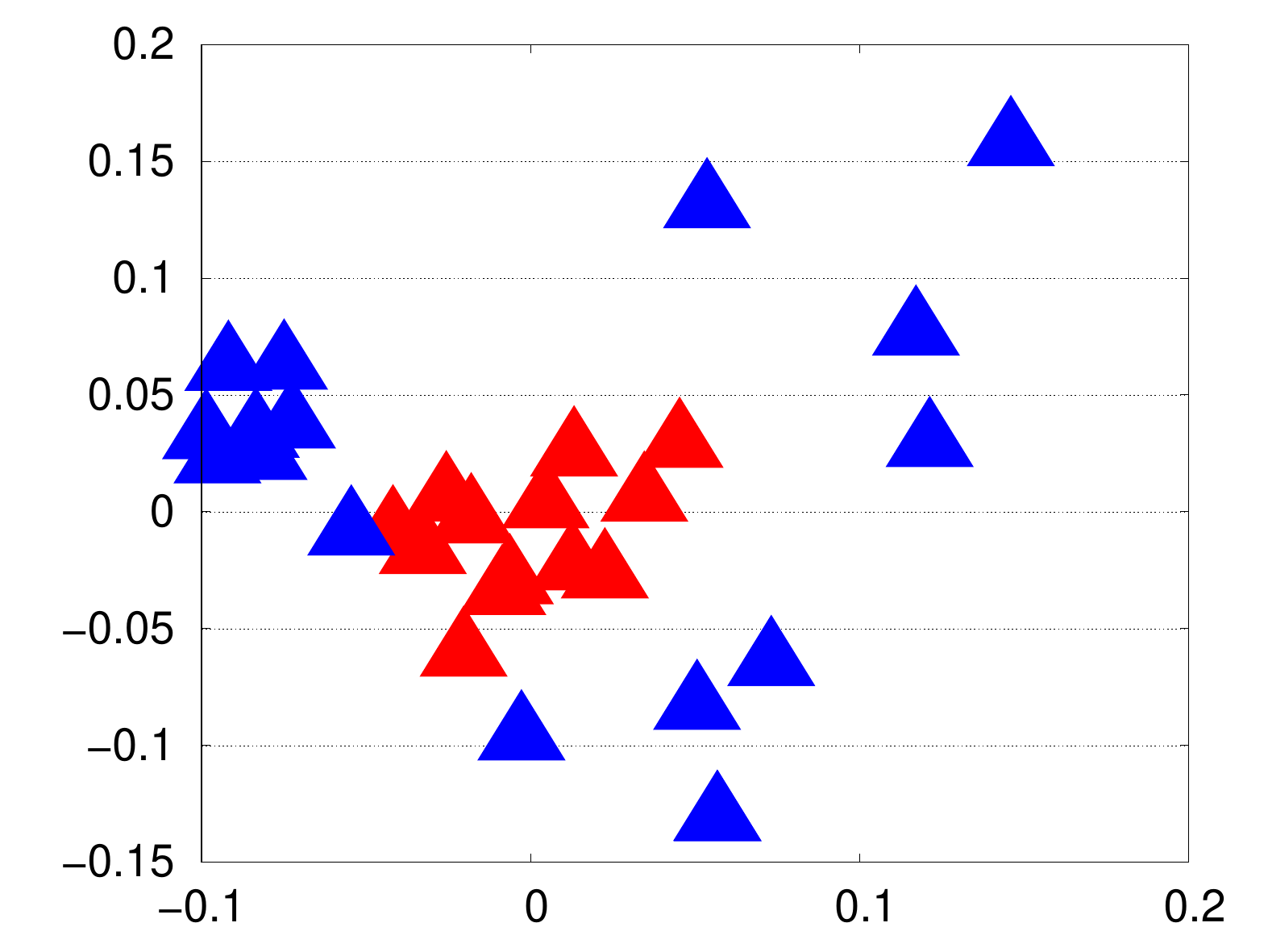}
  \caption{Round 50}
  \label{fig:pca:niid_avg_50}
\end{subfigure}%
\begin{subfigure}{.2\textwidth}
  \centering
  \includegraphics[width=.99\linewidth]{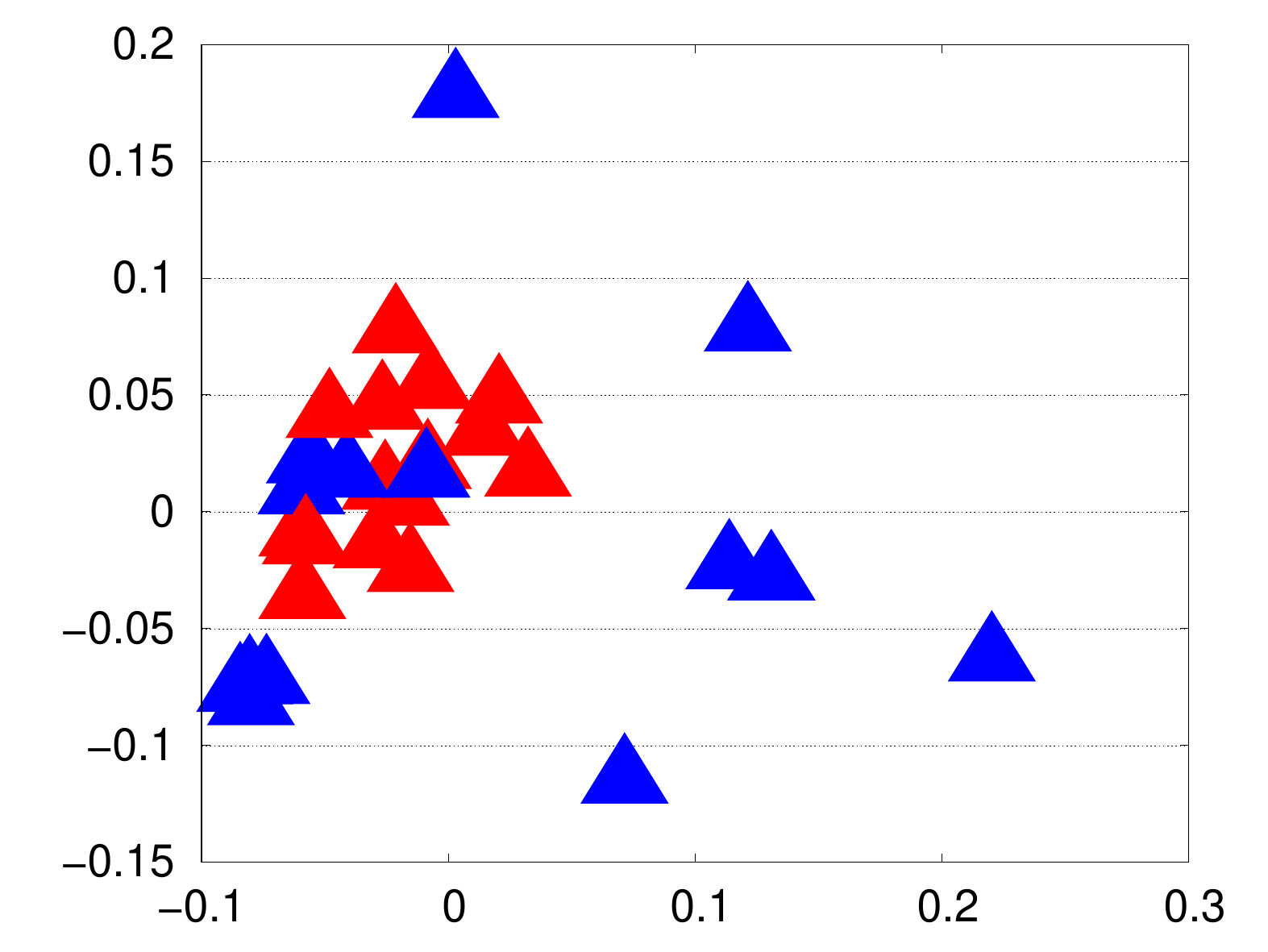}
  \caption{Round 100}
  \label{fig:pca:niid_avg_100}
\end{subfigure}%
\begin{subfigure}{.2\textwidth}
  \centering
  \includegraphics[width=.99\linewidth]{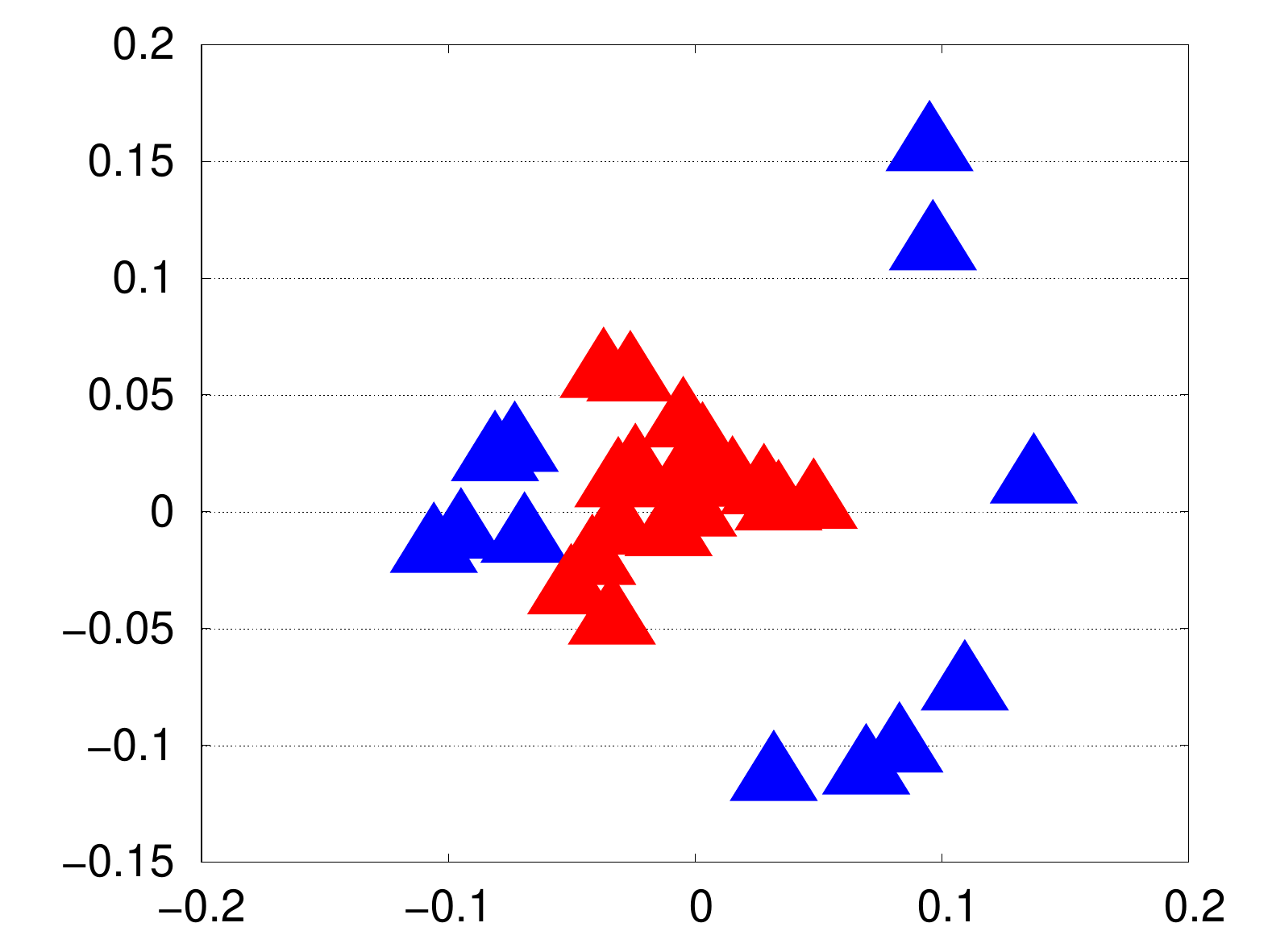}
  \caption{Round 150}
  \label{fig:pca:niid_avg_150}
\end{subfigure}%
\begin{subfigure}{.2\textwidth}
  \centering
  \includegraphics[width=.99\linewidth]{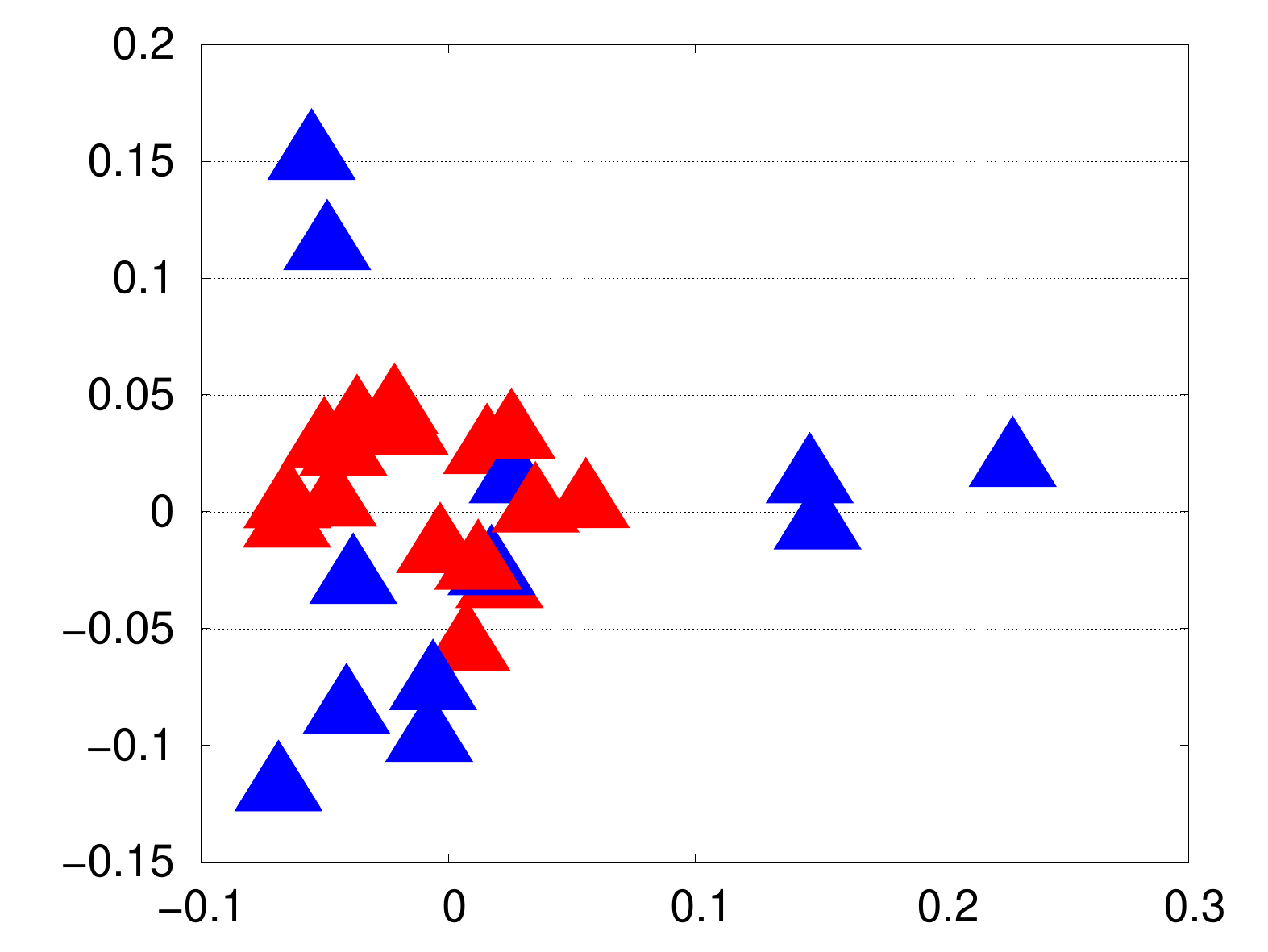}
  \caption{Round 200}
  \label{fig:pca:niid_avg_200}
\end{subfigure}\\
\begin{subfigure}{.20\textwidth}
  \centering
  \includegraphics[width=.99\linewidth]{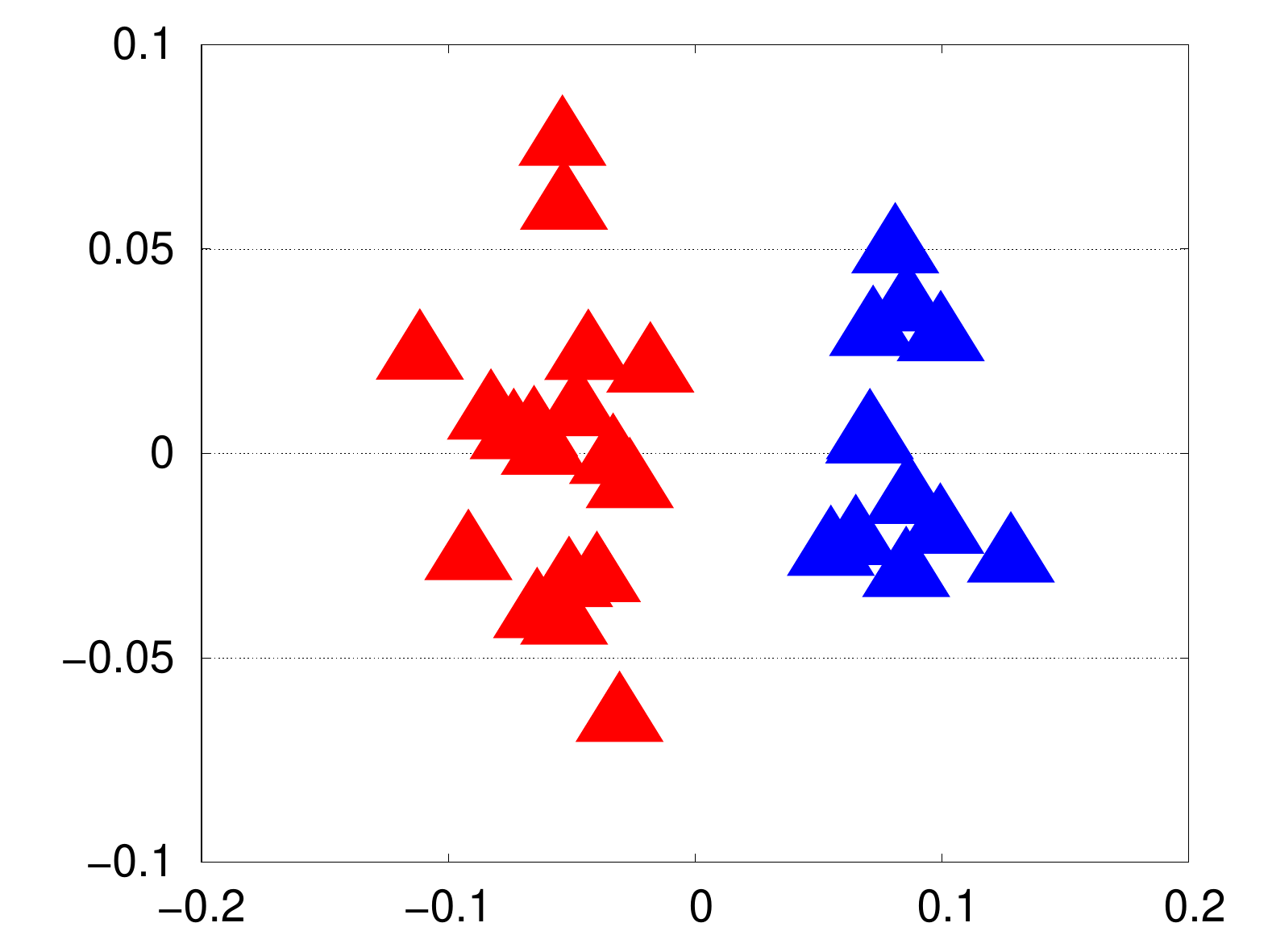}
  \caption{Round 0}
  \label{fig:pca:niid_moe_0}
\end{subfigure}%
\begin{subfigure}{.2\textwidth}
  \centering
  \includegraphics[width=.99\linewidth]{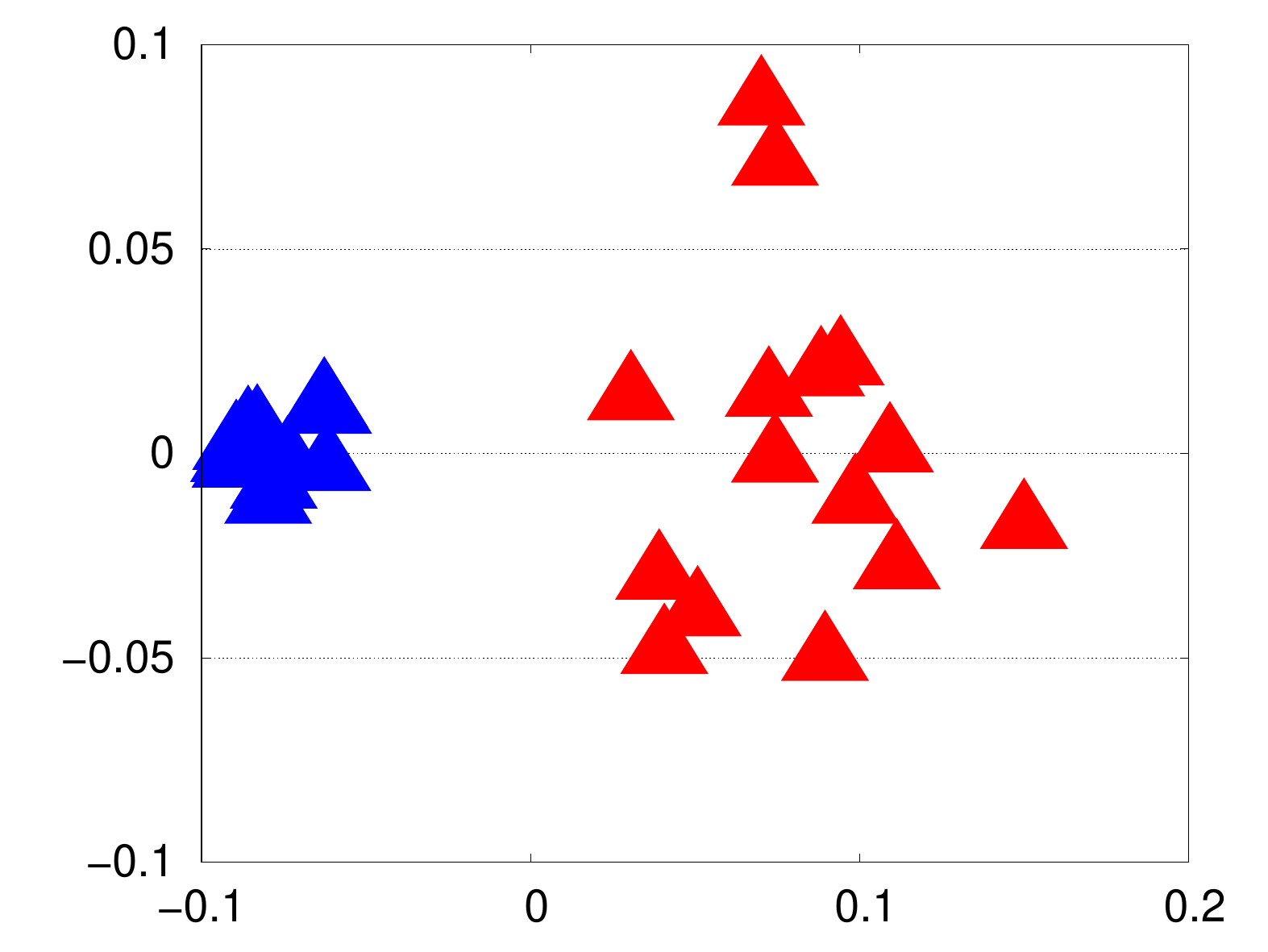}
  \caption{Round 50}
  \label{fig:pca:niid_moe_50}
\end{subfigure}%
\begin{subfigure}{.2\textwidth}
  \centering
  \includegraphics[width=.99\linewidth]{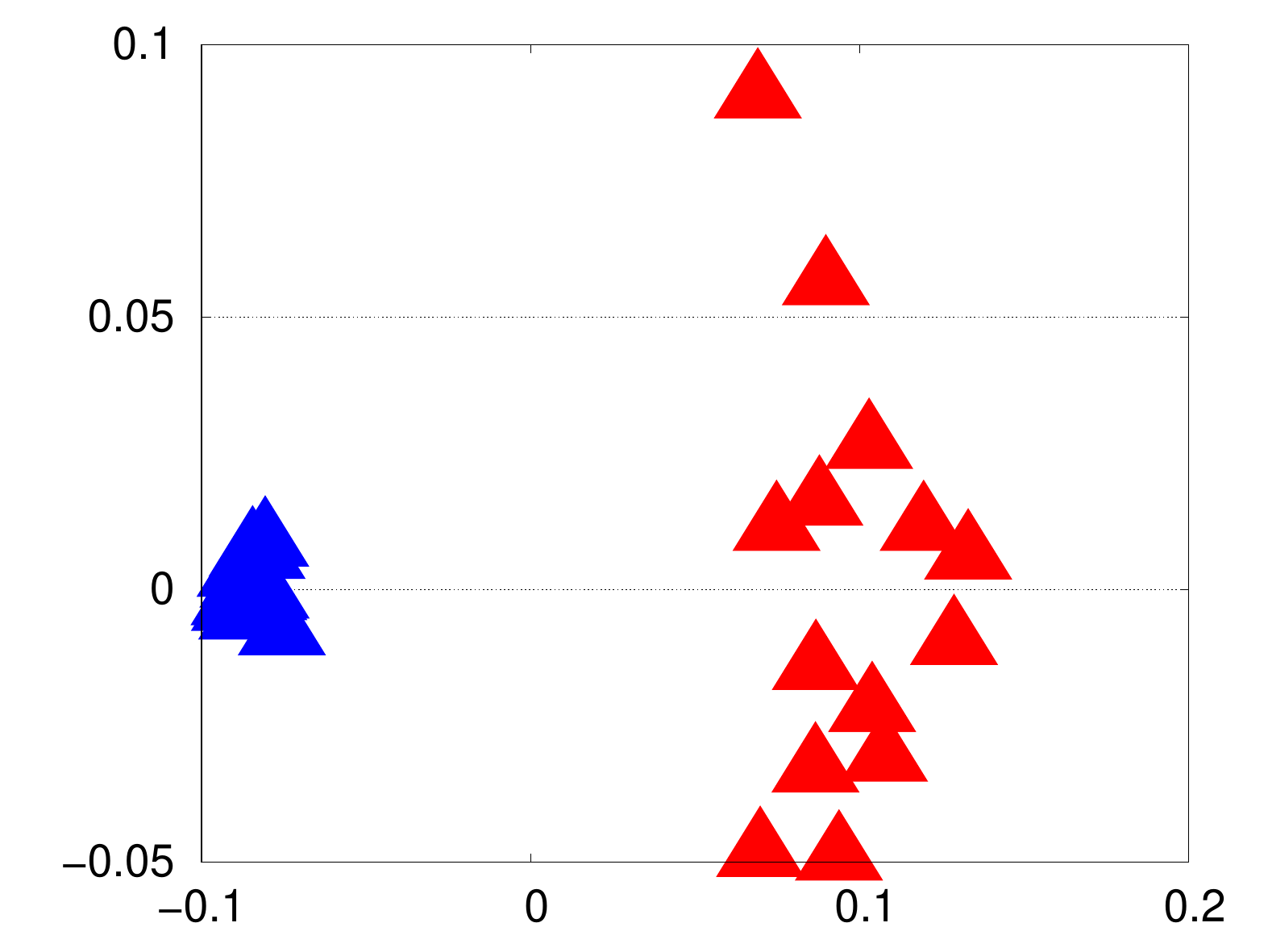}
  \caption{Round 100}
  \label{fig:pca:niid_moe_100}
\end{subfigure}%
\begin{subfigure}{.2\textwidth}
  \centering
  \includegraphics[width=.99\linewidth]{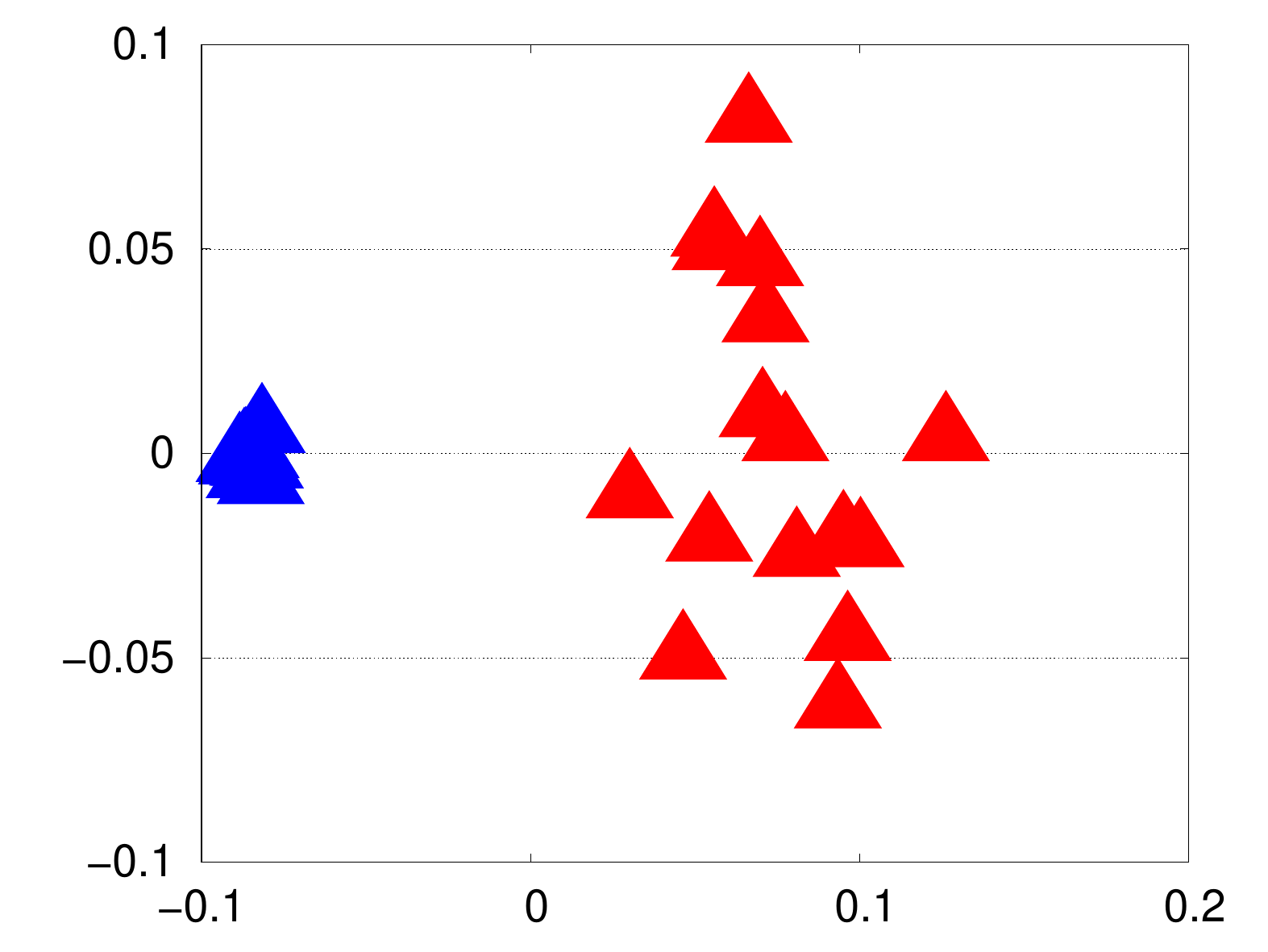}
  \caption{Round 150}
  \label{fig:pca:niid_moe_150}
\end{subfigure}%
\begin{subfigure}{.2\textwidth}
  \centering
  \includegraphics[width=.99\linewidth]{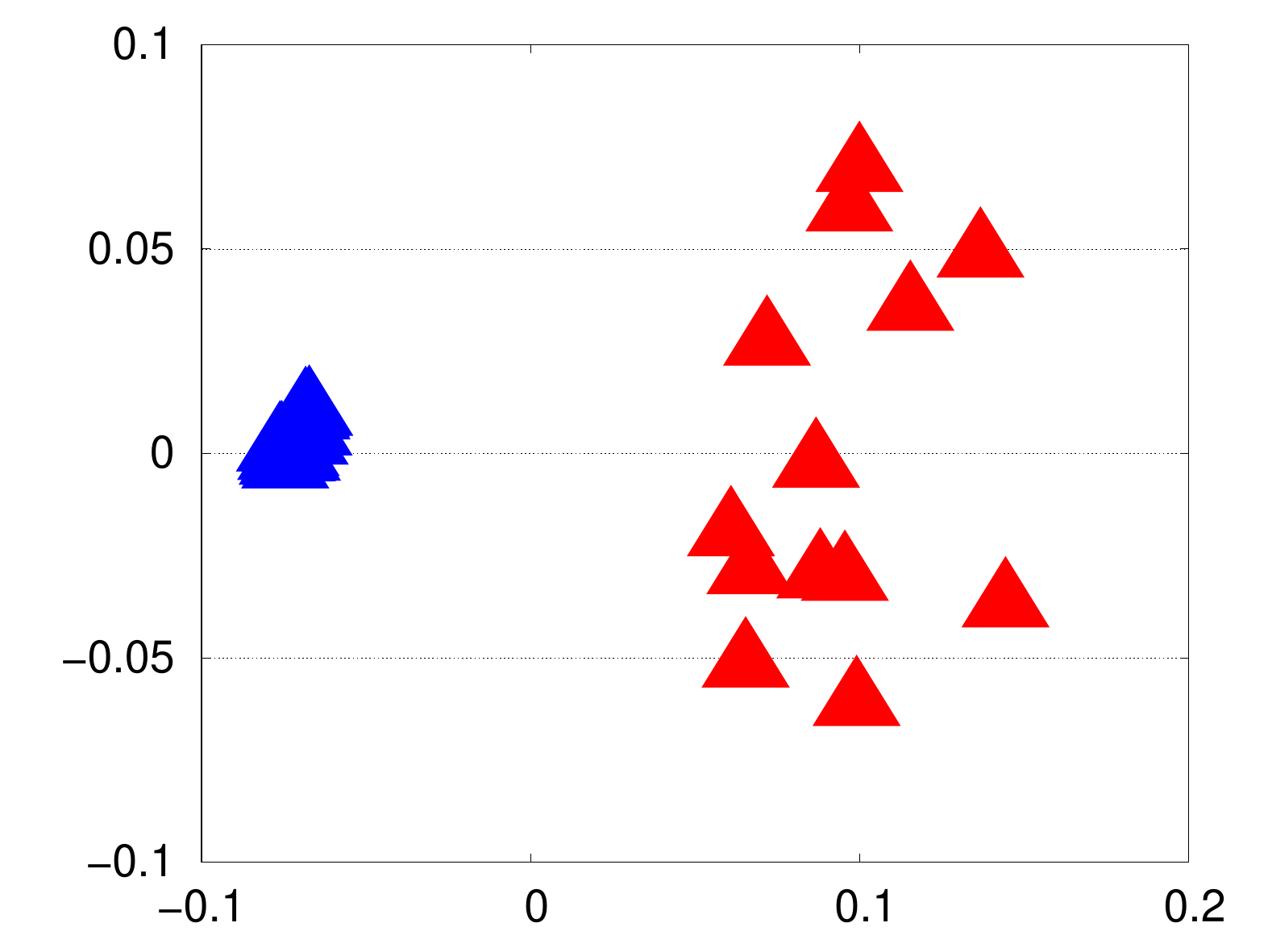}
  \caption{Round 200}
  \label{fig:pca:niid_moe_200}
\end{subfigure}%
\end{center}
\caption{PCA projection on $\mathbb{R}2$ of model parameters for Non-IID data set for FedAVG
(\cref{fig:pca:niid_avg_0,fig:pca:niid_avg_50,fig:pca:niid_avg_100,fig:pca:niid_avg_150,fig:pca:niid_avg_200}) and MOE-FL  
(\cref{fig:pca:niid_moe_0,fig:pca:niid_moe_50,fig:pca:niid_moe_100,fig:pca:niid_moe_150,fig:pca:niid_moe_200}) with 50 attackers
}
\label{fig:pca_niid_50}
\end{figure}

In Figs. \ref{fig:pca_iid_50}-\ref{fig:pca_niid_50}, we study the principal component analysis (PCA) of model parameters distribution of attackers and legitimate users in various rounds for the Fed-Avg and MoE-FL. To illustrate Figs. \ref{fig:pca_iid_50}-\ref{fig:pca_niid_50}, we use the last layer of the neural network parameters of each user $n$ and project them on $\mathbb{R}2$. Model parameters of attackers and legitimate users are illustrated in red and blue, respectively.


In each figure, the first and second row are corresponding to the FedAvg and MoE, respectively. For both IID and non-IID, from Figs. \ref{fig:pca_iid_50}-\ref{fig:pca_niid_50}, the distance of PCA of models of attackers and legitimate users in MOE-FL is larger than that for FedAvg. This shows that by using MOE-FL the weights of legitimate users can converge to the correct model when data sets are not poisoned. However, for FedAvg, the models of legitimate  users are close to the attackers, the models are poisoned and there is no way to distinguish legitimate users and attackers. In non-IID data sets, for 200 rounds, the models of legitimate users in FedAvg are interleaved with attackers than MoE-FL, meaning the models of legitimate users are affected by the updates from the server. However, for MoE-FL, the models of legitimate users are still close together and far from the attackers, showing that the weighted average in MoE-FL eliminates the effect of the attackers. These figures highlight the importance of more sophisticated aggregation methods for FL to have a more robust behaviour against attackers.

\section{Conclusion}
In this paper, by the concept of mixture of experts (MoE), we revisit a structure for federated learning algorithm to distinguish the outliers/attackers and poisoned/outdated data sets of users. Our proposed approach can profoundly introduce robustness in FL against the attackers. This approach outperforms the FedAvg algorithm by 70\% where a larger number of attackers with a more sophisticated attacks are present. The potential applications of MoE in the concept of distributed and federated learning are considered as a future work of this paper.


%

\appendices
\section{Appendix A: Proof of Lemma 1}
Let's study the behaviour of objective function in \eqref{MOEoptimization} and MOEA over the IID local data sets without attackers for $N \gg 1$ as

     \begin{eqnarray}\label{AppendixA}
   \lefteqn{ \mathbb{E}_{\textbf{z}} (\|\sum_{n\in \mathcal{N}} \rho_n \textbf{w}_n- \textbf{w}_0\|)}\\&& \approx \mathbb{E}_{\textbf{z}} (\|\sum_{n\in \mathcal{N}} \rho_n\nabla V_n(\textbf{w}) - \nabla V(\textbf{w}) - (\underbrace{\nabla V_0(\textbf{w}_0)}_{\sum_{n\in \mathcal{N}}\nabla V_n(\textbf{w}_0)} - \nabla V(\textbf{w}))\|)\nonumber \\&&\approx \|\sum_{n\in \mathcal{N}} \rho_n (\frac{1}{M_n}\sum_{\textbf{z}_n} \nabla  {v}_n(\textbf{w}, \textbf{z}_n) - \frac{1}{N}\sum_{n\in \mathcal{N}} \sum_{\textbf{z}_n} \frac{1}{M_n}\nabla  {v}_n(\textbf{w}, \textbf{z}) - \underbrace{\left(\sum_{n\in \mathcal{N}}\nabla V_n(\textbf{w}) - \sum_{n\in \mathcal{N}}\nabla V_n(\textbf{w})\right)}_{\text{for} N \gg 1, \text{IID assumption, this sum is equal to} 0}\| \nonumber \\&&
     \underbrace{\approx}_{\text{from IID assumption}} \sum_{\textbf{z}_n} \frac{1}{M_n}\nabla  {v}_n(\textbf{w}, \textbf{z}) \times \underbrace{\| \sum_{n \in \mathcal{N}} \rho_n -1 \|}_{0} =0. \nonumber
     \end{eqnarray}

\eqref{AppendixA} shows that MoE-FL for the IID data sets is unbiased function regardless of MoEA method and since both $\textbf{w}_n$ and $\textbf{w}_0$ have similar probability distribution. Therefore, the performance cannot be changed by MoE in this case. The converges rate is related to the number of users participating in each round in FL \cite{koloskova2020unified}. For FedAvg, the number of users per each iteration is fixed and the performing is related to the number of users per each epoch i.e., $N$ and the accuracy is related to $O(\frac{1}{\sqrt{N}})$ for non-convex and $O(\frac{1}{N})$ for convex loss function e.g., \cite{yi2020primaldual}. However, for MoE-FL, if there exists zero $\alpha_n$, the convergence rate is increased compared to FedAvg. Similar deduction can be presented for the case that all users are attackers. Still, the performance of FedAvg for convergence of all attackers scenario is better than that MoE-FL.

For non-IID data scenarios without attackers and when $\sigma_n=\sigma$ and $\eta_n=\eta$, we have  
 \begin{eqnarray}\label{AppendixA-2}
   \lefteqn{ \mathbb{E}_{\textbf{z}} (\|\sum_{n\in \mathcal{N}} \rho_n \textbf{w}_n- \textbf{w}_0\|)}\\&& \approx \mathbb{E}_{\textbf{z}} (\|\sum_{n\in \mathcal{N}} \rho_n\nabla V_n(\textbf{w}) - \nabla V(\textbf{w}) - (\underbrace{\nabla V_0(\textbf{w}_0)}_{\sum_{n\in \mathcal{N}}\nabla V_n(\textbf{w}_0)=} - \nabla V(\textbf{w}))\|)\nonumber \\&& \leq  \mathbb{E}_{\textbf{z}} (\|\sum_{n\in \mathcal{N}} \rho_n\nabla V_n(\textbf{w}) - \nabla V(\textbf{w})\|) + \mathbb{E}_{\textbf{z}} ( \|\underbrace{\nabla V_0(\textbf{w}_0)}_{\sum_{n\in \mathcal{N}}\nabla V_n(\textbf{w}_0)} - \nabla V(\textbf{w}))\|)\nonumber \\&& \leq 2 \times N (\sum_{n\in \mathcal{N}} \underbrace{\|\rho_n \mathbb{E}_{\textbf{z}} ( \nabla  {v}_n(\textbf{w}, \textbf{z}_n) ) \|}_{\rho_n \times \eta_n} +2 \times \underbrace{\|\nabla V(\textbf{w})\|}_{\leq \eta} + \underbrace{\mathbb{E}_{\textbf{z}} \left(\sum_{n\in \mathcal{N}}\nabla v_n(\textbf{w},\textbf{z}) - \sum_{n\in \mathcal{N}}\nabla V_n(\textbf{w})\right)}_{\sigma}\| \nonumber \\&&
     \underbrace{\leq}_{\text{from non-IID assumption}, \sigma^1_n=\sigma, \eta^1_n=\eta} 2 N \sum_{n\in \mathcal{N}}\rho_n \eta_n +2\eta  +\sigma = 2N \eta \underbrace{\sum_{n \in \mathcal{N}} \rho_n}_1 +\sigma \neq 0. \nonumber
     \end{eqnarray}
\eqref{AppendixA-2} shows that the MoE-A is biased for the non-IID local data sets and the number of users who are participating in the learning algorithm as well as the features of data sets can affect om the solution. By some minor modifications in line 3 of \eqref{AppendixA-2} and considering that the $\rho_n \leq 1$, we can attain some analysis for the case that $\mathcal{D}_0$ is non-pure in MoE-F:
\begin{eqnarray}\label{AppendixA-3}
   \lefteqn{ \mathbb{E}_{\textbf{z}} (\|\sum_{n\in \mathcal{N}} \rho_n \textbf{w}_n- \textbf{w}_0\|)}\\ && \leq  \mathbb{E}_{\textbf{z}} (\|\sum_{n\in \mathcal{E}} \rho_n\nabla \tilde{V}_n(\textbf{w}) + \sum_{n\in \mathcal{N}/\mathcal{E}} \rho_n\nabla {V}_n(\textbf{w})- \nabla V(\textbf{w})\|) + \mathbb{E}_{\textbf{z}} ( \|\underbrace{\nabla V_0(\textbf{w}_0)}_{\sum_{n\in \mathcal{N}}\nabla V_n(\textbf{w}_0)} - \nabla V(\textbf{w}))\|)\nonumber \\&& \leq  \sum_{n\in \mathcal{E}} \underbrace{\| \mathbb{E}_{\textbf{z}} ( \nabla \tilde{v}_n(\textbf{w}, \textbf{z}_n)- \nabla V_n(\textbf{w}) ) \|}_{\eta_n^2} +\sum_{n\in \mathcal{N}/\mathcal{E}} \underbrace{\| \mathbb{E}_{\textbf{z}} ( \nabla {v}_n(\textbf{w}, \textbf{z}_n)- \nabla V_n(\textbf{w}) ) \|}_{\eta_n^1} \nonumber\\&&\quad \quad \quad \quad + \|\mathbb{E}_{\textbf{z}} \sum_{n\in \mathcal{E}}\underbrace{\nabla \tilde{v}_n(\textbf{w},\textbf{z}) - \nabla V_n(\textbf{w})}_{\sigma^2_n}\| + \|\mathbb{E}_{\textbf{z}} \sum_{n\in \mathcal{N}/\mathcal{E}}\underbrace{\nabla {v}_n(\textbf{w},\textbf{z}) - \nabla V_n(\textbf{w})}_{\sigma^1_n}\| \nonumber \\&&
     \underbrace{\leq}_{\sigma^1_n=\sigma^1, \sigma^2_n=\sigma^2, \eta^1_n=\eta^1 ,\eta^2_n=\eta^2 } (N-E)\times (\sigma^1+\eta^1)+E\times (\sigma^2+\eta^2) \neq 0. \nonumber
     \end{eqnarray}
The above analysis show that the bias is linearly dependent on the number of eavesdroppers and their variance. For the case that $\mathcal{D}_0$ just contains the data sets from legitimate users, we will have 
\begin{eqnarray}\label{AppendixA-4}
   \lefteqn{ \mathbb{E}_{\textbf{z}} (\|\sum_{n\in \mathcal{N}} \rho_n \textbf{w}_n- \textbf{w}_0\|)}\\ && \leq  \mathbb{E}_{\textbf{z}} (\|\sum_{n\in \mathcal{E}} \rho_n\nabla \tilde{V}_n(\textbf{w}) + \sum_{n\in \mathcal{N}/\mathcal{E}} \rho_n\nabla {V}_n(\textbf{w})- \nabla V(\textbf{w})\|) + \mathbb{E}_{\textbf{z}} ( \|\underbrace{\nabla V_0(\textbf{w}_0)}_{\sum_{n\in \mathcal{N}}\nabla V_n(\textbf{w}_0)} - \nabla V(\textbf{w}))\|)\nonumber \\&& \leq  \sum_{n\in \mathcal{E}} \underbrace{\| \mathbb{E}_{\textbf{z}} ( \nabla \tilde{v}_n(\textbf{w}, \textbf{z}_n)- \nabla V_n(\textbf{w}) ) \|}_{\eta_n^2} +\sum_{n\in \mathcal{N}/\mathcal{E}} \underbrace{\| \mathbb{E}_{\textbf{z}} ( \nabla {v}_n(\textbf{w}, \textbf{z}_n)- \nabla V_n(\textbf{w}) ) \|}_{\eta_n^1} \nonumber\\&&+ \|\mathbb{E}_{\textbf{z}} \sum_{n\in \mathcal{N}/\mathcal{E}}\underbrace{\nabla {v}_n(\textbf{w},\textbf{z}) - \nabla V_n(\textbf{w})}_{\sigma^1_n}\| \nonumber  \underbrace{\leq}_{\sigma^1_n=\sigma^1, \sigma^2_n=\sigma^2, \eta^1_n=\eta^1 ,\eta^2_n=\eta^2 } (N-E)\times (\sigma^1+\eta^1)+E\times (\eta^2) \neq 0. \nonumber
     \end{eqnarray}

\section{Appendix B: Proof of Lemma 2} 
\begin{figure}[ht!]
\begin{center}
  \includegraphics[width=0.3\textwidth]{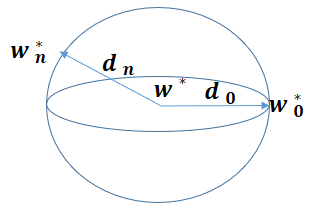}
  \caption{The distances of users ($d_n$) and the server ($d_0$) to the optimal point $\textbf{w}^*$}
\label{fig:forproofs}
\end{center}
\end{figure}

In this part, we are looking to see what are the difference between the server model to $\textbf{w}^*$ (i.e., $d_0$ in Fig. \ref{fig:forproofs}) and the users and users distance to $\textbf{w}^*$  for  $\mathcal{N}/\mathcal{E}$ (i.e., $d_n$ in Fig. \ref{fig:forproofs}). These distances show the error of convergence to the optimal $\textbf{w}^*$. The accuracy of SGD solutions for local SGD in FL setup depends on two parameters  (Theorem 3 in \cite{NEURIPS2020_45713f6f}). For the IID data sets of users when the optimization problem is convex, if we have one legitimate user in a system, the distance of its model to the server model is equal to $\frac{L}{t}+ \frac{\sigma}{\sqrt{t}}$. Therefore, the maximum distance between the model of the server and the legitimate user is equal to $2 \times (\frac{L}{t}+ \frac{\sigma}{\sqrt{t}})$. Any attackers whose variance of the model is greater than this distance can be detected by the server. The same approach can be used for the case that $\mathcal{D}_0$ includes samples from attackers (See Table 1 in \cite{NEURIPS2020_45713f6f} for the results related to \eqref{noise2}).

\section{More Iterative Approaches for MOE-FL} In general, $\mathcal{D}_0$ can bias the results of MoE
since the server dominates the solutions of all other users. Consequently, it is essential to consider the trade-off between decreasing the number of iterations and increasing the bias and over-fitting. To handle this issues, we can apply elastic SGD (ESGD) as explained in Table 2.
\begin{table}[t]
	\caption{Elastic based approaches or Newton Algorithm for Proposed FL setup} \centering \vspace{-0.0 in}
	\begin{tabular}{l}
		\hline \hline \textbf{Step 0}: Users sends their public data chunks to the server   
		\\ Server runs a model for $\tau$ epoch and send the initial weights to all users
		
		\\ \textbf{Iterative Algorithm}:\\ For $t=1,\cdots,T$ and $1\ll T$, for all $n \in[0,1, \cdots, N] $, Update\\ 
		$\quad \quad \quad\textbf{w}_n^{t+1}=(1-\alpha)\textbf{w}_n^{t}-\eta g (\textbf{w}_n^{t})+\alpha \hat{\textbf{w}}_n^{t} $		
		\\
			$\quad \quad \quad\hat{\textbf{w}}_n^{t+1}=(1-\beta)\hat{\textbf{w}}_n^{t-1}-\eta g (\textbf{w}_n^{t})+\beta 
			\bar{\textbf{w}}_n^{t} $	
		\\
		For all $n \in \mathcal {N}$, user $n$ sends its weight to the server 
		\\ Server uses \eqref{MOEoptimization} or \eqref{MOEsoft} to find $\rho_n$ and updates \eqref{updatesss}, 	
		\\ Then server sends $\textbf{w}^t$ to all users in $\mathcal{N}$\\
		If $\|\textbf{w}^{t-1}-\textbf{w}^{t}\|_{2}\leq \zeta$, End; Otherwise $t=t+1$, continue;
		\\ \hline \hline
		\vspace{-0.0 in}
	\end{tabular}\label{distributedalgorith3}
\end{table}



\ifCLASSOPTIONcaptionsoff
  \newpage
\fi



\bibliographystyle{IEEEtran}
\bibliography{ieee-journal}
\end{document}